\documentclass[11pt]{article}

\usepackage[preprint]{acl}

\usepackage{times}
\usepackage{latexsym}

\usepackage[T1]{fontenc}

\usepackage[utf8]{inputenc}

\usepackage{microtype}

\usepackage{inconsolata}

\usepackage{graphicx}

\usepackage{hyperref}
\usepackage{url}
\usepackage{booktabs}
\usepackage{tabularx}
\usepackage{array}
\usepackage{amsmath}
\usepackage{amssymb}
\usepackage{wrapfig}
\usepackage{xspace}
\usepackage{amsthm}
\usepackage{tikz}
\usepackage{makecell}
\usepackage{colortbl}
\usepackage{arydshln}
\usepackage{enumitem}
\usepackage{algorithm}
\usepackage{algpseudocode}
\usepackage{dblfloatfix}
\usepackage{xcolor}
\usepackage[many]{tcolorbox}
\usepackage{fontspec}  
\usepackage{minted} 
\tcbuselibrary{breakable}       

\usepackage{lineno}

\newcommand{\rev}[1]{#1}

\definecolor{lightgreen}{HTML}{D6F3F2}

\definecolor{darkblue}{rgb}{0, 0, 0.5}
\hypersetup{colorlinks=true, citecolor=darkblue, linkcolor=darkblue, urlcolor=darkblue}

\newtheoremstyle{compactthm}
  {0pt}        
  {0pt}        
  {\itshape}   
  {}           
  {\bfseries}  
  {.}          
  {0.5em}      
  {}           

\theoremstyle{compactthm}

\newtheoremstyle{compactremark}
  {1pt}        
  {0pt}        
  {\normalfont} 
  {}           
  {\itshape}   
  {.}          
  {0.5em}      
  {}

\theoremstyle{compactremark}

\title{
\rev{Reclaiming Low-Quality Instruction Data with Rule-Guided LLM Reconstruction}
}

\author{%
    Zile Yang$^{1}$\enskip\enskip
    Ling Li$^{1}$\enskip\enskip
    Na Di$^{1}$\enskip\enskip
    Jinlong Pang$^{2}$\enskip\enskip
    Yao Zhou\enskip\enskip\\
    {\bfseries Hao Cheng$^{3}$}\enskip\enskip 
    {\bfseries Bo Han$^{3}$}\enskip\enskip
    {\bfseries Jiaheng Wei$^{1}$}\thanks{Corresponding Author}\\
    \textsuperscript{1}The Hong Kong University of Science and Technology (Guangzhou) \enskip \\
    \textsuperscript{2}University of California, Santa Cruz \enskip
    \textsuperscript{3}Hong Kong Baptist University \enskip \\
    \{zileyang, lingli, nadi, jiahengwei\}@hkust-gz.edu.cn \\
    \{jpang14, haocheng\}@ucsc.edu \enskip\enskip bhanml@comp.hkbu.edu.hk
}

\begin{document}
\maketitle
\begin{abstract}
\rev{
Data selection for supervised fine-tuning (SFT) often discards low-scoring instruction examples that may retain useful supervision. We propose a rule-guided LLM reconstruction framework to reclaim this information. The framework identifies low-scoring samples through LLM-based rating and score correction, selects semantically representative sources, and reconstructs instruction-response pairs through strategy-guided generation, criteria-based validation, and failure-guided repair. From 123,786 low-scoring samples, it constructs 54,888 training examples. Across three 7B/8B base LLMs, the reconstructed data achieves higher average downstream performance than the low-scoring source set and all evaluated data-selection baselines. Despite using substantially fewer examples than the full source dataset ($\approx$300K), it achieves higher reported average scores than full-data training across all three base models. Stepwise ablations identify rule-guided reconstruction as the largest source of observed improvements over direct fusion, while a manual audit reveals remaining limitations in source integration and answer faithfulness. These findings support reconstruction as a complementary approach to selection-based SFT data curation.
}
\end{abstract}

\section{Introduction}
\rev{
Supervised Fine-Tuning (SFT) plays a crucial role in adapting pre-trained Large Language Models (LLMs) to instruction-following tasks. Existing data-selection methods identify informative instruction-response pairs using criteria such as quality, difficulty, diversity, and training utility~\citep{zhou2023lima,gao2021empiricalexplorationqualityfiltering,liu2024automatic,wei2024measuring,pang2024improving,xia2024less}. Despite their differences, these methods primarily retain selected samples for training while excluding lower-priority examples.
}

\begin{figure*}[!t]
    \vspace{-0.5in}
    \centering
    \includegraphics[width=0.99\textwidth]{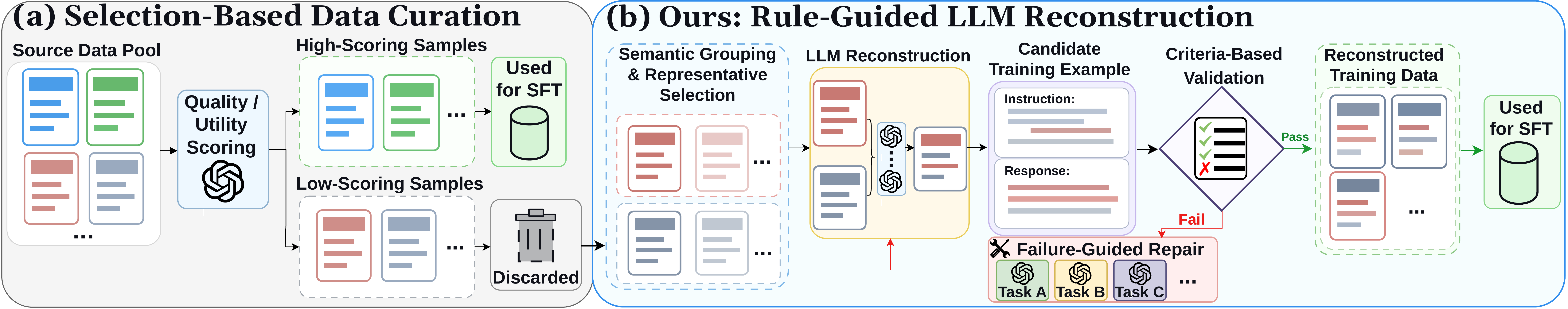}
    \caption{
    \rev{
        Comparison between selection-based data curation and rule-guided LLM reconstruction. (a) Selection-based methods retain existing samples according to quality or utility estimates, typically discarding lower-scoring examples. (b) Our framework reconstructs new training examples from low-scoring data via representative selection, LLM reconstruction, criteria-based validation, and failure-guided repair.
    }
    }
    \label{fig:teaser}
\end{figure*}

\rev{
As illustrated in Figure~\ref{fig:teaser}(a), selection-based data curation leaves the potential value of excluded samples largely unexplored. Low-scoring examples may contain incomplete questions, inadequate answers, or redundant content while still preserving useful concepts or contextual information that complements other samples. Discarding such examples may therefore remove potentially valuable supervision. This motivates a complementary perspective on SFT data curation: \emph{\textbf{rather than only asking which samples should be retained, can useful supervision be reconstructed from samples that would otherwise be discarded?}}
}

\rev{
LLM-based instruction synthesis offers a promising direction. Self-Instruct~\citep{wang2023self} generates synthetic instructions, while MathFusion~\citep{pei2025mathfusion} combines mathematical problems through predefined fusion strategies. However, reconstructing examples from low-scoring data presents additional challenges: sources may contain incomplete, redundant, or inconsistent information, while direct LLM fusion may omit important content, introduce unsupported information, or deviate from the original tasks. Addressing these issues requires identifying complementary source information, validating generated examples, and repairing specific deficiencies.
}

\rev{
To investigate this problem, we propose a \emph{rule-guided LLM reconstruction framework}, illustrated in Figure~\ref{fig:teaser}(b). We first identify low-scoring samples through LLM-based rating and score correction, followed by semantic grouping and representative selection. Our multi-stage reconstruction process then employs task-specific LLM invocations to extract key concepts and analyze source relationships, guiding the selection of predefined fusion strategies. These intermediate analyses guide the integration of complementary information into candidate instruction-response pairs.
}

\rev{
To ensure that candidates adequately preserve and integrate source information, we introduce \emph{criteria-based validation} to assess source-information coverage, question well-formedness, context completeness, and answer consistency. Detected deficiencies are mapped to predefined, task-specific repair instructions, guiding iterative LLM revisions through \emph{failure-guided repair} until the criteria are satisfied or the iteration limit is reached. This provides a structured alternative to direct one-step LLM fusion.
}

\rev{
Starting from 123,786 low-scoring samples, our framework constructs 54,888 instruction-response pairs for downstream SFT. We evaluate three 7B/8B base LLMs: Mistral-7B-v0.3 and Llama-3.1-8B on five benchmarks, and Qwen2.5-7B on four. Across these settings, the reconstructed data achieves higher reported average performance than the original low-scoring data and evaluated data-selection baselines. Despite using substantially fewer training examples, it also achieves higher reported averages than the full source dataset ($\approx$300K samples) across all three model settings.
}

\rev{
Stepwise ablations further show that replacing direct one-step fusion with rule-guided reconstruction increases the five-benchmark average from $45.6 \pm 0.3$ to $52.2 \pm 0.9$ for Mistral-7B-v0.3 and from $51.9 \pm 0.6$ to $57.4 \pm 0.3$ for Llama-3.1-8B. A source-grounded manual audit identifies remaining limitations in source integration and answer faithfulness. Together, these findings demonstrate the potential of reconstructing useful supervision from low-scoring data while highlighting challenges in reconstruction reliability.
}

Our main contributions are summarized as follows:
\begin{itemize}[leftmargin=*, itemsep=0pt, topsep=0pt]

    \item \rev{
    \textbf{Reclaiming underutilized instruction data.}
    We investigate reconstructing useful supervision from low-scoring samples as a complementary approach to selection-based SFT data curation.
    }
    
    \item \rev{
    \textbf{Rule-guided LLM reconstruction.}
    We develop a multi-stage framework combining semantic grouping, representative selection, strategy-guided generation, criteria-based validation, and failure-guided repair.
    }
    
    \item \rev{
    \textbf{Empirical and diagnostic evaluation.}
    We evaluate reconstructed data across three base LLMs, examine stage contributions through stepwise ablations, and characterize reconstruction failures through source-grounded manual analysis.
    }
\end{itemize}

\section{Related Work}
\rev{
\textbf{Instruction Data Selection.  }Existing approaches select instruction-tuning samples based on difficulty, quality, diversity, or training utility. IFD~\citep{li-etal-2024-quantity} measures instruction-following difficulty, while DEITA~\citep{liu2024makesgooddataalignment} combines complexity, quality, and diversity. DS$^2$~\citep{pang2024improving} corrects noisy LLM-generated quality scores and incorporates diversity, whereas LESS~\citep{xia2024less} estimates data influence through gradient similarity. While these methods retain existing samples, our work builds on DS$^2$ for score correction and reconstructs new supervision from low-scoring data.
}\\
\rev{
\textbf{Instruction Synthesis and Transformation.  }Self-Instruct~\citep{wang2023self} generates synthetic instruction-response pairs, while Evol-Instruct~\citep{xu2024wizardlm} iteratively increases instruction complexity. REInstruct~\citep{chen-etal-2024-reinstruct} constructs instruction data from unlabeled text through instruction generation and response rewriting. MathFusion~\citep{pei2025mathfusion} combines mathematical problems using predefined fusion strategies. In contrast, our work reconstructs low-scoring instruction-response pairs through semantic grouping and relation-dependent multi-source reconstruction.
}\\
\rev{
\textbf{Feedback-Guided Refinement.  }Self-Refine~\citep{NEURIPS2023_91edff07} iteratively revises outputs using model-generated feedback, while Auto Evol-Instruct~\citep{zeng-etal-2024-automatic} refines instruction-evolution strategies based on feedback. Our framework instead employs predefined criteria for source coverage, question and context completeness, and answer consistency, mapping detected failures to task-specific repair instructions across two sequential refinement cycles.
}\\
\rev{
General-purpose prompt optimization frameworks, such as DSPy~\citep{khattab2023dspycompilingdeclarativelanguage} and GEPA~\citep{agrawal2026gepareflectivepromptevolution}, optimize LM pipelines or prompts using task-level evaluation signals and feedback. In contrast, our Rule-Guided Prompt Updater (RPU) selects among predefined repair templates for each candidate based on structured validation reports, without learning or searching for new prompts.
}
\section{Method}
\subsection{Framework Overview}
\begin{figure*}[!t]
    \centering
    \includegraphics[width=0.99\textwidth]{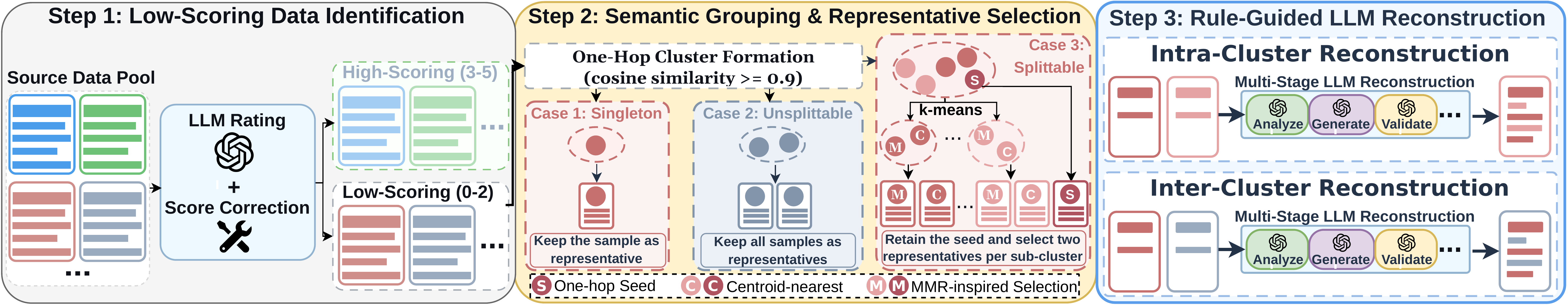}
    \caption{
    \rev{
        Overview of our three-stage rule-guided LLM reconstruction framework.
        Step (1) identifies low-scoring data through LLM-based rating and score correction.
        Step (2) forms one-hop clusters and selects representatives under three cluster cases.
        Step (3) reconstructs instruction-response pairs through multi-stage LLM operations within and across clusters.
    }
    }
    \label{fig:pipeline}
\end{figure*}

\rev{
Our framework reconstructs instruction-response pairs
from low-scoring source data through three stages,
as illustrated in Figure~\ref{fig:pipeline}.
}
\rev{
First, \emph{Low-Scoring Data Identification} uses LLM-based rating and score correction to partition source samples into high-scoring (3-5) and
low-scoring (0-2) subsets, with the latter serving as the source pool for reconstruction.
}

\rev{
Second, \emph{Semantic Grouping and Representative Selection} organizes low-scoring samples into one-hop clusters based on semantic similarity. As shown in Step~2 of Figure~\ref{fig:pipeline}, representative selection follows three cases: singleton clusters retain their sole sample; unsplittable clusters retain all samples; and splittable clusters undergo further k-means sub-clustering, retaining the initial seed sample and selecting two representatives per eligible sub-cluster: one closest to the sub-cluster centroid and another selected using an MMR-inspired scoring function that balances representativeness and diversity.
}

\rev{
Third, \emph{Rule-Guided LLM Reconstruction} constructs new instruction-response pairs from the selected samples through multiple task-specific LLM invocations. As illustrated in Step~3 of Figure~\ref{fig:pipeline}, the framework supports both intra-cluster and inter-cluster reconstruction, integrating information from representative samples within the same cluster or across different clusters, respectively. The reconstruction process incorporates source analysis, strategy-guided generation, criteria-based validation, and failure-guided repair to progressively refine candidate examples. The resulting reconstructed samples form a new training corpus for downstream SFT.
}

\subsection{Low-Scoring Data Identification}

\rev{
\textbf{LLM-Based Quality Rating.  }We employ GPT-4o-mini\footnote{Model version: gpt-4o-mini-2024-07-18} to assess each instruction sample, consisting of an instruction, an optional input, and a response. Following DS$^2$~\citep{pang2024improving}, the rating prompt evaluates rarity, complexity, and informativeness, producing an overall quality score mapped to a discrete scale from 0 to 5. The complete prompt and score mapping are provided in Appendix~\ref{appendix:llm_based_quality_rating}.
}

\rev{
\textbf{LLM-Rating Score Correction.  }To mitigate inaccuracies and inconsistencies in LLM-generated ratings~\citep{zheng2024reliablellmsknowledgebases,pang2024improving}, we adopt the score-curation procedure of DS$^2$, based on the noisy-label transition framework of \citet{zhu2021clusterability}.
}

\rev{
Let $D=\{(\mathbf{x}_n,\tilde{y}_n)\}_{n=1}^{N}$ denote the source data, where $\mathbf{x}_n$ is the embedding of the $n$-th instruction sample\footnote{Embedding model: BAAI/bge-large-en-v1.5}, $\tilde{y}_n\in\{0,\ldots,5\}$ is its initial LLM-rated score, and $y_n$ is the unobserved reference quality label in the same six-class space. Rating errors are modeled by a score transition matrix $\mathbf{T}\in\mathbb{R}^{6\times6}$:
}
\begin{equation*}
\label{eq:score_transition}
\rev{
T_{i,j}=\mathbb{P}(\tilde{y}=j\mid y=i),\quad
i,j\in\{0,\ldots,5\}.
}
\end{equation*}
\rev{
Under the 2-nearest-neighbor (2-NN) clusterability assumption, neighboring samples in embedding space share the same underlying quality class. This allows estimating the transition matrix from observed score agreements using consensus statistics up to the third order, without reference labels~\citep{zhu2021clusterability}. The consensus vector formulations and transition matrix estimation procedure are detailed in Appendix~\ref{appendix:score_transition_matrix_estimation}.
}

\rev{
Following DS$^2$, the estimated transition probabilities identify potentially misrated samples, while neighborhood-based score agreement provides candidate corrections. The resulting curated overall score is denoted by $s_n^{o}\in\{0,\ldots,5\}$.
}

\rev{
\textbf{Low-Scoring Data Partitioning.  }As illustrated in Step~1 of Figure~\ref{fig:pipeline}, samples are partitioned by their curated scores $s^o$: the low-scoring set $\mathbf{S}_{lq}$ with $s^o\in\{0,1,2\}$, and the high-scoring set $\mathbf{S}_{hq}$ with $s^o\in\{3,4,5\}$. The former serves as the source pool for reconstruction. These curated scores are estimates for data partitioning rather than independently verified ground-truth quality annotations.
}

\subsection{Semantic Grouping \& Representative Selection}

\rev{
\textbf{One-Hop Cluster Formation.  }Given the low-scoring set $\mathbf{S}_{lq}$, we randomly select a sample as the one-hop seed and group samples with cosine similarity to the seed of at least 0.9 (Step~2, Figure~\ref{fig:pipeline}). This procedure is repeated to form semantic clusters, each containing samples directly connected to its seed under the similarity threshold.
}

\rev{
\textbf{Cluster-Based Representative Selection.  }Representative selection follows three cases: singleton clusters retain their sole sample, while unsplittable clusters retain all samples. For splittable clusters, we retain the initial one-hop seed and apply k-means~\citep{macqueen1967some} to the remaining samples, selecting the number of sub-clusters $k$ by maximizing the Average Silhouette Score~\citep{rousseeuw1987silhouettes} over a predefined range (Appendix~\ref{appendix:ASS_calculation}). When at least two valid sub-clusters are obtained, each containing at least three samples, we select two representatives per sub-cluster.
}

\rev{
\textbf{Centroid- and Diversity-Based Selection.  }For each eligible sub-cluster, the first representative $r_0$ is the sample closest to the averaged sub-cluster embedding in cosine similarity. The second, $r_{\mathrm{MMR}}$, maximizes an MMR-inspired scoring function~\citep{carbonell1998use} balancing representativeness and diversity:
}
\begin{equation*}
\rev{
\begin{aligned}
&r_{\text{avg}}=\frac{1}{N}\sum_{r_a\in\mathbf{R}}r_a(.),\quad N=|\mathbf{R}|,\\
&r_0=\arg\max_{r_a\in\mathbf{R}}\mathrm{Sim}(r_a,r_{\text{avg}}),\quad \mathbf{S}=\{r_0\},\\
&\mathrm{MMR\ Score}(r_i)=\alpha\,\mathrm{Sim}(r_i,r_{\text{avg}})-\\
&\qquad \qquad \qquad \qquad (1-\alpha) \max_{r_j\in\mathbf{S}}\mathrm{Sim}(r_i,r_j),\\
&r_{\mathrm{MMR}}=\arg\max_{r_i\in\mathbf{R}\setminus\mathbf{S}}\mathrm{MMR\ Score}(r_i).
\end{aligned}
}
\end{equation*}
\rev{
Here, $\mathbf{R}$ is the sub-cluster candidate set, $\mathbf{S}$ is the selected set initialized with $r_0$, $N=|\mathbf{R}|$, $r_{\text{avg}}$ is the averaged embedding, and $\mathrm{Sim}(\cdot,\cdot)$ denotes cosine similarity. The parameter $\alpha\in[0,1]$ controls the trade-off between centroid similarity and redundancy with selected representatives; we set $\alpha=0.2$ to emphasize diversity. If the required sub-cluster conditions are not satisfied, the remaining samples are retained as representatives. The complete selection algorithm is detailed in Algorithm~\ref{alg:representative_selection} (Appendix~\ref{appendix:one_hop_cluster_based_representative_selection})
}

\subsection{Rule-Guided LLM Reconstruction}
\begin{figure*}[!t]
    \centering
    \includegraphics[width=0.99\textwidth]{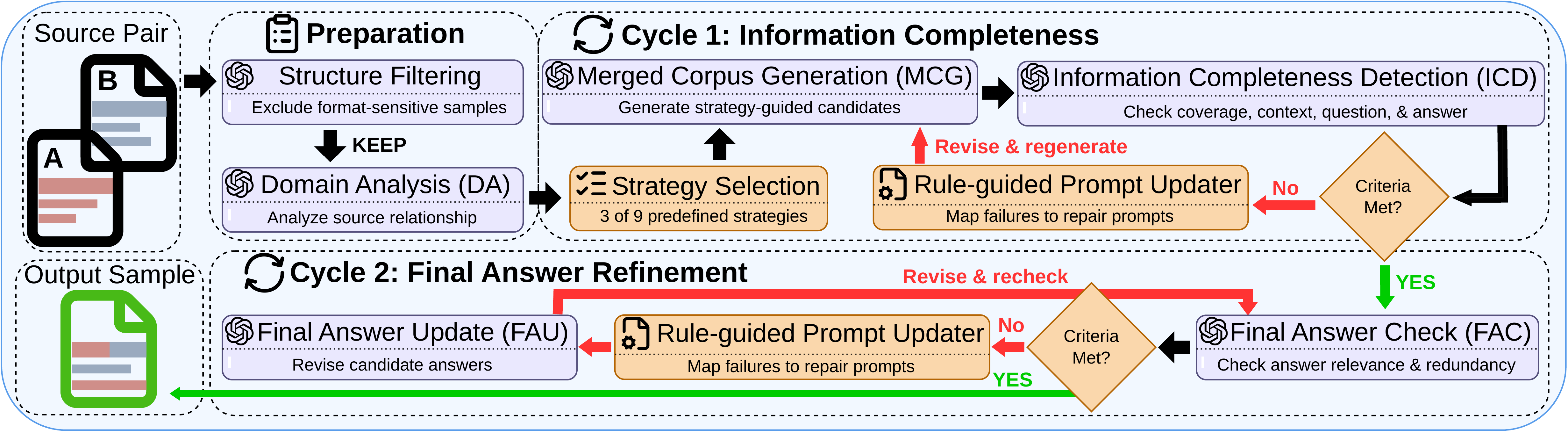}
    \caption{
    \rev{
        Detailed workflow of rule-guided LLM reconstruction.
        Preparation performs structure-dependence filtering
        and Domain Analysis (DA). Cycle 1 generates and
        validates candidates through strategy-guided Merged
        Corpus Generation (MCG), Information Completeness
        Detection (ICD), and rule-guided prompt updates.
        Cycle 2 refines responses through Final Answer Check
        (FAC), Final Answer Update (FAU), and further prompt
        updates. Purple denotes LLM-based operations, orange
        denotes rule-based operations and condition checks,
        and red/green arrows indicate revision/progression
        upon validation.
    }
    }
    \label{fig:step3_logicFlow}
\end{figure*}
\rev{
As illustrated in Figure~\ref{fig:step3_logicFlow}, this stage reconstructs instruction-response pairs from representative samples through task-specific LLM invocations and predefined rules. It supports intra-cluster and inter-cluster reconstruction using sources within or across clusters, respectively, and consists of preparation followed by two sequential refinement cycles.
}\\
\rev{
\textbf{Preparation.  }An LLM-based structure-dependence filter screens the representatives selected in Step 2, excluding samples whose meaning or correctness depends on preserving their original format, such as executable code, tables, and format-sensitive mathematical expressions. Samples classified as \texttt{FILTER} are excluded, while \texttt{KEEP} samples remain eligible. Filtering targets structural dependence rather than task domain, allowing natural-language mathematical or programming instructions to be retained.
}\\
\rev{
For each eligible source pair $(\mathcal{C}_{\mathrm{A}},\mathcal{C}_{\mathrm{B}})$, Domain Analysis (DA) uses the prompt template $\mathcal{P}_{\mathrm{DA}}$ to extract salient terms and semantic relationships, classifying the pair as same-domain, related-domain, or unrelated-domain.
}\\
\rev{
\textbf{Cycle 1: Information Completeness.  }Based on the identified relationship, Strategy Selection assigns three of nine predefined reconstruction strategies inspired by writing-studies literature ~\citep{nelson2023discourse,knobel2017remix,bazerman2003intertextuality}. Merged Corpus Generation (MCG) then produces one candidate instruction-response pair per strategy, with each candidate refined independently.
}\\
\rev{
Information Completeness Detection (ICD) evaluates each candidate for key-term coverage, question quality, context completeness, and the presence of a direct answer. Its structured report identifies retained or missing information and unmet conditions. The Rule-Guided Prompt Updater (RPU) maps detected failures to predefined repair instructions that update the MCG prompt. The revised candidate is then re-evaluated by ICD:
}
\begin{equation*}
\rev{
\begin{aligned}
\mathcal{C}_{\mathrm{AB}}^{i}
&=\mathcal{F}\bigl(
\mathcal{P}_{\mathrm{MCG}}^{i}
(\mathcal{C}_{\mathrm{A}},
\mathcal{C}_{\mathrm{B}},s)\bigr),\\
\mathcal{V}_{\mathrm{ICD}}^{i}
&=\mathcal{F}\bigl(
\mathcal{P}_{\mathrm{ICD}}
(\mathcal{C}_{\mathrm{AB}}^{i})\bigr),\\
\mathcal{P}_{\mathrm{MCG}}^{i+1}
&=\mathrm{RPU}\bigl(
\mathcal{P}_{\mathrm{MCG}}^{i},
\mathcal{V}_{\mathrm{ICD}}^{i}\bigr).\notag
\end{aligned}
}
\end{equation*}
\rev{
Here, $\mathcal{F}$ denotes an LLM-invoking operator, $\mathcal{P}$ a task-specific prompt template, and $s$ the selected reconstruction strategy. $\mathcal{C}_{\mathrm{AB}}^{i}$ is the candidate at iteration $i$, and
$\mathcal{V}_{\mathrm{ICD}}^{i}$ is its structured validation report.
}\\
\rev{
\textbf{Cycle 2: Final Answer Refinement.  }The selected Cycle 1 candidate $\mathcal{C}_{\mathrm{c1}}$ enters the second cycle for answer refinement, where only the response is revised while the instruction and its supporting context remain unchanged. Final Answer Check (FAC) evaluates whether the response directly addresses the question and identifies unnecessary, irrelevant, or redundant information. Based on its structured report, RPU selects predefined repair instructions for Final Answer Update (FAU), after which FAC re-evaluates the revised response:
}
\begin{equation*}
\rev{
\begin{aligned}
\mathcal{V}_{\mathrm{FAC}}^{i}
&=\mathcal{F}\bigl(
\mathcal{P}_{\mathrm{FAC}}
(\mathcal{C}_{\mathrm{c1}}^{i})\bigr),\\
\mathcal{P}_{\mathrm{FAU}}^{i+1}
&=\mathrm{RPU}\bigl(
\mathcal{P}_{\mathrm{FAU}}^{i},
\mathcal{V}_{\mathrm{FAC}}^{i}\bigr),\\
\mathcal{C}_{\mathrm{c1}}^{i+1}
&=\mathcal{F}\bigl(
\mathcal{P}_{\mathrm{FAU}}^{i+1}
(\mathcal{C}_{\mathrm{c1}}^{i})\bigr).\notag
\end{aligned}
}
\end{equation*}
\rev{
Here, $\mathcal{C}_{\mathrm{c1}}^{i}$ denotes the Cycle 2 candidate at iteration $i$, initialized from the selected Cycle 1 output, and $\mathcal{V}_{\mathrm{FAC}}^{i}$ is its answer-validation report.
}\\
\rev{
\textbf{Termination and Candidate Selection.  }Each cycle terminates when all validation criteria are satisfied or the iteration limit is reached. If conditions remain unmet, the candidate with the fewest unsatisfied conditions is selected, with random tie-breaking. The resulting instruction and refined response form a reconstructed corpus $\mathcal{C}_{\mathrm{AB}}$. This procedure is applied independently to candidates generated under different strategies.
}\\
\rev{
The complete procedure is provided in Algorithm~\ref{alg:rule_guided_reconstruction} (Appendix~\ref{appendix:complete_reconstruction_algorithm}), together with the nine reconstruction strategies, validation criteria, and all LLM-driven prompt templates for filtering, DA, MCG, ICD, FAC, and FAU.
}
\section{Experiments}
    
\subsection{Experimental Setup}
\textbf{Source Corpora.  }Our source pool comprises 300,932 instruction samples from Stanford Alpaca~\citep{taori2023stanford}, Flan v2~\citep{longpre2023flan}, OpenAssistant 1~\citep{kopf2023openassistant}, WizardLM~\citep{xu2024wizardlm}, and Dolly~\citep{DatabricksBlog2023DollyV2}, contributing approximately 52K, 100K, 33K, 100K, and 15K samples, respectively. \rev{The reported LQ Set retains 123,786 samples with curated scores of 0-2 after structure-dependence filtering, while the HQ Set contains 131,247 samples scored 3--5.} Detailed source-dataset statistics are provided in Appendix~\ref{appendix:source_corpora}. \\
\textbf{Evaluation Dataset \& Metrics.  }We evaluate on five benchmarks: MMLU~\citep{hendrycks2020measuring}, TruthfulQA~\citep{lin2021truthfulqa}, GSM8K~\citep{cobbe2021training}, BBH~\citep{suzgun2022challenging}, and TyDiQA~\citep{clark2020tydi}. We report Exact Match (EM) for MMLU, GSM8K, and BBH; the Informative-Truthful Rate ($I \times T$) for TruthfulQA; and 1-shot F1 for TyDiQA. Mistral and Llama are evaluated on all five benchmarks, while Qwen excludes TruthfulQA. Detailed evaluation protocols are provided in Appendix~\ref{appendix:evaluation_setup}. \\
\textbf{Base Models \& Rating Model.  }We adopt three widely recognized open-source LLMs as our fine-tuning base models: Mistral-7B-v0.3~\citep{jiang2023mistral7b}, LLaMA-3.1-8B~\citep{grattafiori2024llama}, and Qwen2.5-7B~\citep{qwen2.5}. GPT-4o-mini is used for LLM-based quality rating and reconstruction. \\
\textbf{Baselines.  }
Our framework reconstructs 54,888 training samples from the LQ Set (123,786 samples), comprising 15,488 intra-cluster and 39,400 inter-cluster samples.  All data-selection, instruction-synthesis, and reconstruction-ablation baselines share the same LQ Set source pool as our method, while the Vanilla Base Model, HQ Set, and Full Set serve as separate reference settings. \\
We compare against four control settings (Vanilla Base Model, LQ Set, HQ Set, and Full Set); data-selection baselines including Completion Length, $\mathrm{KNN}_{10}$, Perplexity, Random Selection, AlpaGasus (Random)~\citep{chen2023alpagasus}, IFD~\citep{li-etal-2024-quantity}, Superfiltering~\citep{li2024superfiltering}, DEITA~\citep{liu2024makesgooddataalignment}, DEITA (Our Curated Score), RDS+ and RDS+ (Best)~\citep{ivison2025large}, DS$^2$~\citep{pang2024improving}, and LESS~\citep{xia2024less}; instruction-synthesis baselines Self-Instruct~\citep{wang2023self} and \rev{MathFusion (conditional)~\citep{pei2025mathfusion}}; and three additional reconstruction comparisons: 1-to-1 Rewriting/Enhancement, Direct Corpora Fusion Without Step2\&3, and Direct Corpora Fusion Without Step3.
Unless otherwise specified, the target training data size is 54,888 samples. Detailed baseline implementations are provided in Appendix~\ref{appendix:baseline}. \\
\textbf{Implementation Details.  }For one-hop clustering, we use a cosine similarity threshold of 0.9. For representative selection, we set $\alpha=0.2$ to emphasize diversity. GPT-4o-mini serves as the API-accessed LLM for reconstruction, with temperature 0.4 for Domain Analysis (DA) and 0.2 for the other LLM-based reconstruction operators. We consider both intra-cluster reconstruction, which progressively integrates samples within the same cluster, and inter-cluster reconstruction, which combines samples from different clusters. Each cycle permits at most three candidate generations, including the initial candidate and up to two regenerations. \\
\rev{
For Mistral-7B-v0.3 and Llama-3.1-8B, we evaluate the Full Set, two direct-fusion ablations, and our method using two downstream SFT seeds (42 and 47), reporting mean $\pm$ sample standard deviation in Tables~\ref{tab:main_table_mistral} and~\ref{tab:main_table_llama}. The remaining results in these two tables are from single runs. Complete reconstruction and SFT hyperparameters are provided in Appendix~\ref{appendix:training_setup}.
}
\begin{table}[!htb]
    \setlength{\aboverulesep}{0pt}
    \setlength{\belowrulesep}{0pt}
    \vspace{-0.2cm}
    \caption{
    \rev{
    \textbf{Performance comparison across five benchmarks (Mistral-7B-v0.3).}
    The $\Delta$ row reports the difference between the two-seed mean of Ours and the single-run LQ Set baseline. Bold red and bold black indicate the \textbf{\color{red}{best}} and \textbf{second-best} Average scores, respectively.
    }
    }
    \label{tab:main_table_mistral}
    \centering
    \vspace{-2mm}
    \scalebox{0.35}{
    \begin{tabular}{lcccccc}
        \toprule
        \textbf{Dataset} & \makecell{\textbf{MMLU} \\ (EM $\uparrow$)} & \makecell{\textbf{TruthfulQA} \\ \rev{(I $\times$ T $\uparrow$)}} & \makecell{\textbf{GSM8K} \\ (EM $\uparrow$)} & \makecell{\textbf{BBH} \\ (EM $\uparrow$)} & \makecell{\textbf{TydiQA} \\ (1-Shot F1 $\uparrow$)} & \textbf{Average $\uparrow$}\\
        \midrule
        \multicolumn{7}{>{\columncolor{lightgreen}}c}{\textbf{Base Model: Mistral-7B-v0.3}} \\
        Vanilla Base Model & 59.7 & 30.4 & 38.0 & 47.6 & 54.8 & 46.1 \\
        LQ Set (123786) & 47.5 & 43.7 & 43.5 & 52.7 & 41.1 & 45.7 \\
        HQ Set (131247) & 58.4 & 39.2 & 46.0 & 55.5 & 52.5 & \textbf{50.3} \\
        \rev{Full Set (300932)} & 52.6 $\pm$ 10.5 & 43.1 $\pm$ 0.6 & 43.5 $\pm$ 0.0 & 53.2 $\pm$ 0.9 & 51.4 $\pm$ 2.9 & 48.7 $\pm$ 2.6 \\
        \noalign{%
            \setlength{\arrayrulewidth}{1pt}
            \setlength{\dashlinedash}{5pt}
            \setlength{\dashlinegap}{2pt}
        }%
        \cdashline{1-7} 
        Completion Length & 25.4 & 43.5 & 43.0 & 55.7 & 45.8 & 42.7 \\
        $\mathrm{KNN}_{10}$ & 53.7 & 43.7 & 40.0 & 54.4 & 39.5 & 46.3 \\
        Perplexity & 53.8 & 41.8 & 34.5 & 54.8 & 41.9 & 45.4 \\
        Random Selection & 52.8 & 42.0 & 41.5 & 56.7 & 48.4 & 48.3 \\
        \noalign{%
            \setlength{\arrayrulewidth}{1pt}
            \setlength{\dashlinedash}{5pt}
            \setlength{\dashlinegap}{2pt}
        }%
        \cdashline{1-7} 
        AlpaGasus (Random) & 53.8 & 48.5 & 44.5 & 55.6 & 41.5 & 48.8 \\
        IFD & 40.3 & 43.8 & 44.0 & 49.5 & 33.6 & 42.2 \\
        Superfiltering & 51.8 & 40.7 & 45.0 & 52.6 & 37.8 & 45.6 \\
        DEITA & 44.5 & 39.9 & 43.5 & 50.2 & 46.1 & 44.8 \\
        DEITA (Our Curated Score) & 52.2 & 36.6 & 44.0 & 54.3 & 51.7 & 47.8 \\
        RDS+ & 47.9 & 41.1 & 43.0 & 52.9 & 41.8 & 45.3 \\
        RDS+ (Best) & 51.0 & 43.4 & 46.0 & 54.9 & 44.6 & 48.0 \\
        $\mathrm{DS}^{2}$ & 48.7 & 44.1 & 47.5 & 55.1 & 46.9 & 48.5 \\
        LESS & 54.1 & 46.2 & 44.0 & 53.8 & 50.5 & 49.7 \\
        Self-Instruct & 53.1 & 43.8 & 45.0 & 55.2 & 50.9 & 49.6 \\
        \rev{MathFusion (conditional)} & 51.4 & 58.2 & 36.0	& 51.6 & 41.9 & 47.8 \\
        \cdashline{1-7} 
        1-to-1 Rewriting/ Enhancement & 47.3 & 42.4 & 41.5 & 49.4 & 49.9 & 46.1 \\
        \rev{Direct Corpora Fusion Without Step2\&3} & 40.8 $\pm$ 0.6	& 41.9 $\pm$ 0.7	& 38.3 $\pm$ 1.1 & 49.4 $\pm$ 0.8 &	50.1 $\pm$ 0.8 & 44.1 $\pm$ 0.5 \\
        \rev{Direct Corpora Fusion Without Step3} & 45.3 $\pm$ 0.8 & 42.6 $\pm$ 0.3 & 39.8 $\pm$ 0.4 & 49.7 $\pm$ 0.5	& 50.8 $\pm$ 0.6 & 45.6 $\pm$ 0.3 \\
        \noalign{%
            \setlength{\arrayrulewidth}{1pt}
            \setlength{\dashlinedash}{5pt}
            \setlength{\dashlinegap}{2pt}
        }%
        \cdashline{1-7} 
        \textbf{\rev{Ours}} & 55.7 $\pm$ 4.2 & 49.4 $\pm$ 9.0	& 47.3 $\pm$ 4.6 & 53.7 $\pm$ 0.2 & 55.1 $\pm$ 4.6	& \textbf{\textcolor{red}{52.2 $\pm$ 0.9}} \\
        \rev{$\Delta$ vs. LQ Set} & +8.2 & +5.7 & +3.8 & +1.0 & +14.0 & +6.5 \\
        \bottomrule
        \vspace{-0.5cm}
    \end{tabular}
    }
\end{table}

\begin{table}[!htb]
    \setlength{\aboverulesep}{0pt}
    \setlength{\belowrulesep}{0pt}
    \vspace{-0.5cm}
    \caption{
    \rev{
    \textbf{Performance comparison across five benchmarks (Llama-3.1-8B).}
    The $\Delta$ row reports the difference between the two-seed mean of Ours and the single-run LQ Set baseline. Bold red and bold black indicate the \textbf{\color{red}{best}} and \textbf{second-best} Average scores, respectively.
    }
    }
    \label{tab:main_table_llama}
    \centering
    \vspace{-2mm}
    \scalebox{0.35}{
    \begin{tabular}{lcccccc}
        \toprule
        \textbf{Dataset} & \makecell{\textbf{MMLU} \\ (EM $\uparrow$)} & \makecell{\textbf{TruthfulQA} \\ \rev{(I $\times$ T $\uparrow$)}} & \makecell{\textbf{GSM8K} \\ (EM $\uparrow$)} & \makecell{\textbf{BBH} \\ (EM $\uparrow$)} & \makecell{\textbf{TydiQA} \\ (1-Shot F1 $\uparrow$)} & \textbf{Average $\uparrow$}\\
        \midrule
        \multicolumn{7}{>{\columncolor{lightgreen}}c}{\textbf{Base Model: Llama-3.1-8B}} \\
        Vanilla Base Model & 64.1 & 32.9 & 58.0 & 55.3 & 22.1 & 46.5 \\
        LQ Set (123786) & 52.7 & 44.3 & 57.0 & 61.0 & 43.9 & 51.8 \\
        HQ Set (131247) & 62.3 & 41.8 & 57.5 & 59.3 & 58.9 & 56.0 \\
        \rev{Full Set (300932)} & 62.1 $\pm$ 2.0 & 40.7 $\pm$ 1.9 & 60.8 $\pm$ 0.4 & 59.2 $\pm$ 0.1 & 61.4 $\pm$ 2.1 & 56.8 $\pm$ 1.3 \\
        \noalign{%
            \setlength{\arrayrulewidth}{1pt}
            \setlength{\dashlinedash}{5pt}
            \setlength{\dashlinegap}{2pt}
        }%
        \cdashline{1-7} 
        Completion Length & 59.5 & 45.8 & 60.0 & 58.6 & 61.2 & \textbf{57.0} \\
        $\mathrm{KNN}_{10}$ & 59.3 & 37.6 & 52.0 & 59.5 & 55.7 & 52.8 \\
        Perplexity & 60.5 & 37.5 & 51.0 & 57.8 & 55.0 & 52.4 \\
        Random Selection & 60.2 & 38.0 & 57.0 & 57.7 & 60.4 & 54.7 \\
        \noalign{%
            \setlength{\arrayrulewidth}{1pt}
            \setlength{\dashlinedash}{5pt}
            \setlength{\dashlinegap}{2pt}
        }%
        \cdashline{1-7} 
        AlpaGasus (Random) & 53.0 & 44.1 & 56.0 & 59.4 & 49.0 & 52.3 \\
        IFD & 45.9 & 46.7 & 57.0 & 56.6 & 42.8 & 49.8 \\
        Superfiltering & 58.0 & 43.8 & 61.5 & 56.5 & 51.0 & 54.2 \\
        DEITA & 57.6 & 43.3 & 58.5 & 59.0 & 60.8 & 55.8 \\
        DEITA (Our Curated Score) & 60.0 & 46.8 & 58.0 & 58.1 & 61.3 & 56.8 \\
        RDS+ & 57.1 & 43.6 & 52.5 & 58.6 & 42.0 & 50.8 \\
        RDS+ (Best) & 57.1 & 46.6 & 59.5 & 60.8 & 53.0 & 55.4 \\
        $\mathrm{DS}^{2}$ & 59.9 & 44.8 & 55.5 & 58.2 & 60.8 & 55.8  \\
        LESS & 59.9 & 40.5 & 56.0 & 61.4 & 66.2 & 56.8 \\
        Self-Instruct & 52.1 & 18.3 & 56.5 & 57.9 & 57.3 & 48.4 \\
        \rev{MathFusion (conditional)} & 58.3	& 58.2 & 55.5 & 59.5 & 52.6 & 56.8 \\
        \cdashline{1-7} 
        1-to-1 Rewriting/ Enhancement & 58.3 & 42.5 & 58.5 & 57.9 & 46.1 & 52.7 \\
        \rev{Direct Corpora Fusion Without Step2\&3} & 54.2 $\pm$ 0.1 & 40.0 $\pm$ 1.0 & 59.8 $\pm$ 0.4 & 58.7 $\pm$ 2.5 & 43.4 $\pm$ 0.4 & 51.2 $\pm$ 0.3 \\
        \rev{Direct Corpora Fusion Without Step3} & 54.8 $\pm$ 0.4 & 48.0 $\pm$ 1.1 & 56.5 $\pm$ 0.7 & 57.3 $\pm$ 1.4 &	42.8 $\pm$ 4.5 & 51.9 $\pm$ 0.6 \\
        \noalign{%
            \setlength{\arrayrulewidth}{1pt}
            \setlength{\dashlinedash}{5pt}
            \setlength{\dashlinegap}{2pt}
        }%
        \cdashline{1-7} 
        \textbf{\rev{Ours}} & 59.6 $\pm$ 3.0 & 47.2 $\pm$ 0.8 & 56.5 $\pm$ 2.8 & 60.9 $\pm$ 0.3 & 62.8 $\pm$ 2.1 & \textbf{\textcolor{red}{57.4 $\pm$ 0.3}} \\
        \rev{$\Delta$ vs. LQ Set} & +6.9 & +2.9 & -0.5 & -0.1 & +18.9 & +5.6 \\
        \bottomrule
        \vspace{-2cm}
    \end{tabular}
    }
\end{table}

\subsection{Empirical Observations}
We summarize results for Mistral-7B-v0.3 and Llama-3.1-8B in Tables~\ref{tab:main_table_mistral} and~\ref{tab:main_table_llama}, with additional results provided in Appendix~\ref{appendix:emprical_observations}. \\
\rev{
\textcolor{red}{\textbf{Observation 1: }}\textbf{Low-scoring data retains useful supervision despite variations in selection effectiveness.} As shown in Tables~\ref{tab:main_table_mistral} and~\ref{tab:main_table_llama}, data-selection baselines applied to the LQ Set achieve average scores ranging from 42.2 (IFD) to 49.7 (LESS) for Mistral, and from 49.8 (IFD) to 57.0 (Completion Length) for Llama, compared with LQ Set scores of 45.7 and 51.8, respectively. Notably, Completion Length (57.0) and DEITA (Our Curated Score) (56.8) both exceed the HQ Set (56.0) for Llama. These results suggest that low-scoring samples retain useful supervision, although selection effectiveness depends on the method and base model.
}\\
\rev{
\textcolor{red}{\textbf{Observation 2: }}\textbf{Reconstruction achieves the highest average performance across both base models.} As shown in Tables~\ref{tab:main_table_mistral} and~\ref{tab:main_table_llama}, our method achieves the highest reported five-benchmark averages for Mistral ($52.2 \pm 0.9$) and Llama ($57.4 \pm 0.3$), exceeding the strongest data-selection baselines, LESS (49.7) and Completion Length (57.0), respectively. Compared with the single-run LQ Set baselines, our two-seed mean scores are higher by 6.5 and 5.6 points. These findings support reconstruction as a promising alternative to selecting existing low-scoring samples alone.
}\\
\rev{
\textcolor{red}{\textbf{Observation 3: }}\textbf{Reconstruction improves training-data efficiency, with task-dependent gains.} Using 54,888 reconstructed samples (18.2\% of the 300,932-sample Full Set), our method achieves higher reported averages than both the HQ Set (50.3 and 56.0) and Full Set ($48.7 \pm 2.6$ and $56.8 \pm 1.3$) for Mistral and Llama, respectively. Relative to the LQ Set, substantial gains occur on TyDiQA (+14.0 and +18.9 points) and MMLU (+8.2 and +6.9 points), although Llama shows slight decreases on GSM8K (-0.5) and BBH (-0.1). These results suggest that reconstruction improves the utility of a smaller training corpus, with benefits varying across tasks.
}
\begin{table}[!htb]
    \setlength{\aboverulesep}{0pt}
    \setlength{\belowrulesep}{0pt}
    \caption{
    \rev{
        \textbf{Data-size ablation of reconstructed training data.}
        Results with $\pm$ indicate mean $\pm$ sample standard deviation over two seeds (42 and 47); others are from single runs. The Average is computed over the four benchmarks. Best and second-best Average scores are highlighted in \textbf{\textcolor{red}{bold red}} and \textbf{bold black}, respectively.
    }
    }
    \label{tab:data_size_ablation}
    \centering
    \vspace{-2mm}
    \scalebox{0.5}{
    \begin{tabular}{lccccc}
        \toprule
        \textbf{Dataset} & \makecell{\textbf{MMLU} \\ (EM $\uparrow$)} & \makecell{\textbf{GSM8K} \\ (EM $\uparrow$)} & \makecell{\textbf{BBH} \\ (EM $\uparrow$)} & \makecell{\textbf{TydiQA} \\ (1-Shot F1 $\uparrow$)} & \textbf{Average $\uparrow$}\\
        \midrule
        \multicolumn{6}{>{\columncolor{lightgreen}}c}{\textbf{Base Model: Mistral-7B-v0.3}} \\
        LQ Set (123786) &  47.5 & 43.5 & 52.7 & 41.1 & \rev{46.2} \\
        HQ Set (131247) & 58.4 & 46.0 & 55.5 & 52.5 & \textbf{\color{red}{53.1}} \\
        \rev{Full Set (300932)} & 52.6 $\pm$ 10.5 & 43.5 $\pm$ 0.0 & 53.2 $\pm$ 0.9 & 51.4 $\pm$ 2.9 & 50.1 $\pm$ 3.1 \\
        \noalign{%
            \setlength{\arrayrulewidth}{1pt}
            \setlength{\dashlinedash}{5pt}
            \setlength{\dashlinegap}{2pt}
        }%
        \cdashline{1-6} 
        \rev{Ours-20\%} & 56.7 $\pm$ 3.7 & 42.8 $\pm$ 1.8 & 52.2 $\pm$ 3.4 & 53.2 $\pm$ 3.2 & 51.2 $\pm$ 2.1 \\
        \rev{Ours-60\%} & 56.0 $\pm$ 4.4 & 45.8 $\pm$ 0.4 & 52.5 $\pm$ 0.3 & 53.2 $\pm$ 4.2 & 51.9 $\pm$ 2.0 \\
        \rev{Ours-100\%} & 55.7 $\pm$ 4.2 & 47.3 $\pm$ 4.6 & 53.7 $\pm$ 0.2 & 55.1 $\pm$ 4.6 & \textbf{52.9 $\pm$ 1.1} \\
        \multicolumn{6}{>{\columncolor{lightgreen}}c}{\textbf{Base Model: Llama-3.1-8B}} \\
        LQ Set (123786) & 52.7 & 57.0 & 61.0 & \rev{43.9} & \rev{53.7} \\
        HQ Set (131247) & 62.3 & 57.5 & 59.3 & 58.9 & 59.5 \\
        \rev{Full Set (300932)} & 62.1 $\pm$ 2.0 & 60.8 $\pm$ 0.4 & 59.2 $\pm$ 0.1 & 61.4 $\pm$ 2.1 & \textbf{\color{red}{60.8 $\pm$ 1.1}} \\
        \noalign{%
            \setlength{\arrayrulewidth}{1pt}
            \setlength{\dashlinedash}{5pt}
            \setlength{\dashlinegap}{2pt}
        }%
        \cdashline{1-6} 
        \rev{Ours-20\%} & 60.8 $\pm$ 4.3 & 56.0 $\pm$ 2.1 & 59.2 $\pm$ 3.3	& 54.3 $\pm$ 2.5 & 57.6 $\pm$ 1.8 \\
        \rev{Ours-60\%} & 59.7 $\pm$ 3.7 & 56.3 $\pm$ 1.8 & 59.0 $\pm$ 1.6	& 60.5 $\pm$ 4.7 & 58.8 $\pm$ 0.6 \\
        \rev{Ours-100\%} & 59.6 $\pm$ 3.0 & 56.5 $\pm$ 2.8 & 60.9 $\pm$ 0.3	& 62.8 $\pm$ 2.1 & \textbf{59.9 $\pm$ 0.5} \\
    \bottomrule
    \end{tabular}
    }
    \vspace{-0.5cm}
\end{table}

\subsection{Ablation Study}
\rev{
\textbf{Stepwise Ablation.  }To assess the contributions of semantic grouping and rule-guided LLM reconstruction, we compare the full framework with direct-fusion configurations omitting Steps 2 and 3, or Step 3 alone. As shown in Tables~\ref{tab:main_table_mistral} and~\ref{tab:main_table_llama}, introducing Step 2 increases the five-benchmark average from $44.1 \pm 0.5$ to $45.6 \pm 0.3$ for Mistral-7B-v0.3, and from $51.2 \pm 0.3$ to $51.9 \pm 0.6$ for Llama-3.1-8B. Incorporating Step 3 further raises performance to $52.2 \pm 0.9$ and $57.4 \pm 0.3$, respectively, yielding additional gains of 6.6 and 5.5 points. These results suggest that both stages contribute to downstream performance, with larger gains from rule-guided reconstruction.
}\\
\rev{
\textbf{Data-Size Ablation.  }We evaluate 20\%, 60\%, and 100\% of the 54,888 reconstructed samples, with smaller subsets randomly sampled. As shown in Table~\ref{tab:data_size_ablation}, the four-benchmark average increases from $51.2 \pm 2.1$ to $51.9 \pm 2.0$ and $52.9 \pm 1.1$ for Mistral, and from $57.6 \pm 1.8$ to $58.8 \pm 0.6$ and $59.9 \pm 0.5$ for Llama. Both models show increasing mean performance across these data proportions, although individual benchmark results vary. This trend suggests benefits from additional reconstructed samples without establishing a general scaling law. The single-run results for all five data proportions, including the additional 40\% and 80\% configurations, are provided in Table~\ref{tab:ablation_ours_mistral_llama} (Appendix~\ref{appendix:ablation}).
}

\subsection{Manual Quality Audit}
\begin{table}[t]
\centering
\caption{
\rev{
\textbf{Manual quality audit of 30 reconstructed samples across five criteria.}
}
}
\label{tab:manual_quality_audit}
\small
\setlength{\tabcolsep}{5pt}
\scalebox{1}{
    \begin{tabular}{lcc}
        \toprule
        \textbf{Criterion} & \textbf{Passed} & \textbf{Rate} \\
        \midrule
        Source integration & 20/30 & 66.7\% \\
        Context completeness & 28/30 & 93.3\% \\
        Question clarity & 28/30 & 93.3\% \\
        Answer faithfulness & 18/30 & 60.0\% \\
        Overall usability & 14/30 & 46.7\% \\
        \midrule
        \textbf{All five criteria} &
        \textbf{14/30} & \textbf{46.7\%} \\
        \bottomrule
    \end{tabular}
}
\end{table}
\rev{
To assess the quality of reconstructed data beyond downstream benchmark performance, we manually audit 30 randomly sampled instruction-response pairs from the 54,888 reconstructed samples. Each sample is examined alongside its source corpora using five binary criteria: source integration, context completeness, question clarity, answer faithfulness and consistency, and overall transformation usability. The evaluation criteria and annotation procedure are detailed in Appendix~\ref{appendix:manual_quality_audit}.
}

\rev{
As shown in Table~\ref{tab:manual_quality_audit}, context completeness and question clarity each achieve a 93.3\% pass rate. However, source integration and answer faithfulness reach only 66.7\% and 60.0\%, respectively, with 46.7\% of samples satisfying all five criteria. Further analysis reveals higher source-integration pass rates for pairwise inter-cluster reconstruction (11/13, 84.6\%) than for multi-source intra-cluster reconstruction (9/17, 52.9\%). These findings indicate that well-formed questions do not necessarily guarantee faithful source integration or consistent answers, particularly when multiple sources must be consolidated. Given the small audit sample, these findings should be interpreted descriptively rather than as precise estimates of quality across the entire reconstructed dataset.
}
\section{Conclusion}
\rev{
We introduced a rule-guided LLM reconstruction framework that reclaims useful supervision from low-scoring instruction data. By combining semantic grouping and representative selection with strategy-guided generation, criteria-based validation, and failure-guided repair, our framework constructs 54,888 instruction--response pairs from 123,786 low-scoring samples. Across three base LLMs, fine-tuning on the reconstructed data achieves higher average downstream performance than the original low-scoring set and evaluated data-selection baselines. Stepwise ablations show that rule-guided reconstruction yields the largest observed performance gains. A source-grounded manual audit, however, reveals remaining challenges in source integration and answer faithfulness. These findings support reconstruction as a complementary approach to selection-based SFT data curation while highlighting the importance of improving reconstruction reliability.
}

\section{Limitations}
\rev{
Our framework primarily targets natural-language
instruction data. Reconstruction remains challenging
for structure-dependent inputs, such as code,
tables, and format-sensitive mathematical expressions,
where preserving the original structure is essential
to maintaining meaning or correctness.
We therefore apply LLM-based structure-dependence
filtering before reconstruction, restricting the
current scope of our approach rather than resolving
these structural challenges.
}

\rev{
Although criteria-based validation and failure-guided
repair improve the reconstruction process, they
do not guarantee that every generated example
faithfully preserves its source information.
Our manual quality audit identifies remaining
limitations in source integration and answer
consistency, while its small sample size limits
broader conclusions about corpus-wide quality.
Additionally, reconstruction relies on an external
LLM and multiple API invocations, introducing
offline data-construction costs and potential
sensitivity to the underlying model.
}

\rev{
Furthermore, reconstruction may alter the task and domain composition of the training data alongside changes in model-estimated quality. Therefore, the observed downstream improvements cannot be attributed solely to improved sample quality without distribution-matched comparisons.
}

\bibliography{emnlp2026_reference}

\clearpage
\appendix
\section*{Appendix}

\paragraph{Artifact License and Intended Use.}
All external artifacts used in this work, including datasets, benchmark suites, and pretrained models, are publicly available and used in accordance with their respective licenses, terms of use, and intended research purposes. Any derived artifacts produced in this work are intended for research use only and remain subject to the access conditions of the original artifacts.

\paragraph{Data Privacy and Content Risk.}
We use publicly available datasets and benchmark corpora for research purposes. We do not collect private user data or intentionally include personally identifying information. We also conduct manual inspection of sampled data to check for personally identifying or offensive content, and any such content, if identified, is excluded or anonymized.
\paragraph{Use of Large Language Models.}
Our framework employs GPT-4o-mini for instruction quality assessment, structure-dependence filtering, and multi-stage instruction-response reconstruction, including candidate generation, validation, and refinement. These LLM-based operations are guided by predefined prompts and rules.
\paragraph{Reproducibility and Data Availability.}
To facilitate reproducibility, we provide the reconstruction algorithms, prompt templates, experimental configurations, and evaluation protocols in the Appendix. Upon acceptance, we plan to release our implementation, baseline scripts, and reconstructed instruction-response datasets, including those generated through intra-cluster and inter-cluster reconstruction, subject to applicable licensing restrictions. As the framework involves stochastic LLM generation and fine-tuning, exact outputs and evaluation results may vary across runs. \\

The rest of Appendix is organized as follows:
\begin{itemize}[leftmargin=*]
    \item \rev{\textbf{Appendix A: Quality Rating and Mathematical Preliminaries.}
    We provide the LLM-based quality-rating prompt, score mapping procedure, score transition matrix estimation, and Average Silhouette Score formulation.
    }
    
    \item \textbf{Appendix B: Detailed Algorithms and Reconstruction Prompts.}
    We present the complete semantic grouping and representative selection algorithm, followed by the rule-guided LLM reconstruction algorithm, predefined reconstruction strategies, validation criteria, failure-guided repair rules, and task-specific LLM prompt templates.

    \item \rev{
    \textbf{Appendix C: End-to-End Reconstruction Example.}
    We illustrate the reconstruction process using a concrete execution example, tracing source samples through domain analysis, strategy selection, candidate generation, validation, and failure-guided refinement.
    }
    
    \item \textbf{Appendix D: Experimental Details and Cost Analysis.}
    We provide source-dataset statistics, evaluation protocols, baseline implementations, \rev{reconstruction and SFT hyperparameters}, computational costs, and validation-set requirements.

    \item \textbf{Appendix E: Additional Results and Ablation Studies.}
    We report supplementary model evaluations, additional data-size ablations, and curated score distributions of the source and reconstructed datasets.

    \item \rev{\textbf{Appendix F: Manual Quality Audit and Qualitative Examples.}
    We describe the annotation protocol and five evaluation criteria, report aggregate manual audit results and illustrative case-level judgments, and present representative successful and failed reconstruction examples.
    }
\end{itemize}

\section{Quality Rating and Mathematical Preliminaries}
\label{appendix:preliminary}

This section provides the technical details underlying two components of our framework: LLM-based quality rating and score correction (Section 3.2), and Average Silhouette Score-based sub-cluster selection (Section 3.3).

\subsection{LLM-Based Quality Rating and Score Mapping}
\label{appendix:llm_based_quality_rating}

Following DS$^2$~\citep{pang2024improving}, we employ GPT-4o-mini to assess each instruction sample consisting of an instruction, an optional input, and a response. The rating prompt evaluates four dimensions: Rarity, Complexity, Informativeness, and Overall Rating, each initially scored on a scale from 1 to 10, with higher scores indicating better quality.

Let $r_n \in \{1,\ldots,10\}$ denote the Overall Rating assigned to sample $n$. Following DS$^2$, we map this rating to a six-class score space $\{0,\ldots,5\}$ as follows:
\begin{equation}
\tilde{y}_n =
\begin{cases}
0, & 1 \leq r_n \leq 4,\\
r_n - 4, & 5 \leq r_n \leq 8,\\
5, & 9 \leq r_n \leq 10,
\end{cases}
\label{eq:initial_score_mapping}
\end{equation}
where $\tilde{y}_n$ denotes the resulting initial quality score of sample $n$.

\begin{figure*}[t]
    \centering
    \includegraphics[
        width=\textwidth,
        height=\textheight,
        keepaspectratio
    ]{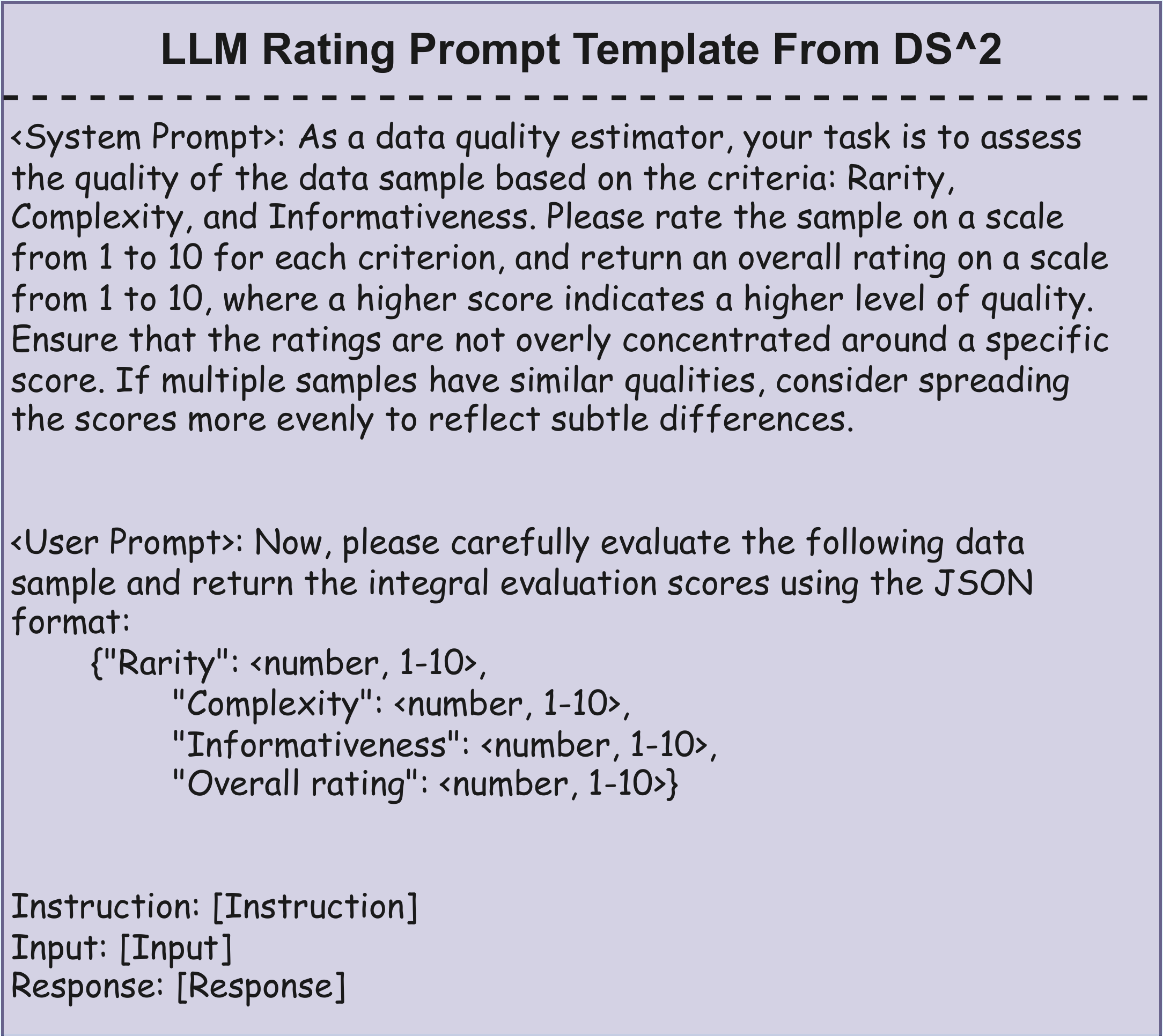}
    \caption{LLM quality-rating prompt template adopted from DS$^2$~\citep{pang2024improving}.}
    \label{appendix:fig_llm_rating_prompt}
\end{figure*}

\paragraph{LLM Rating Prompt Template.}
The complete rating prompt adopted from DS$^2$~\citep{pang2024improving} is provided in Figure~\ref{appendix:fig_llm_rating_prompt}.

\subsection{Score Transition Matrix Estimation}
\label{appendix:score_transition_matrix_estimation}

\paragraph{Score Transition Matrix.}
Following DS$^2$~\citep{pang2024improving}, we model potential inaccuracies in LLM-generated quality ratings using the noisy-label transition framework of \citet{zhu2021clusterability}.

Let $D=\{(\mathbf{x}_n,\tilde{y}_n)\}_{n=1}^{N}$ denote the source data, where $\mathbf{x}_n$ is the embedding of sample $n$, obtained using BAAI/bge-large-en-v1.5, and $\tilde{y}_n$ is its initial LLM-rated score. The unobserved reference quality label is denoted by $y_n$.

Both $\tilde{y}_n$ and $y_n$ are modeled in the same $K$-class space, where $K=6$. Under the class-conditional noise assumption, the score transition matrix $\mathbf{T}\in\mathbb{R}^{K\times K}$ is defined as
\begin{equation}
T_{i,j}
=
\mathbb{P}(\tilde{y}=j\mid y=i),
\quad
i,j\in\{0,\ldots,K-1\}.
\label{eq:app:score_transition_matrix}
\end{equation}
Each entry $T_{i,j}$ represents the probability that a sample belonging to reference quality class $i$ receives the observed rating $j$. The transition matrix is assumed to be independent of sample-specific features, with $T_{i,j}\geq 0$ and $\sum_{j=0}^{K-1}T_{i,j}=1$ for every $i$.

In the ideal case of error-free ratings, $\mathbf{T}$ equals the identity matrix. Off-diagonal transition probabilities therefore characterize potential rating errors under the assumed noise model.

\paragraph{2-NN Score Clusterability.}
To estimate $\mathbf{T}$ without observed reference labels, we adopt the $k$-NN clusterability assumption of \citet{zhu2021clusterability}, following DS$^2$~\citep{pang2024improving}.

Under this assumption, a sample and its $k$ nearest neighbors in the embedding space share the same underlying quality class. As in DS$^2$, we use $k=2$. For an anchor sample $\mathbf{x}_1$ and its two nearest neighbors $\mathbf{x}_2$ and $\mathbf{x}_3$, the assumption implies
\begin{equation}
y_1=y_2=y_3.
\label{eq:app:2nn_clusterability}
\end{equation}

Although the underlying labels are unobserved, the corresponding LLM-rated scores $\tilde{y}_1$, $\tilde{y}_2$, and $\tilde{y}_3$ are available. Their joint rating patterns provide statistical information for estimating the score transition probabilities.

Importantly, this assumption concerns the underlying reference quality labels, rather than requiring neighboring samples to receive identical observed ratings.

\paragraph{Consensus Vectors.}
Following the high-order consensus framework of \citet{zhu2021clusterability}, we use the observed rating patterns among neighboring samples to characterize the latent quality distribution and score transitions.

The framework additionally assumes that, conditioned on their shared underlying quality class, the observed ratings of neighboring samples are independent and follow the same class-conditional transition matrix.

Let
\begin{equation}
\mathbf{p}
=
\left[
\mathbb{P}(y=i)
\right]_{i=0}^{K-1}{}^{\top}
\label{eq:app:quality_class_prior}
\end{equation}
denote the column vector of prior probabilities over the unobserved reference quality classes.

For $g\in\{0,\ldots,K-1\}$, define the cyclically shifted transition matrix
\begin{equation}
\mathbf{T}_g
=
\mathbf{T}\mathbf{H}_g,
\label{eq:app:shifted_transition_matrix}
\end{equation}
where $\mathbf{H}_g$ is the cyclic permutation matrix that shifts the columns of $\mathbf{T}$ to the left by $g$ positions. Using zero-based class indices, the shifted index is defined as $(i+g)_K=(i+g)\bmod K$, such that
\begin{equation}
(\mathbf{T}_g)_{c,i}
=
T_{c,(i+g)_K}.
\label{eq:app:shifted_transition_entry}
\end{equation}

Following the consensus formulation of \citet{zhu2021clusterability} and \citet{pang2024improving}, the first-, second-, and third-order consensus vectors are defined as
\begin{align}
\mathbf{q}^{[1]}
&=
\left[
\mathbb{P}(\tilde{y}_1=i)
\right]_{i=0}^{K-1}{}^{\top}
\nonumber\\
&=
\mathbf{T}^{\top}\mathbf{p},
\label{eq:app:first_order_consensus}
\\[6pt]
\mathbf{q}_{z}^{[2]}
&=
\left[
\mathbb{P}\left(
\tilde{y}_1=i,\,
\tilde{y}_2=(i+z)_K
\right)
\right]_{i=0}^{K-1}{}^{\top}
\nonumber\\
&=
(\mathbf{T}\odot\mathbf{T}_z)^{\top}\mathbf{p},
\label{eq:app:second_order_consensus}
\\[6pt]
(\mathbf{q}_{z,g}^{[3]})_i
&=
\mathbb{P}\left(
\begin{gathered}
\tilde{y}_1=i,\\
\tilde{y}_2=(i+z)_K,\\
\tilde{y}_3=(i+g)_K
\end{gathered}
\right)
\nonumber\\
\mathbf{q}_{z,g}^{[3]}
&=
(\mathbf{T}\odot\mathbf{T}_z
\odot\mathbf{T}_g)^{\top}\mathbf{p},
\label{eq:app:third_order_consensus}
\end{align}
where $\odot$ denotes the element-wise (Hadamard) product, and $i,z,g\in\{0,\ldots,K-1\}$.

These expressions follow from the 2-NN clusterability and conditional independence assumptions. The first-order vector describes the marginal distribution of observed ratings, while the second- and third-order vectors capture joint rating patterns among neighboring samples, including both agreement and disagreement.

The consensus probabilities can be empirically estimated from the frequencies of observed rating configurations within nearest-neighbor groups. These empirical statistics are then used to estimate the unknown transition matrix $\mathbf{T}$ and reference-class distribution $\mathbf{p}$ following the high-order consensus (HOC) estimation procedure of \citet{zhu2021clusterability}, as adopted by DS$^2$~\citep{pang2024improving}. We refer readers to \citet{zhu2021clusterability} for the detailed optimization procedure, theoretical analysis, and identifiability conditions.

\paragraph{Curated Score Generation.}
Following the score-curation procedure of DS$^2$~\citep{pang2024improving}, we use the estimated transition matrix and neighborhood-based rating agreement to identify potentially misrated samples and determine candidate corrected scores.

Specifically, neighborhood-based rating agreement is used to rank samples and suggest candidate corrected scores. The estimated transition matrix and reference-class distribution are used to estimate the number of potentially misrated samples within each observed score class, following the Bayesian error estimation procedure of DS$^2$. The detailed score-correction procedure, including its treatment of imbalanced rating distributions, is described in \citet{pang2024improving}.

We denote the resulting curated overall quality score of sample $n$ by $s_n^{o}\in\{0,\ldots,5\}$. Based on these scores, we partition the source data into low-scoring ($s_n^{o}\in\{0,1,2\}$) and high-scoring ($s_n^{o}\in\{3,4,5\}$) subsets, as described in Section~3.2.

Since the reference quality labels are unobserved, the curated scores represent estimates obtained under the adopted noise and clusterability assumptions, rather than independently verified ground-truth quality annotations.

\subsection{Average Silhouette Score}
\label{appendix:ASS_calculation}

The Average Silhouette Score~\citep{rousseeuw1987silhouettes} evaluates clustering quality by considering both within-cluster cohesion and between-cluster separation.

Consider a partition of embedding vectors into $k$ sub-clusters $C_1,\ldots,C_k$. For a sample $i$ assigned to sub-cluster $C_I$ with $|C_I|>1$, the within-cluster distance is defined as
\begin{equation}
a(i)=
\frac{1}{|C_I|-1}
\sum_{\substack{j\in C_I\\j\neq i}}
d(i,j),
\end{equation}
where $d(i,j)$ denotes the Euclidean distance between the embeddings of samples $i$ and $j$.

The nearest alternative-cluster distance is defined as
\begin{equation}
b(i)=
\min_{J\neq I}
\frac{1}{|C_J|}
\sum_{j\in C_J}d(i,j).
\end{equation}

Here, $a(i)$ measures the average distance from sample $i$ to other members of its assigned cluster, while $b(i)$ measures its smallest average distance to another cluster.

The silhouette coefficient of sample $i$ is then
\begin{equation}
s(i)=
\frac{b(i)-a(i)}
{\max\{a(i),b(i)\}},
\qquad -1\leq s(i)\leq 1.
\end{equation}
For a sample belonging to a singleton cluster, its silhouette coefficient is conventionally set to $s(i)=0$.

A value close to $1$ indicates that a sample is well matched to its assigned cluster and separated from other clusters. Values near $0$ indicate ambiguous assignments, while negative values suggest that a sample may fit another cluster more closely.

For a partition containing $M$ samples, the Average Silhouette Score is
\begin{equation}
\tilde{s}=
\frac{1}{M}\sum_{i=1}^{M}s(i).
\end{equation}

In our representative selection procedure, this score is used to evaluate candidate k-means partitions within each splittable one-hop cluster. The number of sub-clusters $k$ is selected from a predefined candidate range by maximizing the Average Silhouette Score, subject to the validity conditions described in Section~3.3.
\section{Detailed Algorithms and Reconstruction Prompts}

This section provides the detailed procedures for semantic grouping, representative selection, and rule-guided LLM reconstruction. It also presents the predefined reconstruction strategies, validation criteria, failure-guided repair rules, and LLM prompt templates referenced in Sections~3.3 and~3.4.

\subsection{Semantic Grouping and Representative Selection}
\label{appendix:one_hop_cluster_based_representative_selection}

The procedure consists of one-hop cluster formation followed by cluster-based representative selection, as detailed in Algorithm~\ref{alg:representative_selection}.

\paragraph{One-Hop Cluster Formation.}
Given the low-scoring sample set $S_{lq}$, we first randomly shuffle the samples and iteratively select an unassigned sample as the seed of a new one-hop cluster. Each cluster contains its seed and all currently unassigned samples whose cosine similarity with the seed is at least 0.9.

Once a cluster is formed, its members are marked as assigned and excluded from subsequent cluster formation, ensuring that each sample belongs to exactly one cluster. Unlike transitive similarity-based clustering, membership is determined directly by similarity to the selected seed.

\paragraph{Cluster-Based Representative Selection.}
For each one-hop cluster, we retain its initial seed as a representative and process the remaining samples based on their number and sub-clustering validity.

When fewer than three samples remain, all remaining samples are retained. Otherwise, we evaluate candidate k-means partitions and select the number of sub-clusters $k$ that maximizes the Average Silhouette Score, as defined in Appendix~\ref{appendix:ASS_calculation}. The candidate number of sub-clusters ranges from $2$ to $\min(10,|C'|-1)$, where $C'$ denotes the samples remaining after excluding the initial one-hop seed.

If the selected partition contains at least two sub-clusters, each with at least three samples, we select two representatives from each sub-cluster. The first is the sample closest to the sub-cluster centroid in cosine similarity. The second is selected using the MMR-inspired scoring function defined in Section~3.3, with $\alpha=0.2$ to emphasize diversity.

If the sub-clustering conditions are not satisfied, all remaining samples are retained as representatives.

\begin{algorithm*}[t]
\caption{Semantic Grouping and Representative Selection}
\label{alg:representative_selection}

\textbf{Input:} Low-scoring sample set $S_{lq}$ with embeddings $\{\mathbf{x}_n\}$; cosine similarity threshold $\tau=0.9$; MMR weight $\alpha=0.2$; maximum number of sub-clusters $k_{\max}=10$.

\textbf{Output:} Representative sample set $\mathcal{R}$.

\begin{algorithmic}[1]

\State $\mathcal{U}\gets S_{lq}$;
       $\mathcal{G}\gets[\,]$;
       $\mathcal{R}\gets\varnothing$

\Statex \textit{Step 1: One-Hop Cluster Formation}

\For{each sample $u$ in a random permutation of $S_{lq}$}
    \If{$u\in\mathcal{U}$}
        \State $C\gets\{u\}\cup
        \{v\in\mathcal{U}\setminus\{u\}:
        \operatorname{Sim}(\mathbf{x}_u,\mathbf{x}_v)\geq\tau\}$
        \State Append $(u,C)$ to $\mathcal{G}$
        \State $\mathcal{U}\gets\mathcal{U}\setminus C$
    \EndIf
\EndFor

\Statex \textit{Step 2: Cluster-Based Representative Selection}

\For{each $(u,C)\in\mathcal{G}$}
    \State $\mathcal{R}\gets\mathcal{R}\cup\{u\}$
    \Comment{Retain the one-hop seed}
    \State $C'\gets C\setminus\{u\}$

    \If{$|C'|<3$}
        \State $\mathcal{R}\gets\mathcal{R}\cup C'$
    \Else
        \State $\mathcal{K}_C\gets
        \{2,\ldots,\min(k_{\max},|C'|-1)\}$

        \State $(k^*,\{C_1,\ldots,C_{k^*}\})
        \gets\operatorname{BestSilhouetteKMeans}(C',\mathcal{K}_C)$

        \If{$k^*\geq 2$ \textbf{and}
            $\min_j |C_j|\geq 3$}

            \For{each sub-cluster $C_j$}
                \State $\mathbf{r}_{avg,j}\gets
                \frac{1}{|C_j|}
                \sum_{v\in C_j}\mathbf{x}_v$

                \State $r_j^{cent}\gets
                \arg\max_{v\in C_j}
                \operatorname{Sim}(
                \mathbf{x}_v,\mathbf{r}_{avg,j})$

                \State $r_j^{MMR}\gets
                \operatorname{MMRSelect}(
                C_j,\{r_j^{cent}\},\alpha)$

                \State $\mathcal{R}\gets
                \mathcal{R}\cup
                \{r_j^{cent},r_j^{MMR}\}$
            \EndFor

        \Else
            \State $\mathcal{R}\gets\mathcal{R}\cup C'$
            \Comment{Retain all remaining samples}
        \EndIf
    \EndIf
\EndFor

\State \Return $\mathcal{R}$

\end{algorithmic}
\end{algorithm*}

\subsection{Rule-Guided LLM Reconstruction}
\label{appendix:step3}








This subsection provides the implementation details of the rule-guided LLM reconstruction procedure described in Section~3.4. The procedure consists of preparation, relation-dependent strategy selection, and two sequential refinement cycles. LLM-based operators generate and evaluate candidate instruction-response pairs, while a rule-guided controller maps detected failures to predefined repair instructions. The complete procedure is summarized in Algorithm~\ref{alg:rule_guided_reconstruction}.

\subsubsection{Preparation and Structure-Dependence Filtering}
\label{appendix:structure_dependence_filtering}

Before reconstruction, we apply an LLM-based structure-dependence filter to the representative samples obtained from Step 2. The filter determines whether a sample's meaning or correctness depends on preserving its original structural representation.

Specifically, samples containing executable code with syntax or indentation dependencies, tables with essential row-column relationships, format-sensitive mathematical expressions, or other structure-dependent content are excluded when altering their original representations may change their meaning or correctness.

The filter returns a structured decision of either \texttt{KEEP} or \texttt{FILTER}. Samples classified as \texttt{KEEP} remain eligible for reconstruction, whereas those classified as \texttt{FILTER} are excluded.

Importantly, filtering is based on structural dependence rather than task domain. Natural-language mathematical problems and conceptual programming instructions may therefore remain eligible if their semantics do not depend on preserving strict syntax or layout.

The subsequent reconstruction procedure operates on eligible source samples, denoted by $C_A$ and $C_B$. These samples may be drawn from the same semantic cluster for intra-cluster reconstruction or from different clusters for inter-cluster reconstruction. In the intra-cluster setting, the procedure can be applied progressively to incorporate additional representative samples.

\subsubsection{Domain Analysis and Strategy Selection}
\label{appendix:domain_analysis_strategy_selection}

For each eligible source pair $(C_A,C_B)$, Domain Analysis (DA) uses an LLM to extract salient concepts, key terms, and semantic relationships from the two sources. Given the domain-analysis prompt template $P_{\mathrm{DA}}$, the operator produces a structured report containing the extracted source information and a predicted relationship category.

The relationship is classified into one of three categories: \textit{same-domain}, \textit{related-domain}, or \textit{unrelated-domain}. These categories characterize the semantic compatibility of the input sources and determine which reconstruction strategies are applicable.

Strategy Selection then assigns three predefined strategies associated with the predicted relationship category. Across the three categories, the framework employs nine predefined strategies inspired by writing-studies literature~\citep{nelson2023discourse,knobel2017remix,bazerman2003intertextuality}.

Let $\mathcal{S}(d)$ denote the strategy set associated with relationship category $d$, where $|\mathcal{S}(d)|=3$. Rather than allowing unconstrained source combination, each strategy specifies a distinct way of organizing or integrating the source information.

\begin{figure*}[t]
    \centering
    \includegraphics[
        width=\textwidth,
        height=\textheight,
        keepaspectratio
    ]{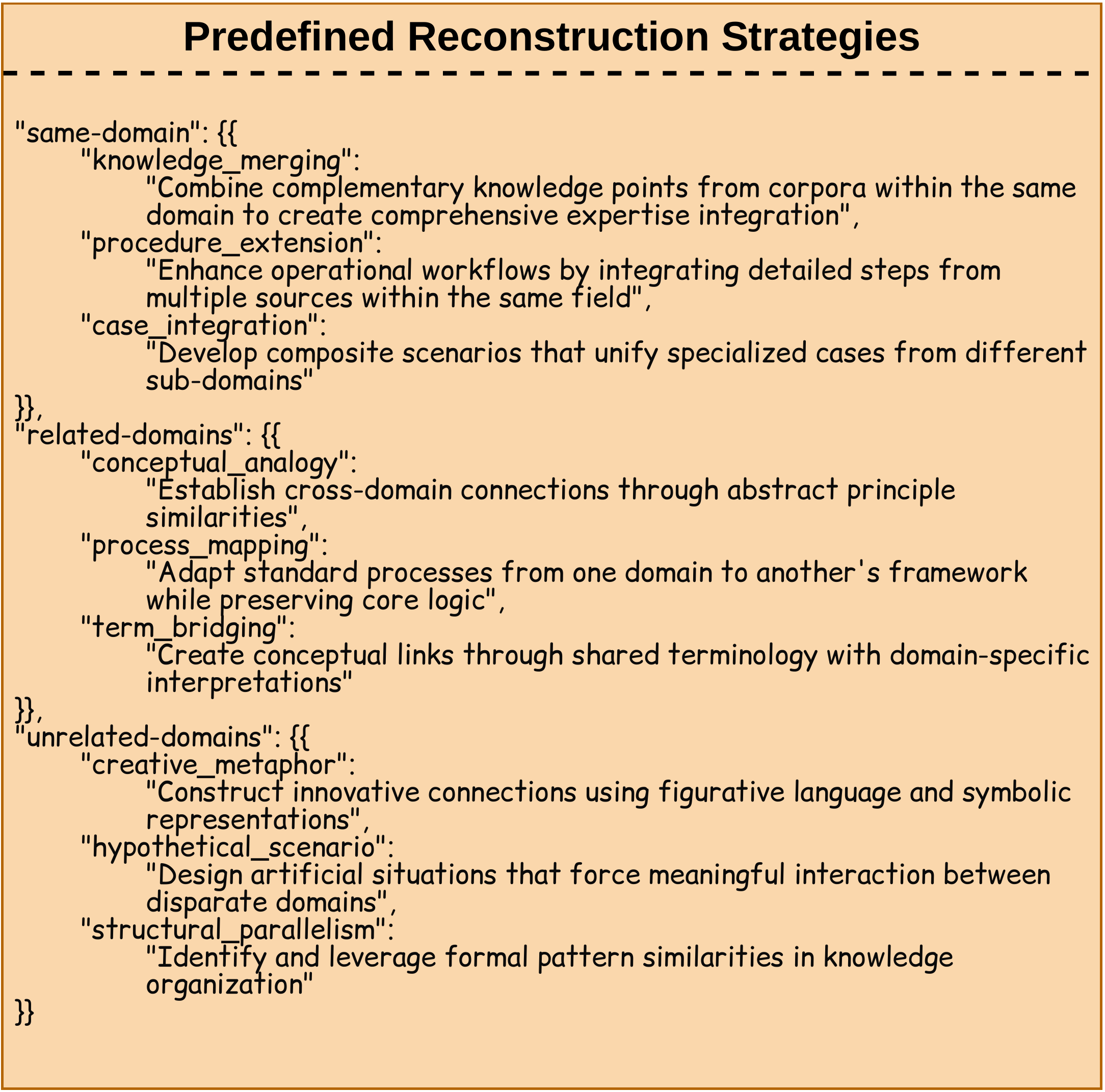}
    \caption{Nine predefined reconstruction strategies grouped by the semantic relationship between source samples. Each relationship category is associated with three strategies that guide subsequent LLM-based reconstruction.}
    \label{fig:app:reconstruction_strategies}
\end{figure*}

As illustrated in Figure~\ref{fig:app:reconstruction_strategies}, the Merged Corpus Generation (MCG) operator generates one candidate instruction-response pair for each selected strategy. The resulting candidates are subsequently validated and refined independently. This separation allows different reconstruction strategies to be explored without conflating their intermediate validation and repair histories.

\subsubsection{Cycle 1: Information Completeness}
\label{appendix:cycle1_information_completeness}

The first refinement cycle constructs an instruction-response candidate with sufficient source-information coverage, a well-formed question, and the context needed to answer it.

For each selected strategy $s\in\mathcal{S}(d)$, MCG generates an initial candidate from the source samples and the strategy-specific instructions. The candidate is then evaluated by Information Completeness Detection (ICD).

ICD produces a structured validation report covering three principal aspects.

\paragraph{Key-Term Coverage.}
ICD examines whether the salient concepts and key terms extracted from the two source samples are meaningfully incorporated into the candidate. The report distinguishes incorporated information from missing or insufficiently integrated information. Mere occurrence of a term in the generated text does not necessarily establish meaningful integration.

\paragraph{Question Quality and Context Completeness.}
ICD determines whether the candidate contains a well-formed and relevant question, whether its intended task is sufficiently specified, and whether the supporting context provides the information necessary to answer the question. Missing contextual information or an underspecified question is identified as a validation failure.

\paragraph{Answer Completeness.}
ICD also checks whether the candidate contains a direct answer to the generated question and whether the answer requires regeneration.

When one or more validation criteria remain unsatisfied, the Rule-Guided Prompt Updater (RPU) selects a predefined MCG repair prompt template according to the combination of detected failures. The resulting instruction updates the MCG prompt for the next generation attempt.

The revised candidate is evaluated again by ICD. This process continues until all criteria are satisfied or the generation limit is reached. We allow up to three candidate generations per cycle, including the initial generation and at most two regeneration attempts.

If multiple candidates have been generated without satisfying all criteria, the candidate with the fewest unmet validation conditions is retained. Ties are resolved randomly. The selected Cycle 1 candidate is denoted by $C_{\mathrm{c1}}$.

\subsubsection{Cycle 2: Final Answer Refinement}
\label{appendix:cycle2_final_answer_refinement}

The second cycle refines the response of the selected Cycle 1 candidate while preserving its instruction and supporting context.

Final Answer Check (FAC) evaluates whether the response directly addresses the question and identifies unnecessary, irrelevant, or redundant information. Its structured validation report specifies whether further answer revision is required.

If a failure is detected, RPU selects the corresponding predefined FAU prompt template based on the combination of validation findings in the FAC report. FAU then revises the response using the selected template, without modifying the instruction or supporting context established in Cycle 1.

The revised response is re-evaluated by FAC. As in Cycle 1, refinement stops when all applicable validation conditions are satisfied or the limit of three candidate generations, including the initial generation, is reached.

If the iteration limit is reached before all conditions are satisfied, the candidate with the fewest unmet conditions is selected, with random tie-breaking.

The final reconstructed instruction-response pair combines the preserved instruction and context from Cycle 1 with the selected response from Cycle 2. This procedure is applied independently to candidates generated under different reconstruction strategies.

\subsubsection{Validation Criteria and Failure-Guided Repair Rules}
\label{appendix:validation_repair_rules}

The rule-guided reconstruction procedure uses structured validation reports to determine whether a generated candidate requires further refinement. Rather than generating repair instructions dynamically, the Rule-Guided Prompt Updater (RPU) selects from a set of predefined prompt templates according to the combination of detected validation failures.

\paragraph{Cycle 1: ICD-Based Prompt Selection.}
Information Completeness Detection (ICD) evaluates generated candidates and returns a structured report containing three principal components: \texttt{term\_coverage}, \texttt{question\_analysis}, and \texttt{answer\_quality}.

The term-coverage component identifies source concepts that are incorporated or missing. The question-analysis component checks whether a suitable question exists and whether additional context is required. The answer-quality component determines whether the response adequately addresses the question.

RPU examines these validation results jointly to determine the applicable MCG prompt template. Different combinations of unmet conditions correspond to different predefined templates, including cases involving missing source concepts, insufficient contextual information, inadequate answers, or multiple simultaneous deficiencies.

Each selected template specifies the corresponding regeneration task, required input information, content-preservation constraints, and expected output structure. MCG then uses this template to revise the current candidate, which is subsequently re-evaluated by ICD.

\paragraph{Cycle 2: FAC-Based Prompt Selection.}
Final Answer Check (FAC) evaluates the response for the presence of a direct answer and for unnecessary, irrelevant, or redundant content.

Based on the FAC report, RPU selects the corresponding predefined Final Answer Update (FAU) prompt template. These templates address answer-specific deficiencies while preserving the instruction and supporting context established in Cycle 1.

\paragraph{Rule-Guided Control and Termination.}
The controller performs conditional template selection based on the structured validation results. It does not compute gradients, update model parameters, or learn a globally optimized prompt.

In each cycle, the generated candidate is repeatedly evaluated and revised until all applicable validation conditions are satisfied or the maximum number of regeneration attempts is reached. If no candidate satisfies all conditions, the candidate with the fewest unmet conditions is selected, with random tie-breaking.

The complete ICD, FAC, and conditional MCG/FAU prompt templates are provided in Appendix~\ref{appendix:reconstruction_prompt_templates}.

\subsubsection{Complete Reconstruction Algorithm}
\label{appendix:complete_reconstruction_algorithm}

Algorithm~\ref{alg:rule_guided_reconstruction} presents the complete procedure for reconstructing candidates from an eligible source pair. The DA, MCG, ICD, FAC, and FAU operators invoke the external LLM using task-specific prompt templates, whereas Strategy Selection, RPU, validation-condition checks, and candidate selection implement the predefined control logic.

The algorithm allows up to three candidate generations per cycle, including the initial generation and at most two regeneration attempts. For candidate selection, $\operatorname{SelectBest}$ returns a candidate satisfying all validation conditions when available; otherwise, it selects the candidate with the fewest unmet conditions, with random tie-breaking.

Here, $\operatorname{RPU}_{\mathrm{MCG}}$ and $\operatorname{RPU}_{\mathrm{FAU}}$ denote rule-based selection of predefined MCG and FAU prompt templates according to the corresponding ICD and FAC validation reports, respectively.

\begin{algorithm*}[t]
\caption{Rule-Guided LLM Reconstruction}
\label{alg:rule_guided_reconstruction}

\textbf{Input:} Eligible source pair $(C_A,C_B)$; task-specific prompt templates $\mathcal{P}$; predefined strategy and repair rules; maximum candidate generations per cycle $B=3$.

\textbf{Output:} Reconstructed candidate set $\mathcal{O}$.

\begin{algorithmic}[1]

\State $\mathcal{O}\gets\varnothing$

\Statex \textit{Preparation: Domain Analysis and Strategy Selection}

\State $\mathcal{D}\gets
\operatorname{DA}(C_A,C_B;P_{\mathrm{DA}})$

\State $d\gets\operatorname{Relationship}(\mathcal{D})$

\State $\mathcal{S}\gets
\operatorname{StrategySelection}(d)$

\State $\{C_s^0\}_{s\in\mathcal{S}}
\gets\operatorname{MCG}
(C_A,C_B,\mathcal{D},\mathcal{S};P_{\mathrm{MCG}})$

\For{each $s\in\mathcal{S}$}

    \Statex \textit{Cycle 1: Information Completeness}

    \State $V_{\mathrm{ICD}}^0
    \gets\operatorname{ICD}
    (C_s^0;P_{\mathrm{ICD}})$

    \State $\mathcal{H}_1
    \gets\{(C_s^0,V_{\mathrm{ICD}}^0)\}$

    \For{$t=1$ to $B-1$}
        \If{$\operatorname{Pass}(V_{\mathrm{ICD}}^{t-1})$}
            \State \textbf{break}
        \EndIf

        \State $P_{\mathrm{MCG}}^t
        \gets\operatorname{RPU}_{\mathrm{MCG}}
        (V_{\mathrm{ICD}}^{t-1})$

        \State $C_s^t
        \gets\operatorname{MCG}
        (C_A,C_B,\mathcal{D},s,C_s^{t-1};
        P_{\mathrm{MCG}}^t)$

        \State $V_{\mathrm{ICD}}^t
        \gets\operatorname{ICD}
        (C_s^t;P_{\mathrm{ICD}})$

        \State $\mathcal{H}_1
        \gets\mathcal{H}_1
        \cup\{(C_s^t,V_{\mathrm{ICD}}^t)\}$
    \EndFor

    \State $C_{\mathrm{c1}}
    \gets\operatorname{SelectBest}(\mathcal{H}_1)$

    \Statex \textit{Cycle 2: Final Answer Refinement}

    \State $A_s^0\gets C_{\mathrm{c1}}$

    \State $V_{\mathrm{FAC}}^0
    \gets\operatorname{FAC}
    (A_s^0;P_{\mathrm{FAC}})$

    \State $\mathcal{H}_2
    \gets\{(A_s^0,V_{\mathrm{FAC}}^0)\}$

    \For{$t=1$ to $B-1$}
        \If{$\operatorname{Pass}(V_{\mathrm{FAC}}^{t-1})$}
            \State \textbf{break}
        \EndIf

        \State $P_{\mathrm{FAU}}^t
        \gets\operatorname{RPU}_{\mathrm{FAU}}
        (V_{\mathrm{FAC}}^{t-1})$

        \State $A_s^t
        \gets\operatorname{FAU}
        (A_s^{t-1};P_{\mathrm{FAU}}^t)$
        \Comment{Revise response only}

        \State $V_{\mathrm{FAC}}^t
        \gets\operatorname{FAC}
        (A_s^t;P_{\mathrm{FAC}})$

        \State $\mathcal{H}_2
        \gets\mathcal{H}_2
        \cup\{(A_s^t,V_{\mathrm{FAC}}^t)\}$
    \EndFor

    \State $C_{\mathrm{final}}
    \gets\operatorname{SelectBest}(\mathcal{H}_2)$

    \State $\mathcal{O}
    \gets\mathcal{O}\cup\{C_{\mathrm{final}}\}$

\EndFor

\State \Return $\mathcal{O}$

\end{algorithmic}
\end{algorithm*}

\subsubsection{LLM-Driven Prompt Templates}
\label{appendix:reconstruction_prompt_templates}

This subsection provides the complete prompt templates used by the LLM-based operators in the reconstruction pipeline, including structure-dependence filtering, Domain Analysis (DA), Merged Corpus Generation (MCG), Information Completeness Detection (ICD), Final Answer Check (FAC), and Final Answer Update (FAU).

The prompts specify the corresponding task instructions, required input information, validation criteria, and expected output structures. For failure-guided refinement, the Rule-Guided Prompt Updater (RPU) selects among predefined MCG and FAU templates based on the structured ICD and FAC reports, respectively, as described in Appendix~\ref{appendix:validation_repair_rules}.

\begin{figure*}[p]
    \centering
    \includegraphics[
        width=\textwidth,
        height=\textheight,
        keepaspectratio
    ]{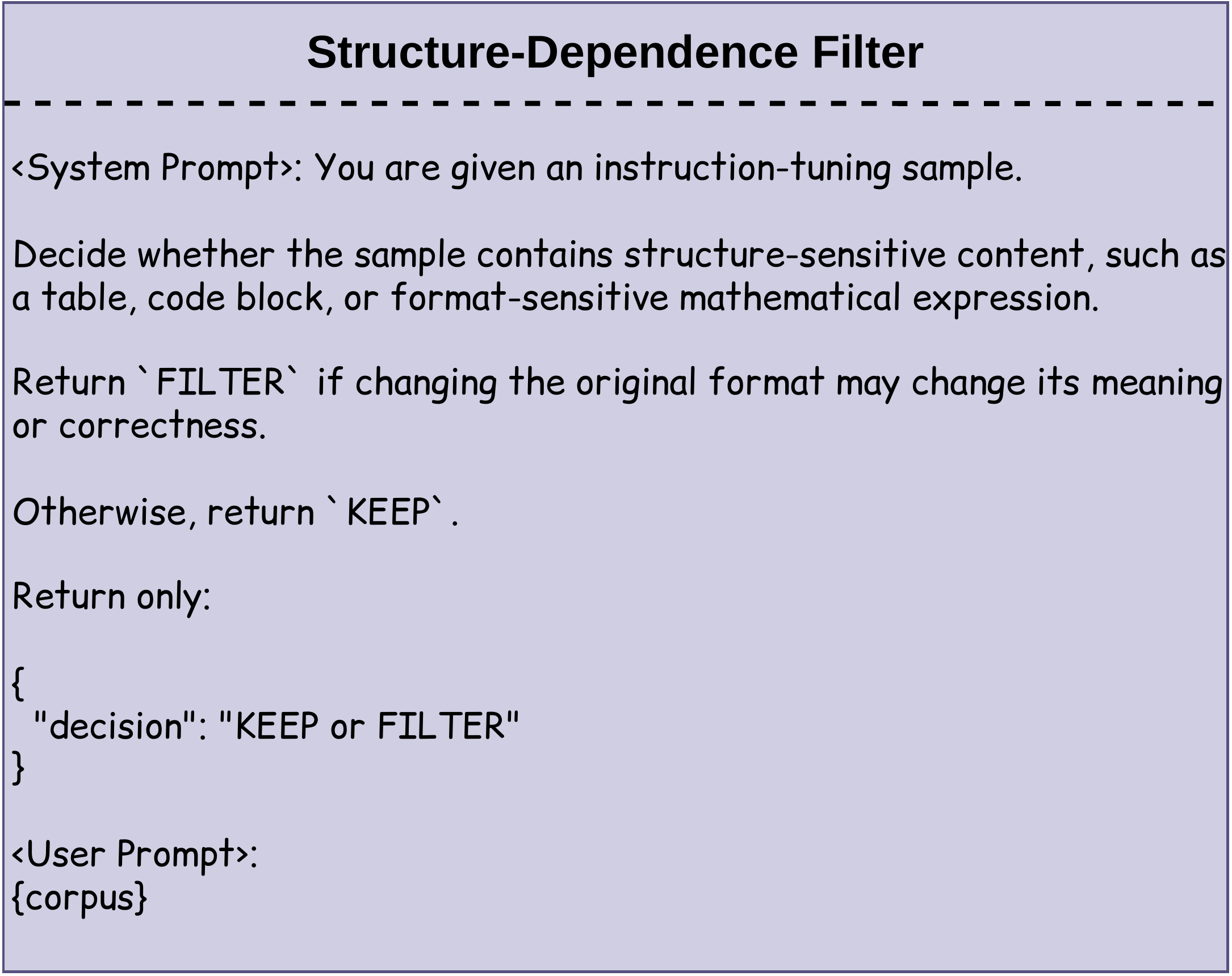}
    \caption{Structure-dependence filtering prompt used to identify source samples requiring strict preservation of their original format.}
    \label{fig:app:filter_prompt}
\end{figure*}

\paragraph{Structure-Dependence Filtering Prompt.}
The structure-dependence filtering prompt is shown in Figure~\ref{fig:app:filter_prompt}. It classifies each source sample as \texttt{KEEP} or \texttt{FILTER} according to whether preserving its original structural representation is necessary to maintain its meaning or correctness.

\begin{figure*}[p]
    \centering
    \includegraphics[
        width=\textwidth,
        height=\textheight,
        keepaspectratio
    ]{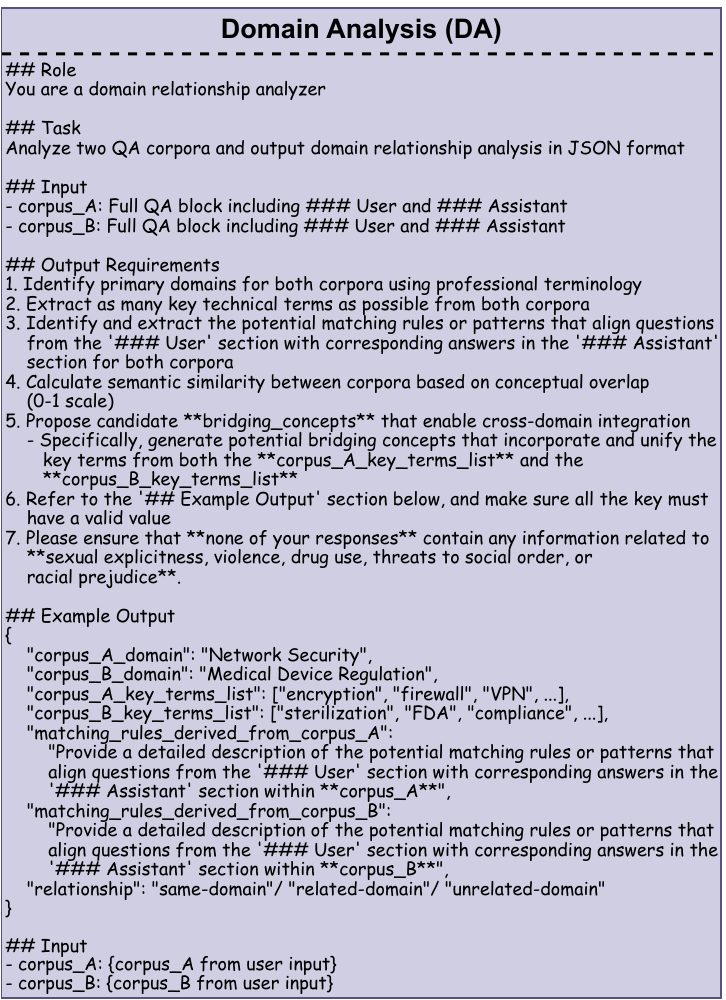}
    \caption{Prompt template for Domain Analysis (DA).}
    \label{fig:app:da_prompt}
\end{figure*}

\paragraph{Domain Analysis Prompt.}
Figure~\ref{fig:app:da_prompt} presents the DA prompt, which extracts domain information, salient source concepts, and semantic matching patterns, and classifies the relationship between the two source samples.

\begin{figure*}[p]
    \centering
    \includegraphics[
        width=\textwidth,
        height=\textheight,
        keepaspectratio
    ]{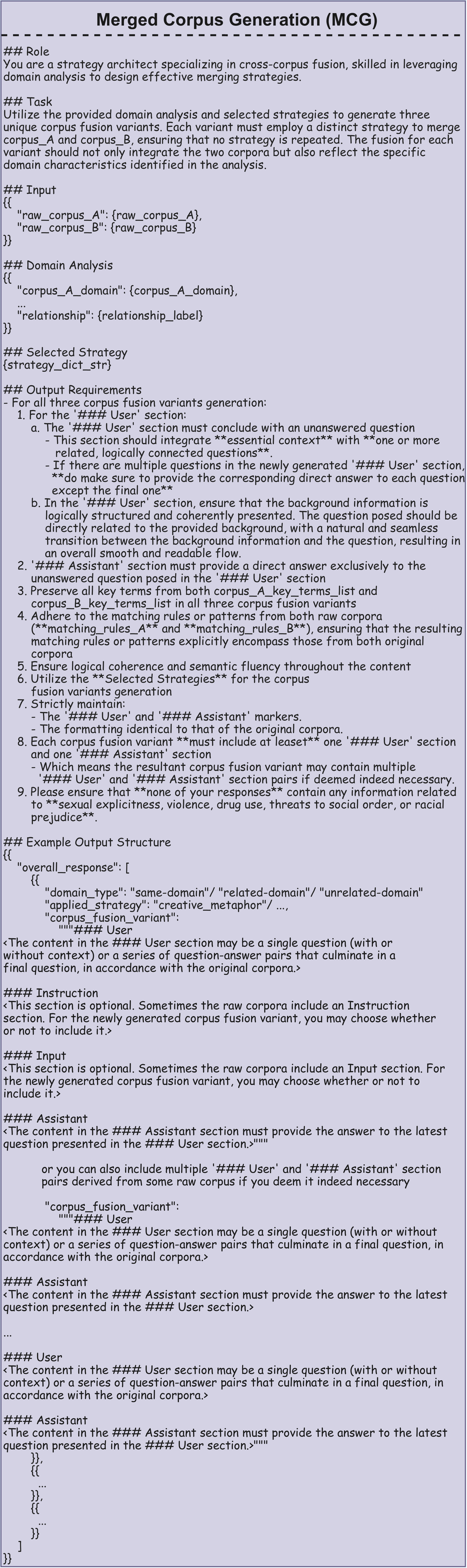}
    \caption{Initial prompt template for Merged Corpus Generation (MCG).}
    \label{fig:app:mcg_initial_prompt}
\end{figure*}

\paragraph{Initial Merged Corpus Generation Prompt.}
The initial MCG prompt is provided in Figure~\ref{fig:app:mcg_initial_prompt}. Given the source samples, domain-analysis report, and selected reconstruction strategies, it generates three strategy-specific candidate instruction-response pairs.

\begin{figure*}[p]
    \centering
    \includegraphics[
        width=\textwidth,
        height=\textheight,
        keepaspectratio
    ]{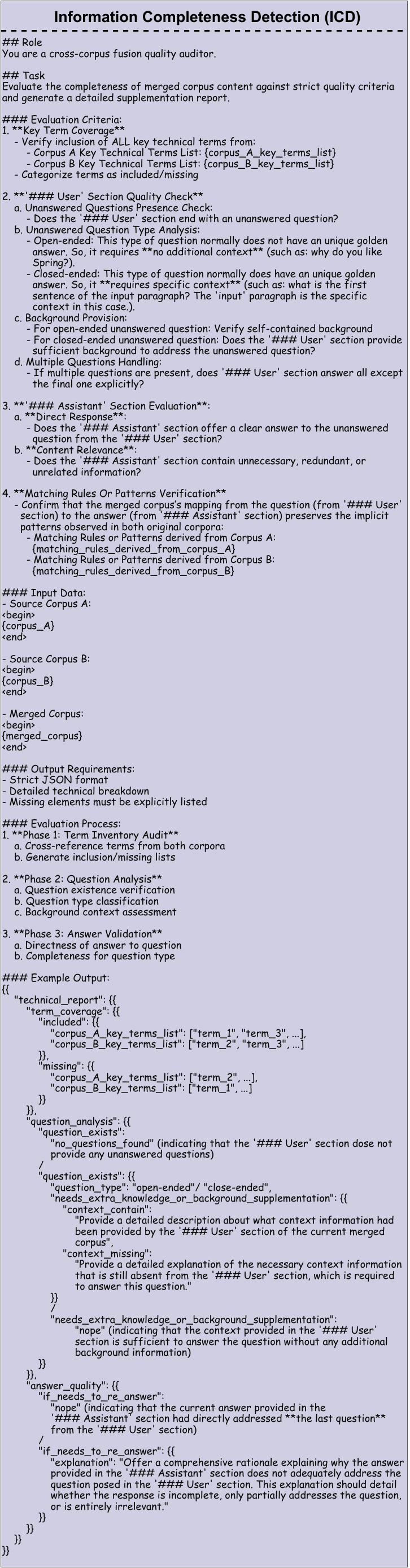}
    \caption{Prompt template for Information Completeness Detection (ICD).}
    \label{fig:app:icd_prompt}
\end{figure*}

\paragraph{Information Completeness Detection Prompt.}
Figure~\ref{fig:app:icd_prompt} presents the ICD prompt used in Cycle 1. It produces a structured validation report covering source-term incorporation, question quality, contextual completeness, and answer completeness. The resulting report provides the conditions used by RPU to select the subsequent MCG prompt template.

\begin{figure*}[p]
    \centering
    \includegraphics[
        width=\textwidth,
        height=\textheight,
        keepaspectratio
    ]{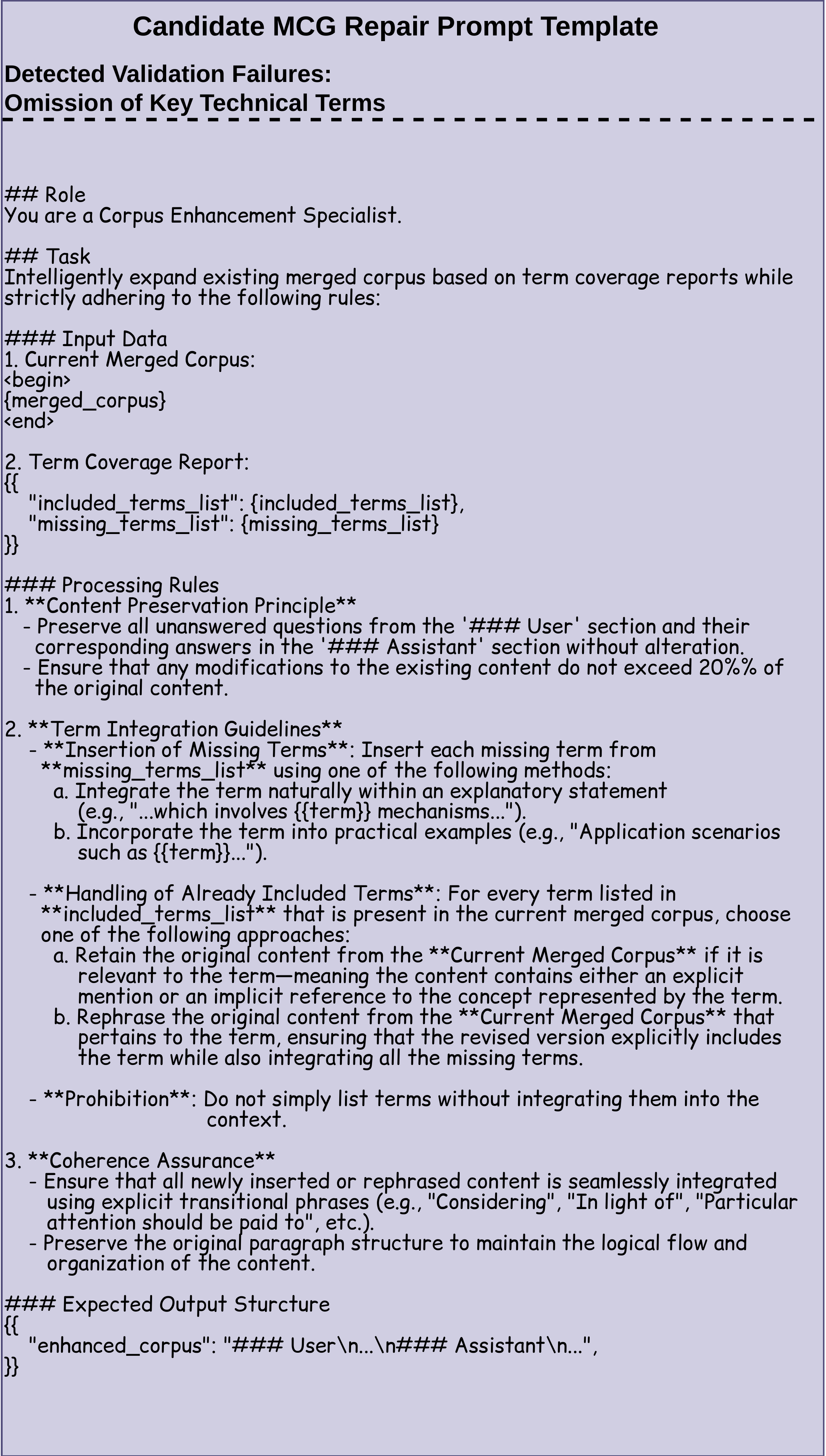}
    \caption{Conditional MCG repair prompt template for missing key technical terms.}
    \label{fig:app:mcg_repair_terms}
\end{figure*}

\begin{figure*}[p]
    \centering
    \includegraphics[
        width=\textwidth,
        height=\textheight,
        keepaspectratio
    ]{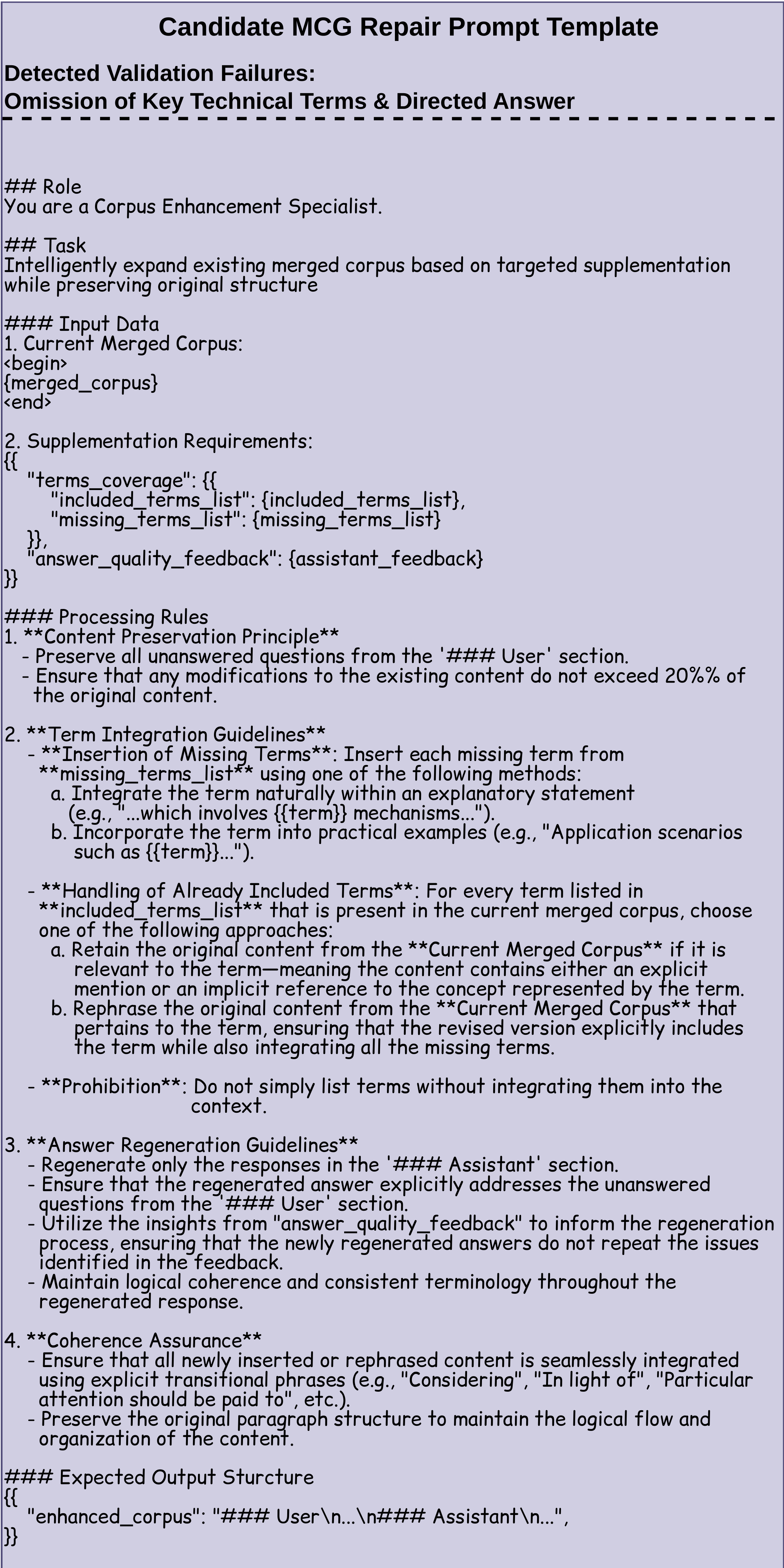}
    \caption{Conditional MCG repair prompt template for missing key technical terms and a direct answer.}
    \label{fig:app:mcg_repair_terms_answer}
\end{figure*}

\begin{figure*}[p]
    \centering
    \includegraphics[
        width=\textwidth,
        height=\textheight,
        keepaspectratio
    ]{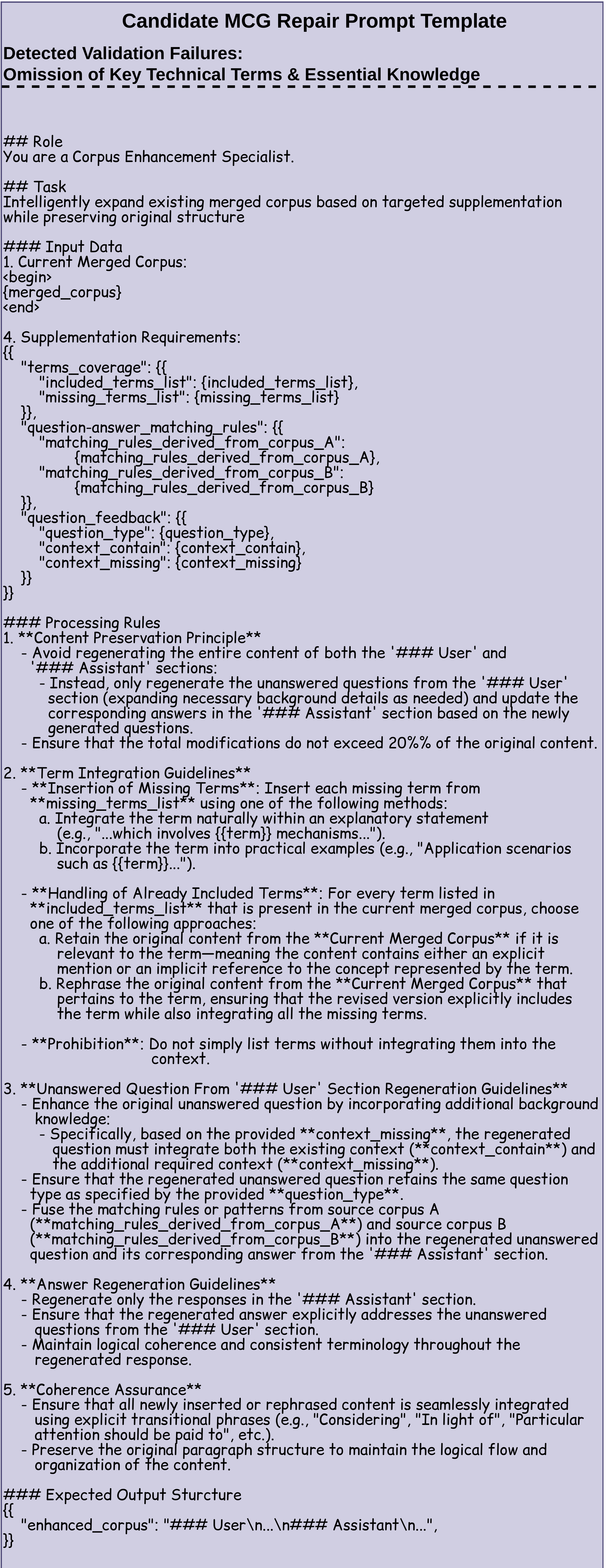}
    \caption{Conditional MCG repair prompt template for missing key technical terms and essential contextual knowledge.}
    \label{fig:app:mcg_repair_terms_context}
\end{figure*}

\begin{figure*}[p]
    \centering
    \includegraphics[
        width=\textwidth,
        height=\textheight,
        keepaspectratio
    ]{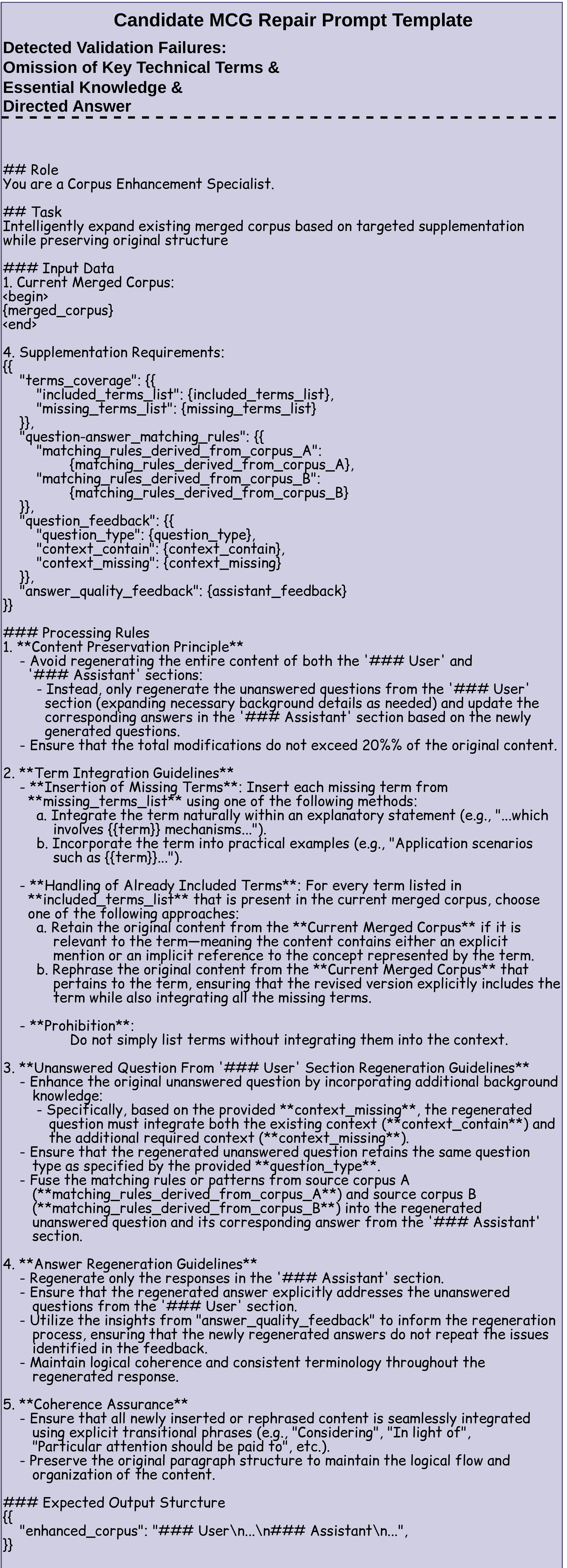}
    \caption{Conditional MCG repair prompt template for missing key technical terms, essential contextual knowledge, and a direct answer.}
    \label{fig:app:mcg_repair_terms_context_answer}
\end{figure*}

\begin{figure*}[p]
    \centering
    \includegraphics[
        width=\textwidth,
        height=\textheight,
        keepaspectratio
    ]{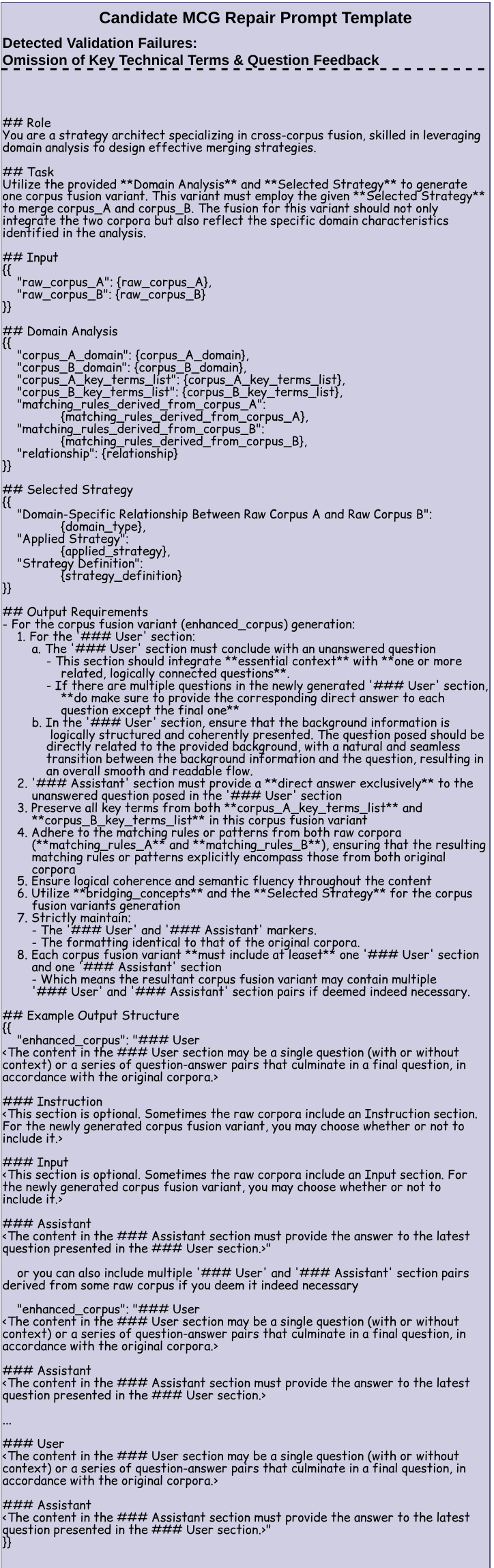}
    \caption{Conditional MCG repair prompt template for missing key technical terms and question-related deficiencies.}
    \label{fig:app:mcg_repair_terms_question}
\end{figure*}

\begin{figure*}[p]
    \centering
    \includegraphics[
        width=\textwidth,
        height=\textheight,
        keepaspectratio
    ]{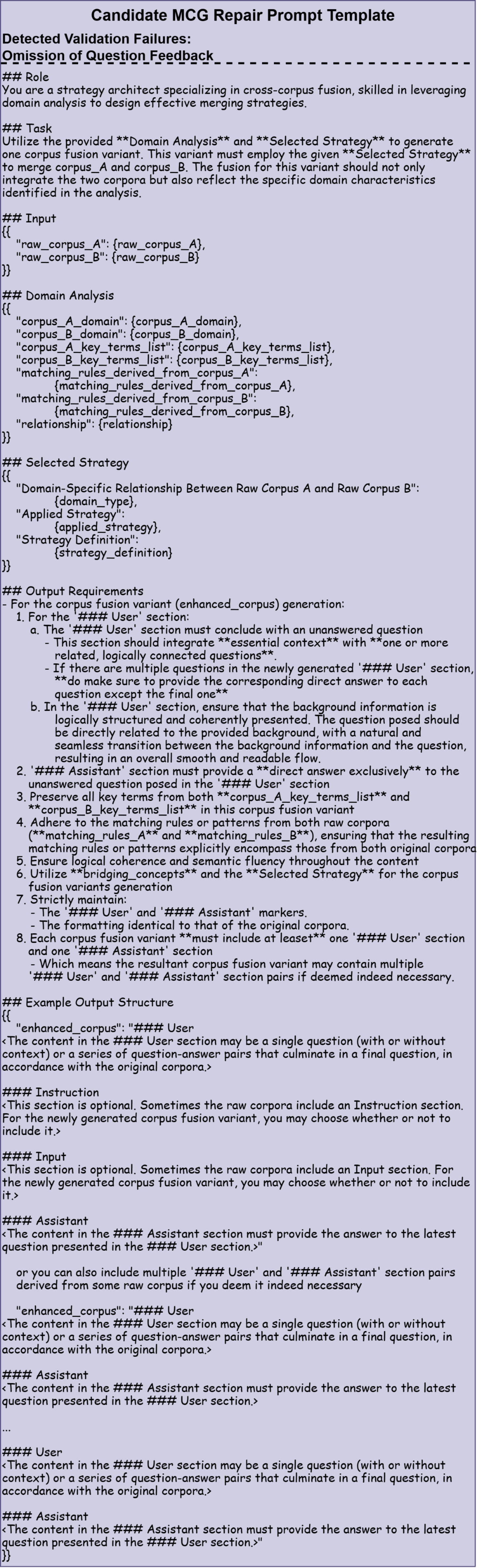}
    \caption{Conditional MCG repair prompt template for question-related deficiencies.}
    \label{fig:app:mcg_repair_question}
\end{figure*}

\begin{figure*}[p]
    \centering
    \includegraphics[
        width=\textwidth,
        height=\textheight,
        keepaspectratio
    ]{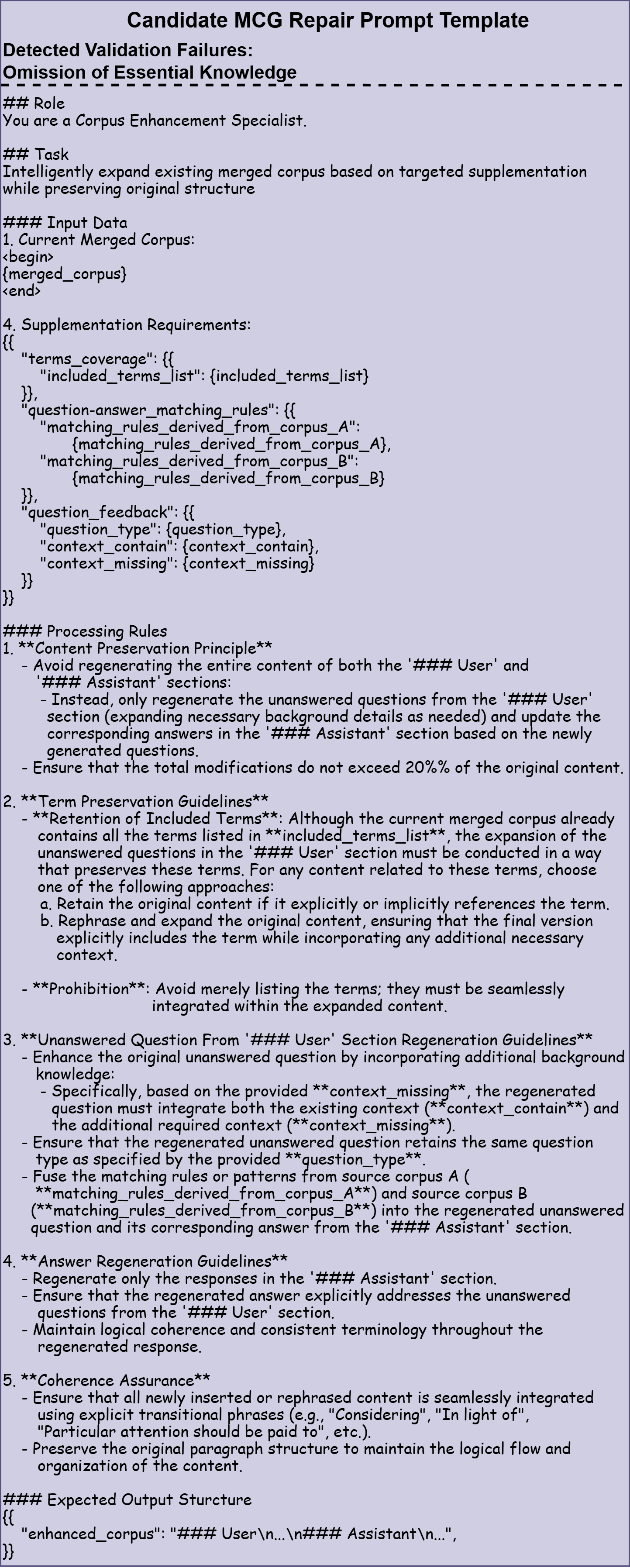}
    \caption{Conditional MCG repair prompt template for missing essential contextual knowledge.}
    \label{fig:app:mcg_repair_context}
\end{figure*}

\begin{figure*}[p]
    \centering
    \includegraphics[
        width=\textwidth,
        height=\textheight,
        keepaspectratio
    ]{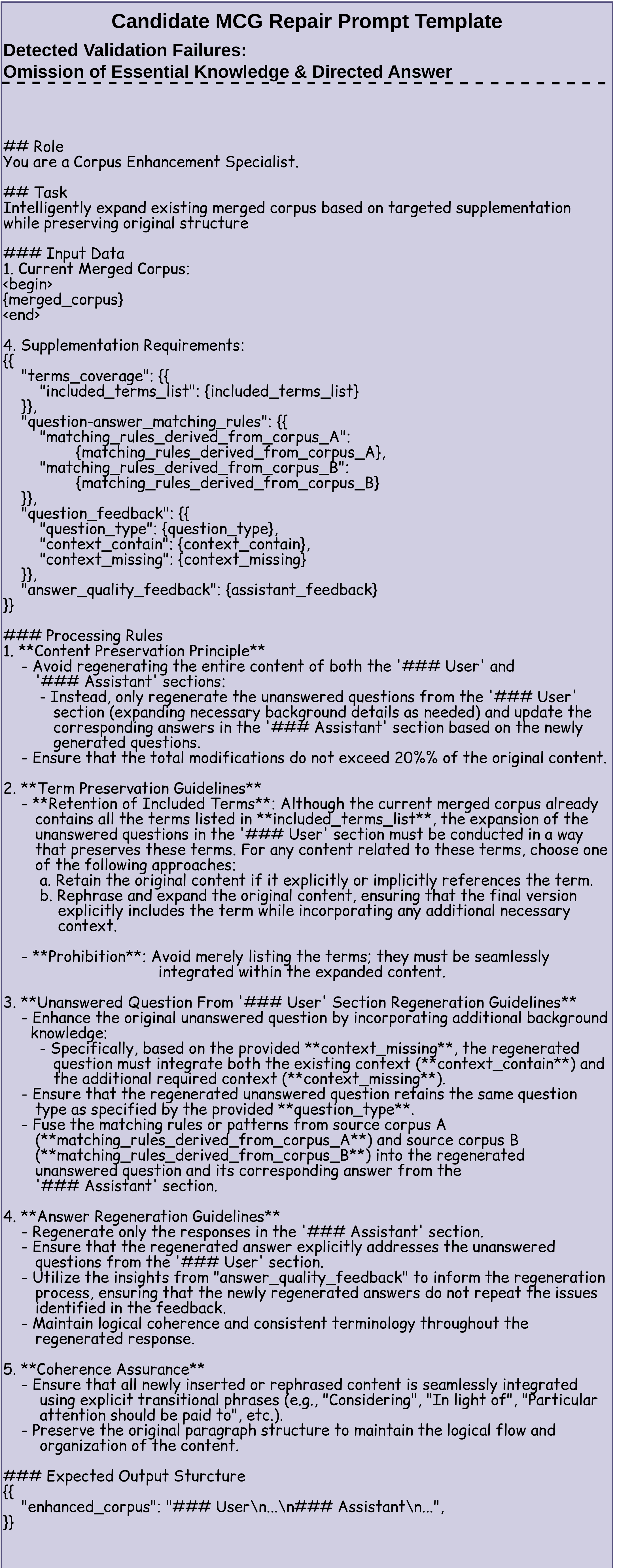}
    \caption{Conditional MCG repair prompt template for missing essential contextual knowledge and a direct answer.}
    \label{fig:app:mcg_repair_context_answer}
\end{figure*}

\begin{figure*}[p]
    \centering
    \includegraphics[
        width=\textwidth,
        height=\textheight,
        keepaspectratio
    ]{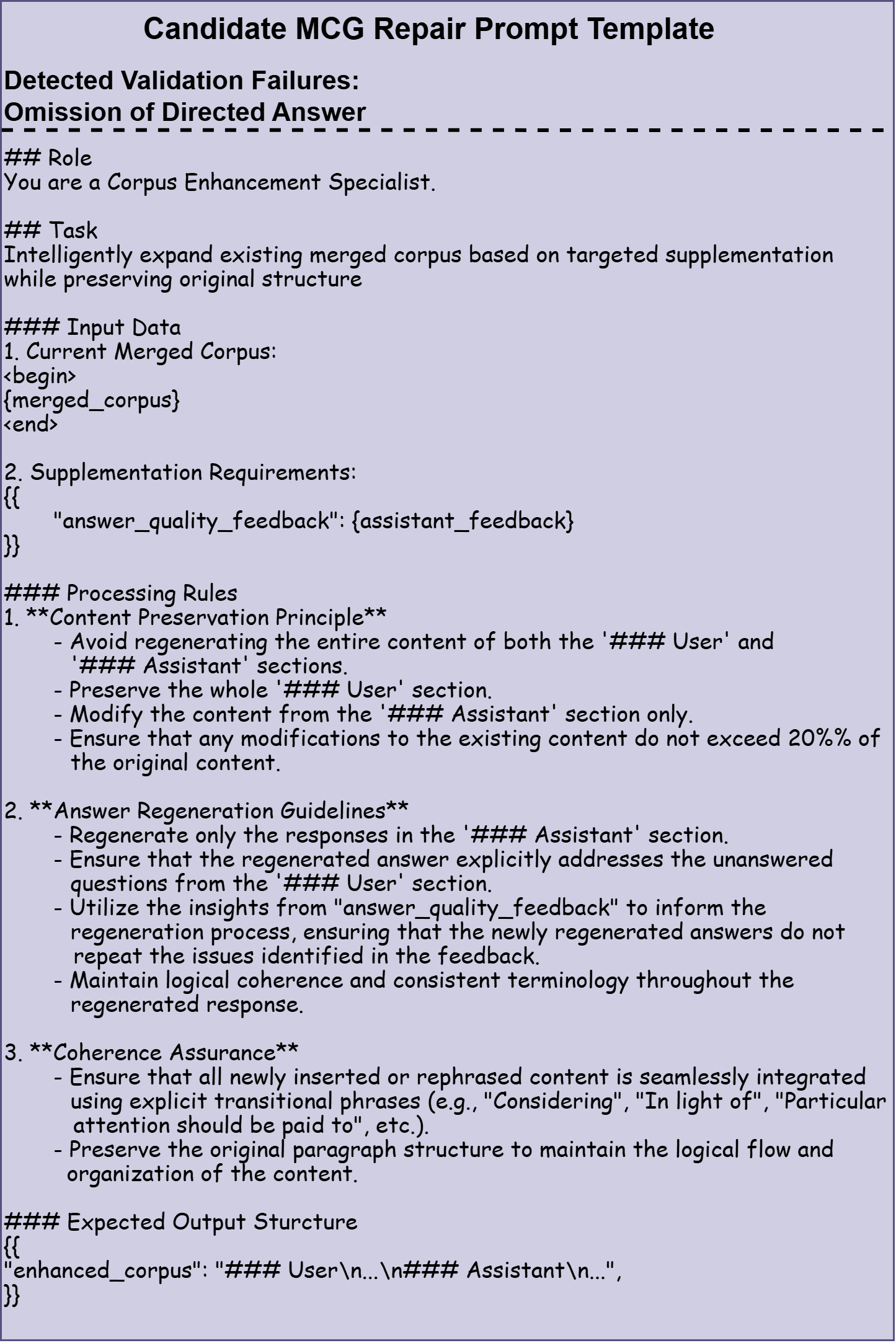}
    \caption{Conditional MCG repair prompt template for a missing direct answer.}
    \label{fig:app:mcg_repair_answer}
\end{figure*}

\paragraph{Conditional MCG Repair Prompt Templates.}
When ICD identifies unmet validation conditions, RPU selects a predefined MCG repair prompt template according to the combination of detected failures. The nine conditional templates are provided in Figures~\ref{fig:app:mcg_repair_terms}--\ref{fig:app:mcg_repair_answer}.

These templates cover different combinations of missing source concepts, insufficient contextual information, question-related deficiencies, and inadequate answers. Each template specifies the relevant feedback, content-preservation requirements, and regeneration instructions. The selected template is used to revise the current candidate, which is subsequently re-evaluated by ICD.

\begin{figure*}[p]
    \centering
    \includegraphics[
        width=\textwidth,
        height=\textheight,
        keepaspectratio
    ]{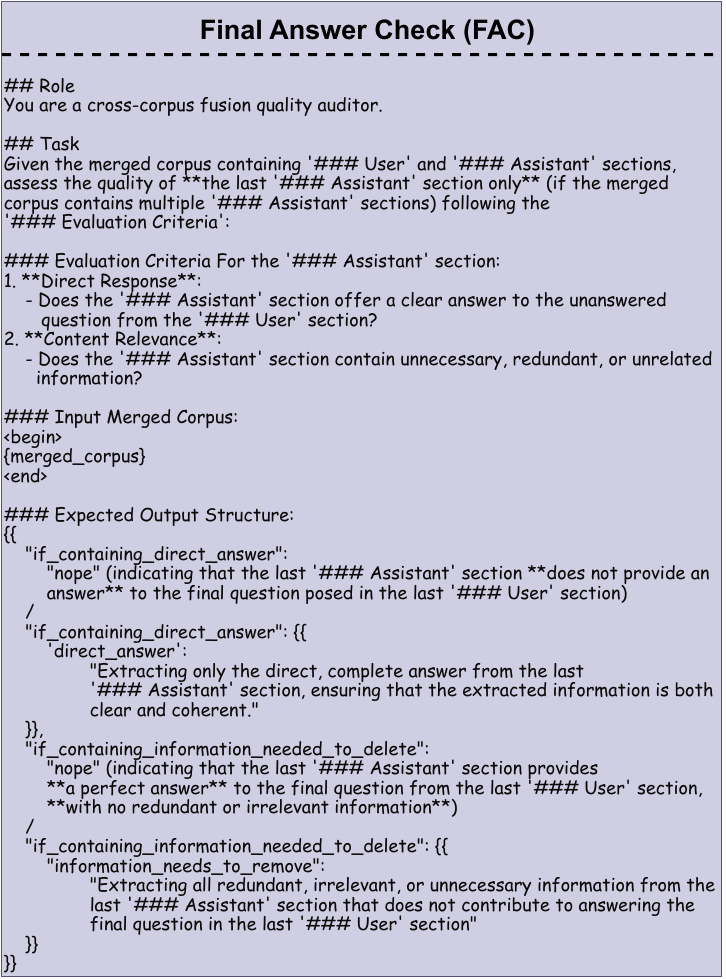}
    \caption{Prompt template for Final Answer Check (FAC).}
    \label{fig:app:fac_prompt}
\end{figure*}

\paragraph{Final Answer Check Prompt.}
The FAC prompt used in Cycle 2 is shown in Figure~\ref{fig:app:fac_prompt}. It evaluates whether the response directly addresses the final question and identifies unnecessary, irrelevant, or redundant information. The structured FAC report determines which predefined FAU prompt template is selected.

\begin{figure*}[p]
    \centering
    \includegraphics[
        width=\textwidth,
        height=\textheight,
        keepaspectratio
    ]{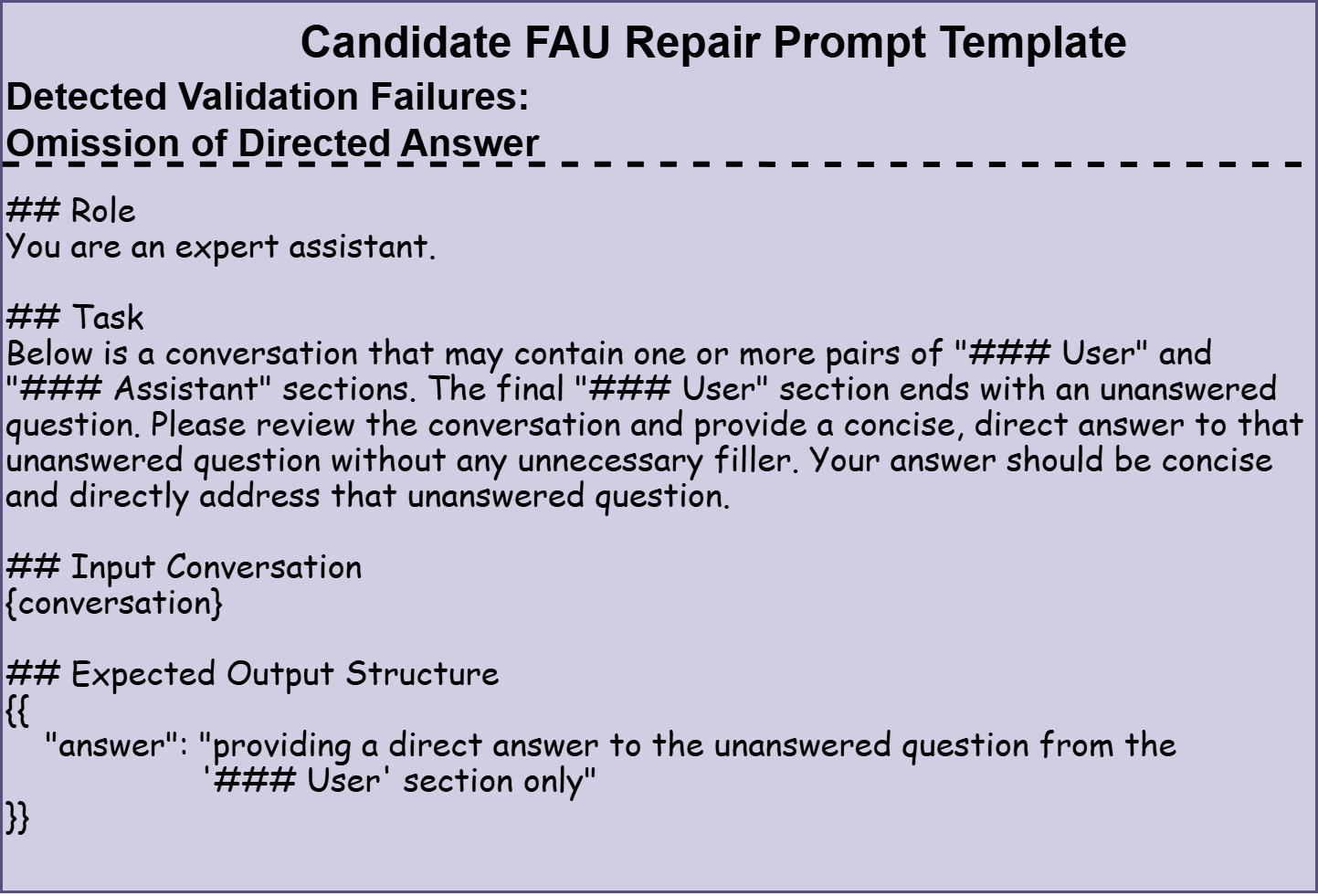}
    \caption{Conditional FAU repair prompt template for generating a missing direct answer.}
    \label{fig:app:fau_repair_answer}
\end{figure*}

\begin{figure*}[p]
    \centering
    \includegraphics[
        width=\textwidth,
        height=\textheight,
        keepaspectratio
    ]{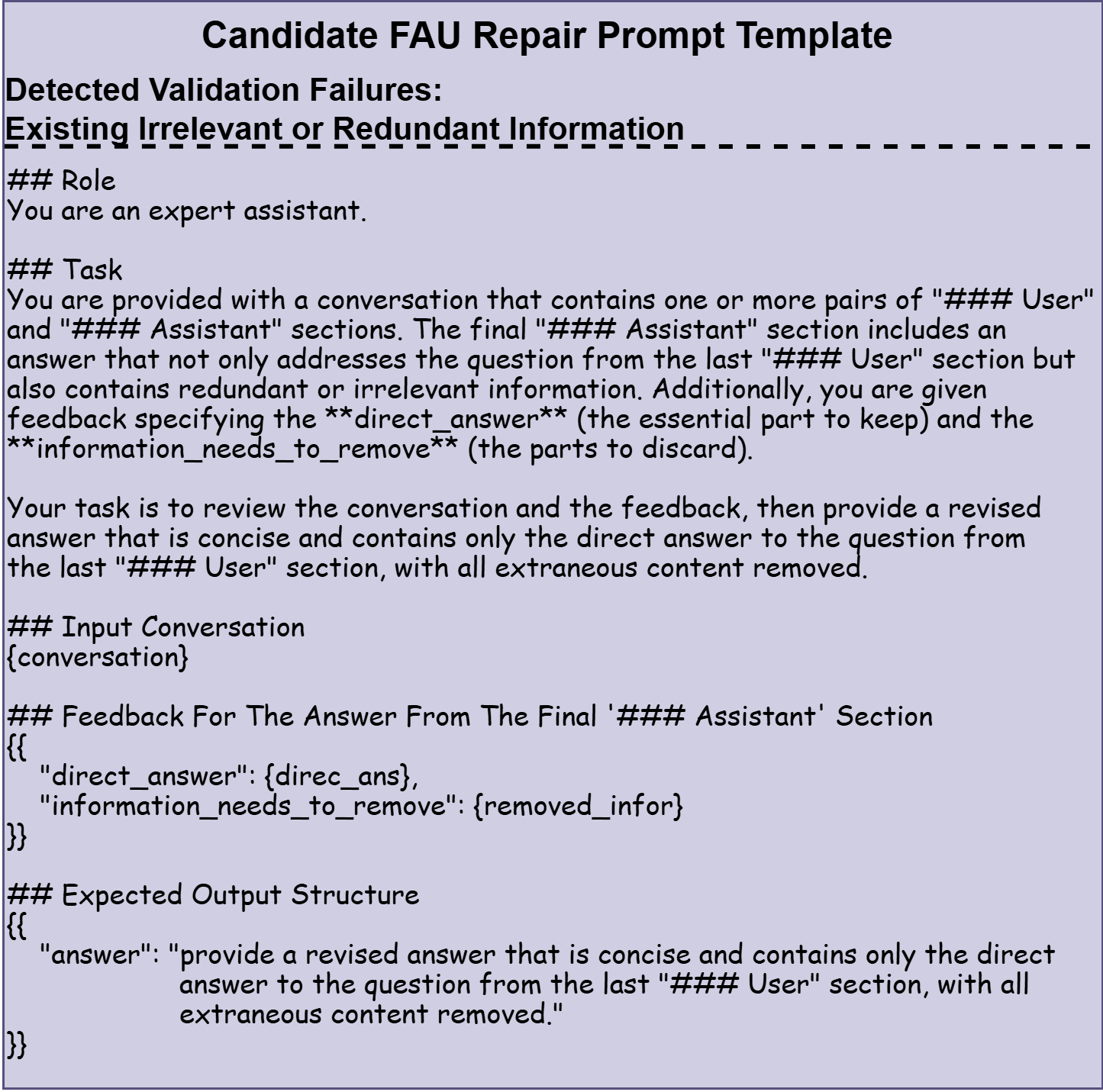}
    \caption{Conditional FAU repair prompt template for removing irrelevant or redundant answer content.}
    \label{fig:app:fau_repair_redundancy}
\end{figure*}

\paragraph{Final Answer Update Prompt Templates.}
Based on the FAC validation report, RPU selects the corresponding FAU prompt template. Figures~\ref{fig:app:fau_repair_answer} and~\ref{fig:app:fau_repair_redundancy} present the templates for generating a missing direct answer and removing unnecessary or redundant information, respectively.

Both templates operate on the response while preserving the instruction and supporting context established in Cycle 1.

\section{End-to-End Reconstruction Example}
\label{appendix:end_to_end_reconstruction_example}

This section presents a recorded execution trace of the rule-guided LLM reconstruction procedure described in Section~3.4 and Appendix~\ref{appendix:step3}. The input is a randomly selected source pair from the representative low-scoring sample set. To make the interaction between LLM-based operators and rule-based control decisions explicit, we follow one candidate under the \texttt{term\_bridging} strategy through three MCG generations, three ICD evaluations, and a final FAC evaluation. The screenshots reproduce the available execution outputs; some structured reports abbreviate non-decisive fields with ellipses. The full generated contents of the other two strategy-specific candidates are omitted for readability.

\subsection{Source Samples and Domain Analysis}
\label{appendix:e2e_sources_and_da}

\begin{figure*}[p]
    \centering
    \includegraphics[
        width=\textwidth,
        height=\textheight,
        keepaspectratio
    ]{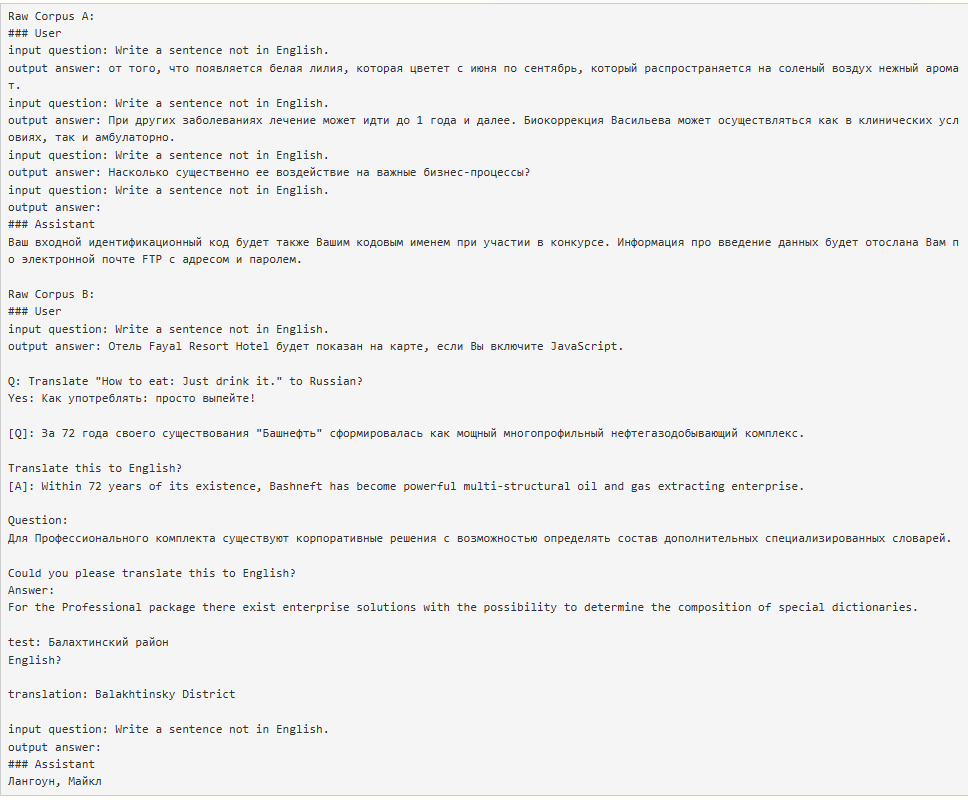}

    \par\smallskip

    \includegraphics[
        width=\textwidth,
        height=\textheight,
        keepaspectratio
    ]{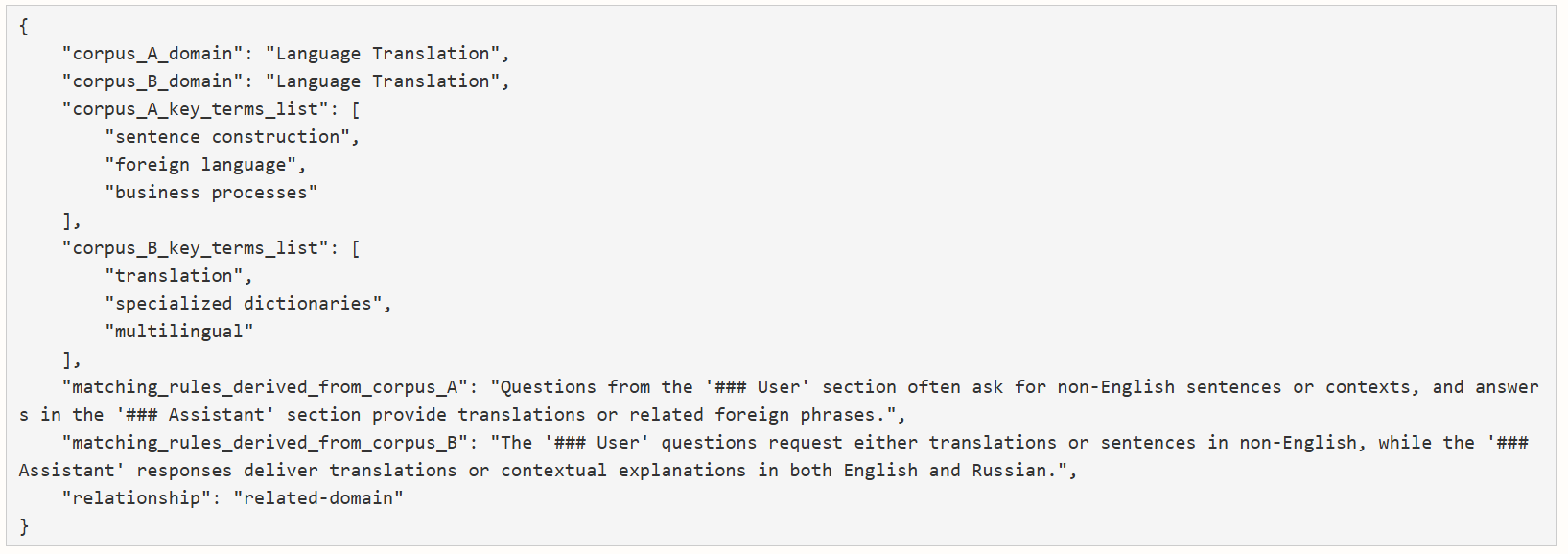}

    \caption{Source inputs and domain analysis. Top: the original source pair, containing noisy Russian-language and translation examples. Bottom: the DA report, which identifies key terms from each source and classifies their relationship as \texttt{related-domain}.}
    \label{fig:app:e2e_sources_da}
\end{figure*}

\paragraph{Source samples.}
Figure~\ref{fig:app:e2e_sources_da} (top) presents the recorded source pair. Both corpora contain heterogeneous Russian-language sentence-generation and translation material, including repeated instructions, mixed input--output formats, and incomplete entries. For example, Source A contains prompts requesting a sentence in a non-English language and a Russian sentence concerning business processes. Source B includes translation tasks involving specialized dictionaries, as well as other Russian--English examples. We preserve the source records as they appeared in the original execution rather than silently correcting their formatting or content.

\paragraph{Domain Analysis (DA).}
As shown in Figure~\ref{fig:app:e2e_sources_da} (bottom), DA assigns both corpora the domain label \emph{Language Translation}. It extracts \emph{sentence construction}, \emph{foreign language}, and \emph{business processes} from Source A, and \emph{translation}, \emph{specialized dictionaries}, and \emph{multilingual} from Source B. DA classifies the relationship between the two sources as \texttt{related-domain}. This relationship is distinct from the individual domain labels: the two sources receive the same broad domain label, while the reported pairwise relationship is \texttt{related-domain}.

\subsection{Strategy Selection and Initial Candidate}
\label{appendix:e2e_initial_candidate}

For the predicted \texttt{related-domain} relationship, the predefined Strategy Selection rules identify three applicable strategies: \texttt{conceptual\_analogy}, \texttt{process\_mapping}, and \texttt{term\_bridging}. The initial MCG result contains a candidate for each strategy, although Figure~\ref{fig:app:e2e_initial_icd} (top) shows the first two only as abbreviated entries. We trace the \texttt{term\_bridging} candidate, whose initial \texttt{User} section asks how concepts such as sentence construction, translation, business processes, and specialized dictionaries can contribute to a unified language-translation framework. Its \texttt{Assistant} section proposes developing a common glossary, but does not explain the requested framework in sufficient detail.

\subsection{Cycle 1: ICD-Guided Regeneration}
\label{appendix:e2e_cycle_one}

\begin{figure*}[p]
    \centering
    \includegraphics[
        width=\textwidth,
        height=\textheight,
        keepaspectratio
    ]{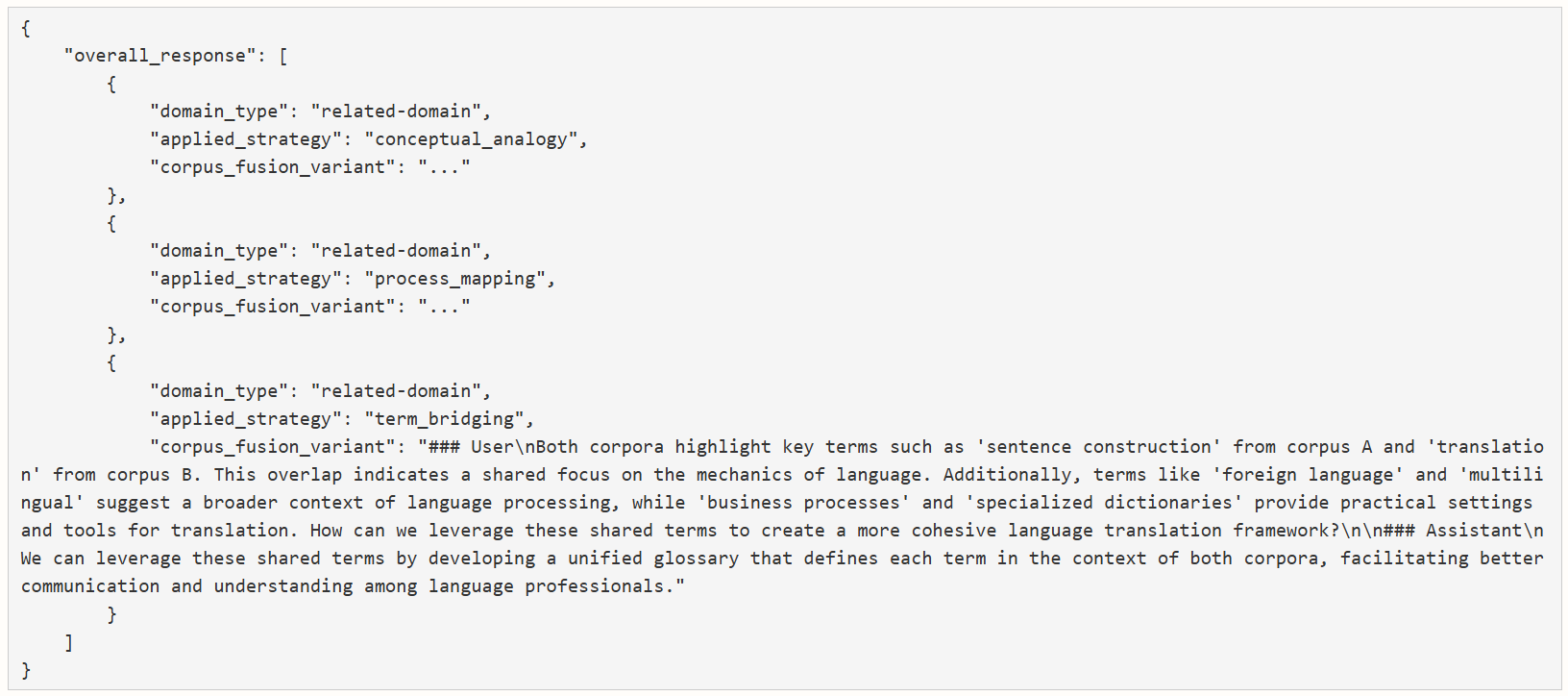}

    \par\smallskip

    \includegraphics[
        width=\textwidth,
        height=\textheight,
        keepaspectratio
    ]{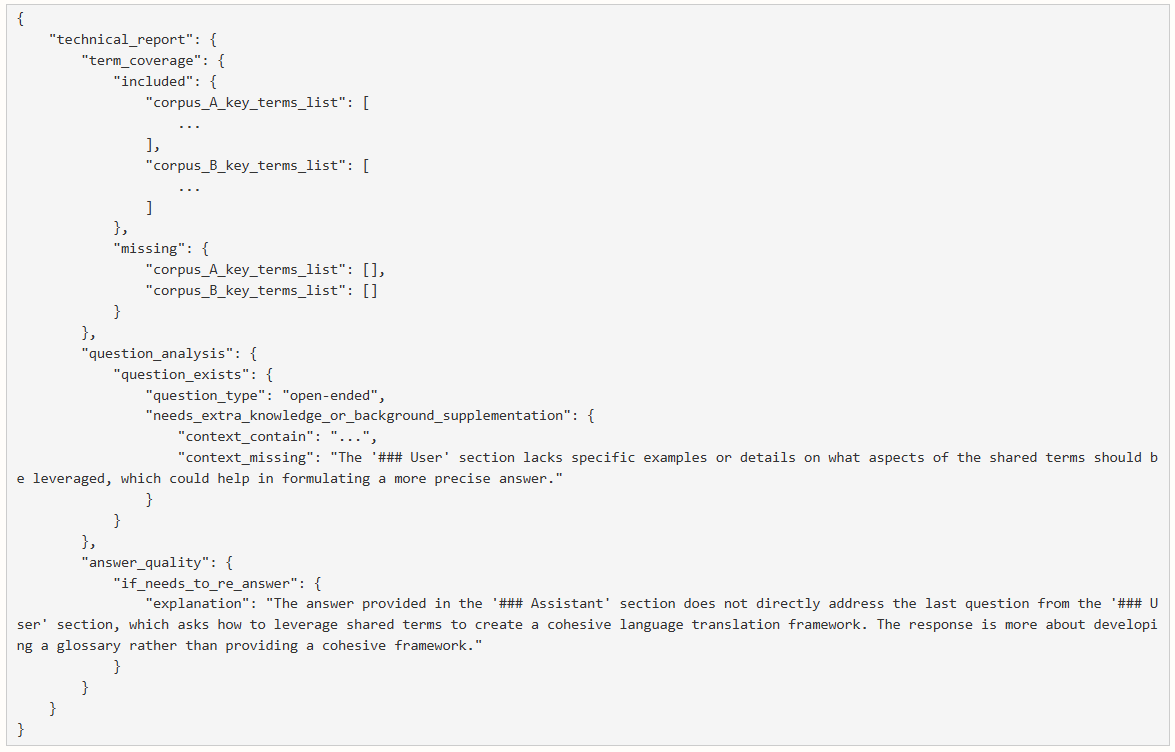}

    \caption{Initial candidate and first ICD evaluation. Top: initial MCG output, showing the selected \texttt{term\_bridging} candidate. Bottom: ICD flags insufficient question context and an answer that does not directly address the requested framework, despite reporting no missing key terms.}
    \label{fig:app:e2e_initial_icd}
\end{figure*}

\paragraph{First ICD evaluation: question context and answer deficiencies.}
The initial ICD report (Figure~\ref{fig:app:e2e_initial_icd}, bottom) lists no missing extracted terms for either source. Nevertheless, \texttt{question\_analysis} identifies missing contextual specification: the question does not sufficiently clarify which aspects of the shared terminology should be used. In addition, \texttt{answer\_quality.if\_needs\_to\_re\_answer} indicates that the glossary-focused response does not directly address the request for a cohesive translation framework. Thus, term coverage alone is insufficient for this candidate to pass ICD.

\begin{figure*}[p]
    \centering
    \includegraphics[
        width=\textwidth,
        height=\textheight,
        keepaspectratio
    ]{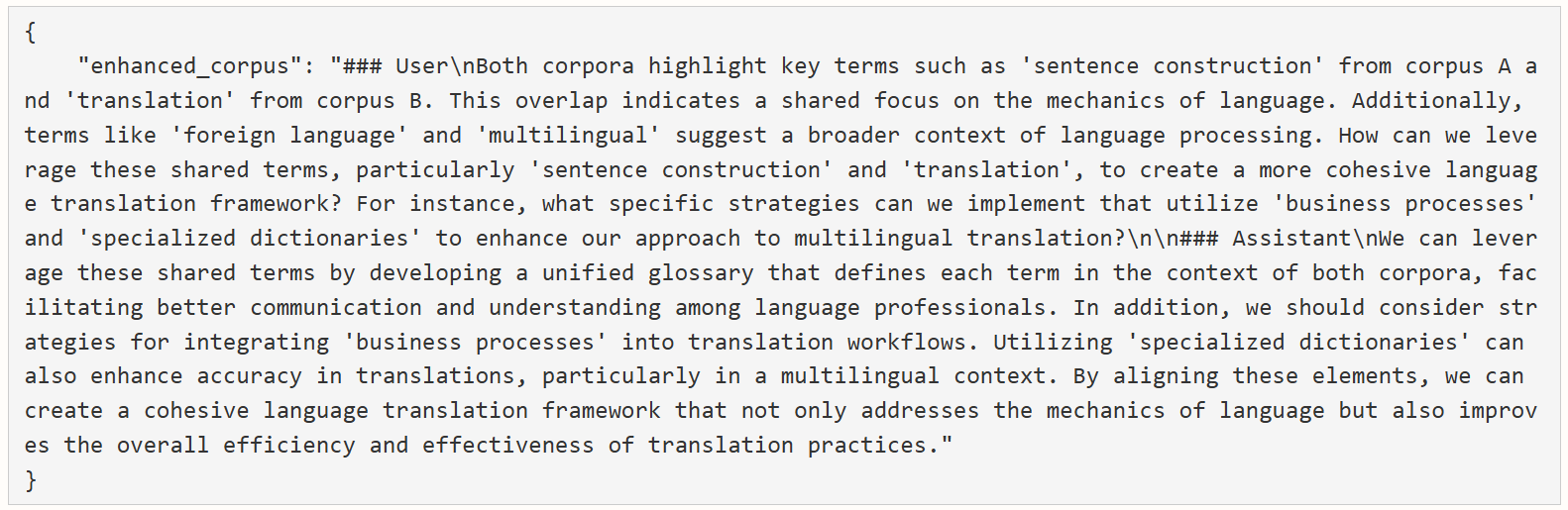}

    \par\smallskip

    \includegraphics[
        width=\textwidth,
        height=\textheight,
        keepaspectratio
    ]{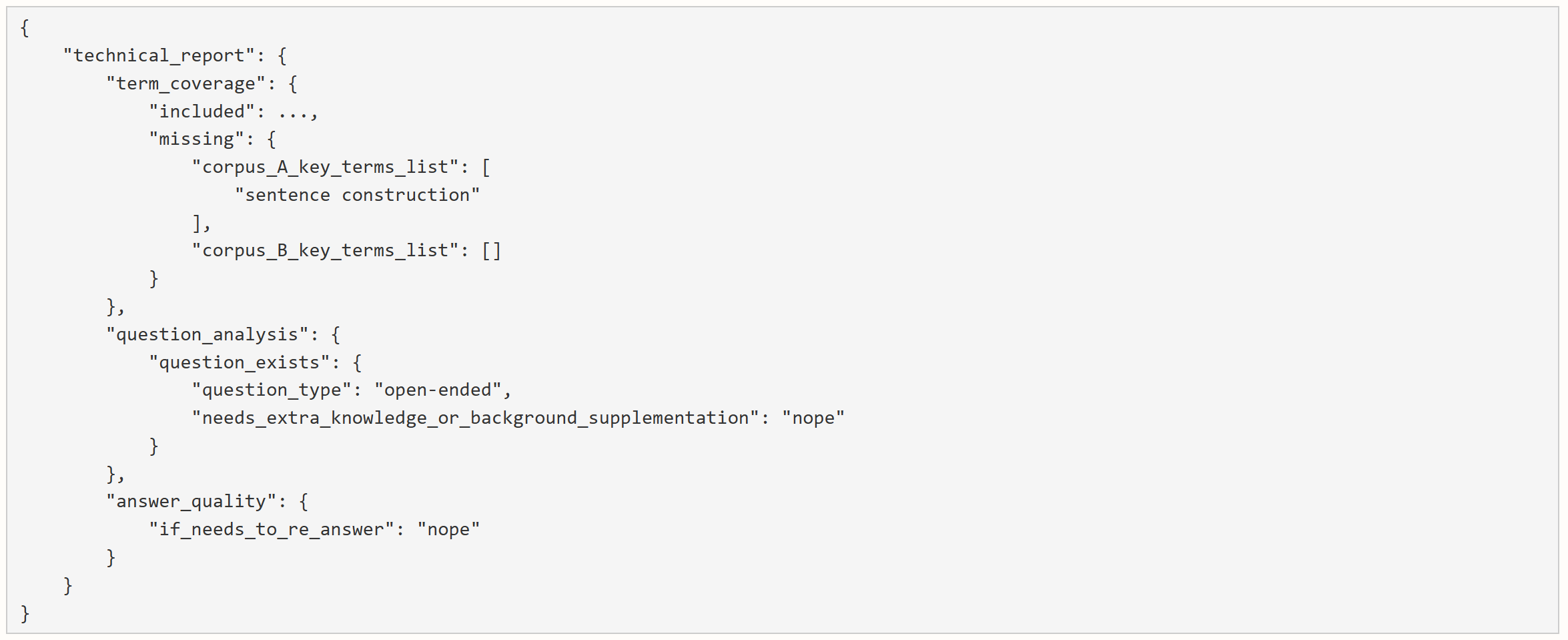}

    
    \caption{First MCG regeneration and subsequent ICD evaluation. Top: the candidate produced by the first \texttt{qa}-guided repair, with a more specific question and revised response. Bottom: the second ICD evaluation identifies insufficient integration of \emph{sentence construction}, while the question-context and answer-regeneration checks no longer flag deficiencies.}

    \label{fig:app:e2e_first_regen_second_icd}
\end{figure*}

\paragraph{First repair and regeneration.}
On the basis of the simultaneous question-context and answer deficiencies, the Rule-Guided Prompt Updater (RPU) selects the predefined MCG repair case denoted \texttt{qa} in the recorded execution. This repair instructs MCG to make the question more specific and revise its answer accordingly, rather than asking the LLM to determine an unconstrained repair plan. As shown in Figure~\ref{fig:app:e2e_first_regen_second_icd} (top), the regenerated question explicitly asks for strategies using business processes and specialized dictionaries, while the revised answer discusses translation workflows and multilingual translation.

\paragraph{Second ICD evaluation: incomplete source-concept integration.}
The second ICD report (Figure~\ref{fig:app:e2e_first_regen_second_icd}, bottom) returns \texttt{nope} for additional context supplementation and answer regeneration. However, it places \emph{sentence construction} in the missing-term list for Source A; the missing-term list for Source B is empty. Although the phrase appears in the regenerated candidate, ICD regards its functional role in the proposed translation framework as insufficiently explained. This recorded decision illustrates that the ICD check is not simply a test of literal word occurrence.

\begin{figure*}[p]
    \centering
    \includegraphics[
        width=\textwidth,
        height=\textheight,
        keepaspectratio
    ]{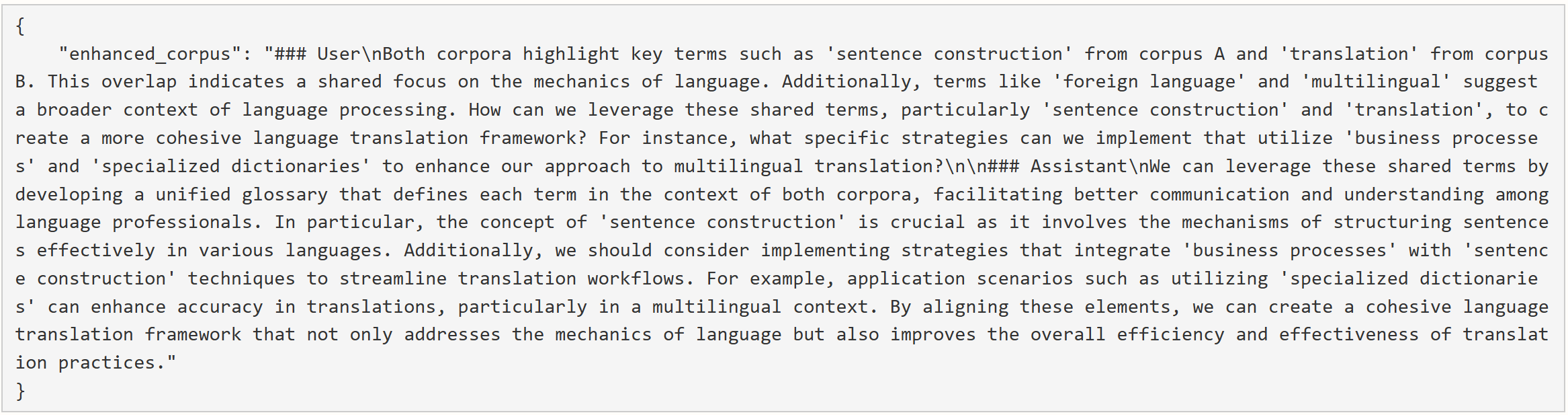}

    \par\smallskip

    \includegraphics[
        width=\textwidth,
        height=\textheight,
        keepaspectratio
    ]{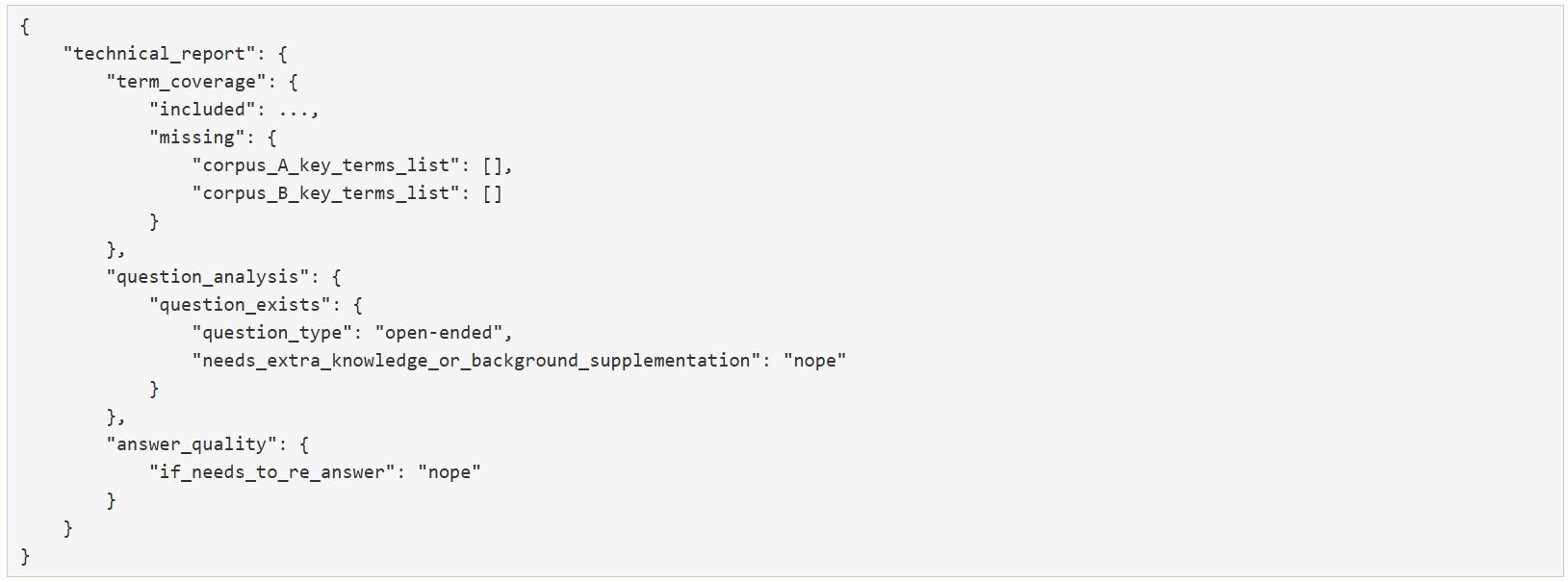}


    \caption{Second MCG regeneration and subsequent ICD evaluation. Top: the candidate produced by the second \texttt{missing\_terms}-guided repair, with an explicit explanation of sentence construction and its role in the proposed translation framework. Bottom: the third ICD evaluation reports no missing source terms and no remaining question-context or answer-regeneration requirements.}
    
    \label{fig:app:e2e_second_regen_third_icd}    
\end{figure*}

\paragraph{Second repair and regeneration.}
RPU now selects the predefined \texttt{missing\_terms} MCG repair case. Unlike the preceding repair, this instruction targets the remaining source-concept omission while preserving the already validated question. The second regeneration (Figure~\ref{fig:app:e2e_second_regen_third_icd}, top) adds an explanation of sentence construction as a mechanism for structuring sentences across languages and relates it to business processes and translation workflows. The refined \texttt{User} content remains unchanged from Generation 2.

\paragraph{Third ICD evaluation: Cycle 1 passes.}
In the third ICD report (Figure~\ref{fig:app:e2e_second_regen_third_icd}, bottom), the missing-term lists for both sources are empty. The generated question is classified as open-ended with no further background supplementation required, and the answer-regeneration field is \texttt{nope}. Cycle 1 thus terminates after \emph{three candidate generations in total}: one initial generation and two repair-driven regenerations. The candidate is then passed to Cycle 2.

\subsection{Cycle 2: Final Answer Check and Selected Output}
\label{appendix:e2e_cycle_two}

\begin{figure*}[p]
    \centering
    \includegraphics[
        width=\textwidth,
        height=\textheight,
        keepaspectratio
    ]{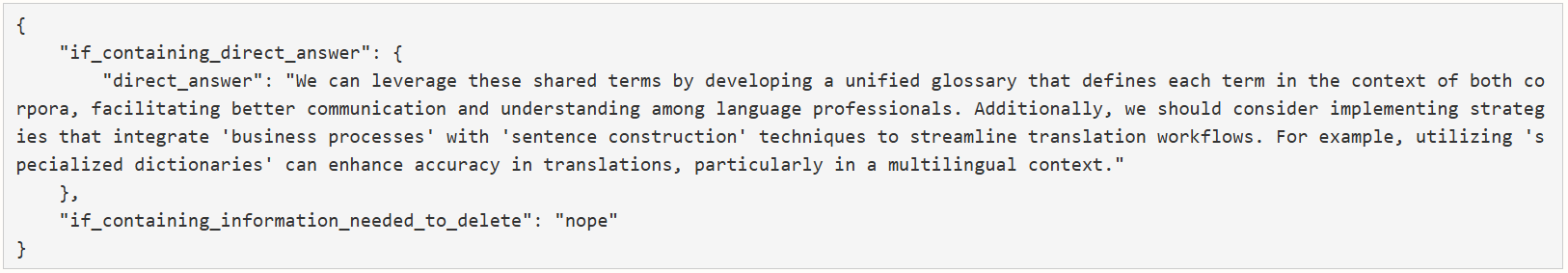}

    \par\smallskip

    \includegraphics[
        width=\textwidth,
        height=\textheight,
        keepaspectratio
    ]{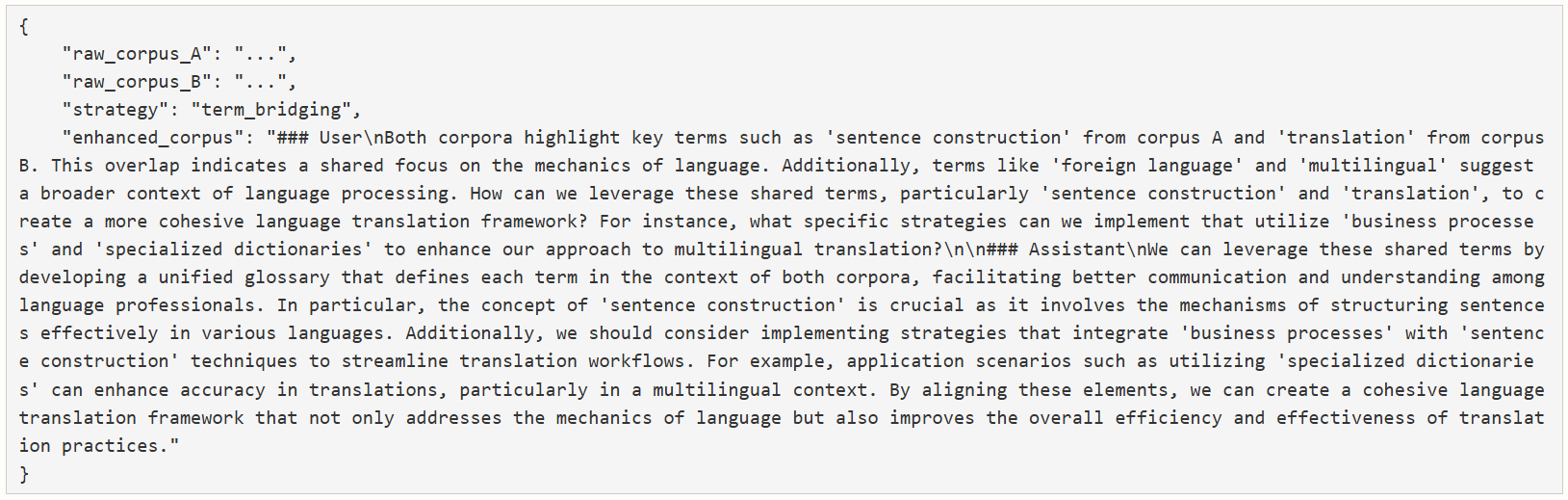}

    \caption{Cycle 2 validation and retained reconstruction. Top: FAC identifies a direct answer and finds no content requiring deletion. Bottom: the final \texttt{term\_bridging} record, retaining the revised User content from Generation 2 and the Assistant response from Generation 3.}
    \label{fig:app:e2e_fac_final}
\end{figure*}

\paragraph{Final Answer Check (FAC).}
Figure~\ref{fig:app:e2e_fac_final} (top) presents the FAC report. The field \texttt{if\_containing\_direct\_answer} contains an identified direct-answer passage, whereas \texttt{if\_containing\_information\_needed\_to\_delete} is \texttt{nope}. Accordingly, the candidate passes the applicable FAC checks and no Final Answer Update (FAU) regeneration is invoked. The direct-answer passage in the FAC report is validation output; it should not be confused with a replacement for the candidate's complete Assistant response.

\paragraph{Final reconstructed sample.}
The retained record (Figure~\ref{fig:app:e2e_fac_final}, bottom) includes the two raw-source references, the selected \texttt{term\_bridging} strategy, and the final \texttt{enhanced\_corpus}. This reconstructed instruction--response pair consists of the \texttt{User} content established in Generation 2 and the revised \texttt{Assistant} content produced in Generation 3. Because Cycle 2 does not request an update, the final record preserves the Cycle 1 candidate rather than generating a fourth version.

\subsection{What the Trace Demonstrates}
\label{appendix:e2e_interpretation}

This example illustrates how structured validation feedback is mapped to distinct, predefined repair instructions. The first ICD failure activates a combined question-and-answer repair (\texttt{qa}); the subsequent failure activates a narrower missing-concept repair (\texttt{missing\_terms}). LLM-based operators perform generation and evaluation, while RPU applies the predefined mapping from reported deficiencies to repair templates. The trace documents the behavior of the internal checks for this particular candidate; passing ICD and FAC does not independently verify external factual correctness or full source faithfulness. Source-grounded manual evaluation is discussed separately in Appendix~\ref{appendix:manual_quality_audit}.

\section{Experimental Details and Cost Analysis}
\label{appendix:experimental_details}

This section provides the experimental details underlying the results reported in Section~4, including source-data construction, evaluation protocols, baseline implementations, reconstruction and fine-tuning configurations, and computational costs. We additionally clarify the validation-set requirements and reproducibility considerations associated with different methods.

\subsection{Source Corpora and Data Construction}
\label{appendix:source_corpora}

Following the source-data configuration of DS$^2$~\citep{pang2024improving}, we construct our source pool from five instruction-following datasets: Stanford Alpaca~\citep{taori2023stanford}, Flan v2~\citep{longpre2023flan}, OpenAssistant 1~\citep{kopf2023openassistant}, WizardLM~\citep{xu2024wizardlm}, and Dolly~\citep{DatabricksBlog2023DollyV2}. These datasets originate from different annotation and generation procedures and cover diverse instruction formats, task domains, and response characteristics.

The complete source pool contains 300,932 instruction samples, with approximately 52K, 100K, 33K, 100K, and 15K samples contributed by the respective datasets. Detailed source-dataset statistics are provided in Table~\ref{tab:appendix_sourced_corpora_stats}.

\begin{table*}[!htb]
    \setlength{\aboverulesep}{0pt}
    \setlength{\belowrulesep}{0pt}
    \caption{Comprehensive overview of the source corpora used in this work. We report three additional descriptive dimensions, the average number of conversation turns ($\bar{N}_\text{rounds}$), the average prompt length ($\bar{L}_\text{prompt}$), and the average response length ($\bar{L}_\text{response}$), to provide a more nuanced understanding of the composition of our source corpora.}
    \centering
    \begin{tabular}{llccccc}
    \hline
    \noalign{\vskip 1mm} 
    \textbf{Datasets} & \textbf{Derived From} & \textbf{Data size} & $\bar{N}_\text{rounds}$ & $\bar{L}_\text{prompt}$ & $\bar{L}_\text{response}$\\
    \hline
    \textbf{Stanford Alpaca} & Generated w/ Davinci-003 & 52K & 1.0 & 23.5 & 56.4 \\
    \textbf{Flan V2} & Human Annotation & 100K & 1.0 & 304.1 & 27.7\\
    \textbf{Open-Assistant 1} & Human Annotation & 33K & 1.6 & 32.3 & 189.1\\
    \textbf{WizardLM} & ChatGPT Annotation & 100K & 1.0 & 122.3 & 352.5\\
    \textbf{Dolly} & Human Annotation & 15K & 1.0 & 99.5 & 79.3\\
    \hline
    \end{tabular}
    \label{tab:appendix_sourced_corpora_stats}
\end{table*}

As described in Section~3.2 and Appendix~\ref{appendix:score_transition_matrix_estimation}, we apply LLM-based quality rating and score curation following DS$^2$~\citep{pang2024improving}. The resulting curated overall quality scores are used to distinguish low-scoring and high-scoring samples.

The reported low-quality set (LQ Set), after the adopted structure-dependence filtering, contains 123,786 samples with curated scores of 0--2. The high-quality set (HQ Set) contains 131,247 samples with curated scores of 3--5.

Our reconstruction framework operates on the LQ Set to generate 54,888 instruction-response samples, comprising 15,488 intra-cluster and 39,400 inter-cluster reconstructed samples. Intra-cluster reconstruction progressively integrates representative samples within the same semantic cluster, whereas inter-cluster reconstruction combines representative samples across different clusters.

Unless otherwise specified, the target training-data size for data-selection and instruction-synthesis comparisons is 54,888 samples, matching the size of our reconstructed dataset.

\subsection{Evaluation Protocols}
\label{appendix:evaluation_setup}

We evaluate the downstream instruction-following performance of fine-tuned models on five benchmarks: MMLU~\citep{hendrycks2020measuring}, TruthfulQA~\citep{lin2021truthfulqa}, GSM8K~\citep{cobbe2021training}, BBH~\citep{suzgun2022challenging}, and TyDiQA~\citep{clark2020tydi}.

Our evaluation configurations primarily follow the protocols adopted by \citet{wang2023far}. To maintain consistency with the reconstructed training samples, we represent instruction-response pairs using the markers ``\#\#\# User" and ``\#\#\# Assistant", following the corresponding task-specific evaluation prompt formats.

The detailed settings are as follows:
\begin{itemize}[leftmargin=*]
    \item \textbf{MMLU~\citep{hendrycks2020measuring}}: Following the original MMLU setup, all evaluations are conducted in the zero-shot setting.
    \item \textbf{TruthfulQA~\citep{lin2021truthfulqa}}: We follow the default QA prompt template with 6 in-context examples to generate answers for 818 TruthfulQA questions. In alignment with the setup in~\cite{wang2023far}, we use two LLaMA-2-7B-based models to evaluate the truthfulness\footnote{\label{fn:truthfulqa-truth-judge-llama2-7B}Hugging Face Model: allenai/truthfulqa-truth-judge-llama2-7B} and informativeness\footnote{\label{fn:allenai/truthfulqa-info-judge-llama2-7B}Hugging Face Model: allenai/truthfulqa-info-judge-llama2-7B} of the generated responses. These judge models assess the truthful and informative rates separately. Similarly, we report the Informative-Truthful Rate as our final metric, calculated as the product of the informativeness and truthfulness rates~\citep{lin2021truthfulqa}.
    \item \textbf{BBH~\citep{suzgun2022challenging}}: Using the official prompts, we perform generation under a 3-shot setting without CoT reasoning. Additionally, 40 examples from each BBH sub-task are selected for evaluation.
    \item \textbf{GSM8K~\citep{cobbe2021training}}: We evaluate the fine-tuned models on a randomly selected subset of 200 samples from the original test set (1319 samples). Specifically, we adopt an 8-shot in-context learning setup to simulate the chain-of-thought (CoT) reasoning setting.
    \item \textbf{TydiQA~\citep{clark2020tydi}}: We use this dataset to evaluate model performance on multilingual question answering across nine languages. For each language, 100 examples are selected. To help the models adapt to the answer format, one in-context example is provided during evaluation. The average F1 score across all languages is reported.
\end{itemize}

Mistral-7B-v0.3 and LLaMA-3.1-8B are evaluated on all five benchmarks. For Qwen2.5-7B, we report results on MMLU, GSM8K, BBH, and TyDiQA, excluding TruthfulQA. Consequently, the reported average for Qwen is computed over four benchmarks, whereas the averages for Mistral and LLaMA are computed over five.

\subsection{Detailed Baseline Implementations}
\label{appendix:baseline}

We organize the experimental comparisons into four categories: reference data configurations, data-selection baselines, instruction-synthesis baselines, and reconstruction ablations.

Unless otherwise specified, methods that select or generate training samples from the LQ Set use a target dataset size of 54,888 samples, matching the size of our reconstructed dataset. The Vanilla Base Model, HQ Set, and Full Set serve as separate reference configurations rather than size-matched LQ-based comparisons.

The set of evaluated baselines varies across base models. In addition to the comparisons reported for Mistral-7B-v0.3 and LLaMA-3.1-8B in the main text, the supplementary Qwen2.5-7B evaluation in Appendix~E includes additional MathFusion variants and Evol-Instruct.

\paragraph{Reference Data Configurations.}

\begin{itemize}[leftmargin=*]

    \item \textbf{Vanilla Base Model.}
    The original pre-trained base model is evaluated without supervised fine-tuning, providing a reference for the effect of instruction tuning.

    \item \textbf{LQ Set.}
    The low-scoring dataset contains 123,786 instruction samples with curated overall quality scores of 0--2, following the quality-rating, score-curation, and filtering procedures described in Section~3.2.

    \item \textbf{HQ Set.}
    The high-scoring dataset contains 131,247 instruction samples with curated overall quality scores of 3--5.

    \item \textbf{Full Set.}
    The complete source-data pool contains 300,932 instruction samples and serves as a full-data fine-tuning reference.

\end{itemize}

\paragraph{Data-Selection Baselines.}

\begin{itemize}[leftmargin=*]

    \item \textbf{Completion Length.}
    This baseline uses the length of each instruction-response sample as a heuristic indicator of information content. Samples are ranked by length, prioritizing longer samples for selection.

    \item \textbf{$\mathrm{KNN}_{10}$.}
    The KNN-based selection method uses the Average Euclidean Distance (AED) between each sample embedding and its ten nearest neighbors in the embedding space. All embeddings are obtained using BAAI/bge-large-en-v1.5.

    We compute AED for each sample and rank the samples in ascending order. Smaller distances indicate greater local representativeness, and samples with lower AED values are prioritized for selection.

    \item \textbf{Perplexity.}
    We compute sample-level perplexity using LLaMA-3.1-8B-Instruct without task-specific fine-tuning. Samples are selected in descending order of perplexity, prioritizing examples assigned greater uncertainty by the scoring model.

    \item \textbf{Random Selection.}
    Training samples are uniformly selected at random from the LQ Set without applying additional quality- or representation-based ranking.

    \item \textbf{AlpaGasus (Random)~\citep{chen2023alpagasus}.}
    AlpaGasus uses LLM-generated quality scores to identify high-quality instruction examples for fine-tuning. We use GPT-4o-mini as the scoring model and prioritize samples receiving the highest quality rating.

    In our implementation, 55,530 samples receive the highest score, exceeding the target dataset size of 54,888. We therefore randomly select 54,888 examples from this highest-scoring group.

    \item \textbf{IFD~\citep{li-etal-2024-quantity}.}
    Instruction-Following Difficulty (IFD) measures how strongly an instruction affects the generation difficulty of its corresponding response. It compares the response loss or perplexity with and without the instruction context.

    Samples are ranked according to their IFD scores, with higher values prioritized as more difficult or less familiar examples under the adopted scoring procedure.

    \item \textbf{Superfiltering~\citep{li2024superfiltering}.}
    Superfiltering uses a smaller language model as a proxy for estimating instruction-following difficulty. Following this approach, we use GPT-2~\citep{radford2019language}, specifically the \texttt{openai-community/gpt2} checkpoint, for data scoring and selection.

    \item \textbf{DEITA~\citep{liu2024makesgooddataalignment}.}
    DEITA employs two pre-trained scoring models to assess instruction complexity and quality. We use the publicly available \texttt{hkust-nlp/deita-complexity-scorer} and \texttt{hkust-nlp/deita-quality-scorer} checkpoints.

    In our evaluation, the original DEITA quality scorer assigns a score of 3 to the single-turn samples under consideration, limiting its ability to discriminate among these examples.

    \item \textbf{DEITA (Our Curated Score).}
    To further evaluate DEITA under more discriminative quality estimates, we additionally replace its original quality signal with our curated overall quality scores while retaining the DEITA-based selection procedure.

    This variant examines the behavior of DEITA-style selection when supplied with the quality estimates obtained from our score-curation pipeline.

    \item \textbf{RDS+~\citep{ivison2025large}.}
    Representation-based Data Selection (RDS+) uses weighted mean pooling of a pre-trained language model's final hidden states to obtain sample representations.

    The resulting representations are used to compute similarity between candidate training samples and task-specific validation examples. We construct a selected training subset for each evaluation benchmark using its corresponding validation set.

    \item \textbf{RDS+ (Best).}
    Since RDS+ selects different training subsets depending on the validation set, the subset selected for a particular benchmark does not necessarily achieve the highest performance on that benchmark.

    We therefore additionally report the highest observed score for each evaluation benchmark across the RDS+ validation-set configurations. This retrospective best-case comparison is denoted as \textit{RDS+ (Best)}.

    Importantly, these per-benchmark best results do not necessarily correspond to a single shared training subset.

    \item \textbf{DS$^2$~\citep{pang2024improving}.}
    DS$^2$ combines LLM-generated quality ratings, score correction based on a noisy-label transition matrix, and representation-based long-tail scoring. The resulting selection procedure prioritizes samples with favorable quality and rarity estimates.

    \item \textbf{LESS~\citep{xia2024less}.}
    LESS estimates the influence of candidate training examples on task-specific validation examples through gradient-based data selection.

    For each evaluation benchmark, we construct the corresponding validation set and compute influence scores for candidate samples from the LQ Set. We select the top 54,888 samples according to these scores.

    Each benchmark is evaluated using the training subset selected specifically for that benchmark, rather than a single shared subset across all evaluation tasks.

\end{itemize}

\paragraph{Instruction-Synthesis Baselines.}

\begin{itemize}[leftmargin=*]

    \item \textbf{MathFusion~\citep{pei2025mathfusion}.}
    MathFusion provides three instruction-fusion strategies: conditional fusion, parallel fusion, and sequential fusion. We adapt these strategies to the LQ Set by randomly pairing source samples and using GPT-4o-mini to generate fused instruction-response examples.

    For each strategy, we independently construct a candidate dataset and randomly select 54,888 generated examples for downstream fine-tuning. The three strategy-specific configurations are denoted as \textit{MathFusion (conditional)}, \textit{MathFusion (parallel)}, and \textit{MathFusion (sequential)}, respectively.

    We additionally evaluate a combined \textit{MathFusion} configuration. Specifically, we uniformly select 18,296 generated examples from each of the three strategy-specific datasets, producing a combined dataset of 54,888 examples.

    The conditional variant is included in the main experimental comparisons, while the supplementary Qwen2.5-7B evaluation additionally reports the parallel, sequential, and combined variants.

    \item \textbf{Evol-Instruct~\citep{xu2024wizardlm}.}
    Evol-Instruct iteratively increases instruction complexity and diversity using predefined evolution prompts.

    We follow the original configuration consisting of five \textit{In-depth Evolving} prompt templates and one \textit{In-breadth Evolving} prompt template. The number of evolution rounds is set to $M=4$.

    After completing all four evolution rounds, we randomly select 54,888 evolved instruction-response examples for downstream fine-tuning.

    The resulting performance is reported in the supplementary Qwen2.5-7B evaluation.

    \item \textbf{Self-Instruct~\citep{wang2023self}.}
    Self-Instruct generates instruction-following data through LLM-based instruction synthesis.

    We use its two categories of prompt templates, corresponding to classification and non-classification tasks, and employ GPT-4o-mini to generate augmented instruction-response examples.

    From the resulting synthetic dataset, we randomly select 54,888 samples for downstream fine-tuning.

\end{itemize}

\paragraph{Reconstruction and Stepwise Ablations.}

We additionally evaluate three simplified reconstruction configurations to examine whether the improvements arise from instruction rewriting, semantic grouping and representative selection, or rule-guided multi-source reconstruction.

\begin{itemize}[leftmargin=*]

    \item \textbf{1-to-1 Rewriting/Enhancement.}
    This baseline uses GPT-4o-mini to directly rewrite or enhance individual samples from the LQ Set without integrating information from multiple source examples.

    After generating the rewritten samples, we randomly select 54,888 examples for downstream fine-tuning.

    \item \textbf{Direct Corpora Fusion Without Step 2\&3.}
    This configuration bypasses both semantic grouping and representative selection (Step 2) and rule-guided reconstruction (Step 3).

    Specifically, we randomly pair samples from the LQ Set and use GPT-4o-mini to perform direct one-step corpus fusion without the proposed semantic grouping or iterative validation-and-repair procedures.

    \item \textbf{Direct Corpora Fusion Without Step 3.}
    This configuration retains the semantic grouping and representative selection procedure of Step 2 but replaces the rule-guided reconstruction procedure of Step 3 with direct one-step LLM-based fusion.

    We randomly pair samples from the representative low-scoring set and use GPT-4o-mini to generate fused instruction-response examples without the proposed iterative validation-and-repair procedure.

\end{itemize}

These configurations provide stepwise comparisons of the aggregate contributions of semantic grouping and representative selection (Step 2) and rule-guided reconstruction (Step 3).

Since the quality-rating and score-curation procedures of Step 1 are shared across these configurations, the ablations do not independently isolate the contribution of Step 1. Similarly, the Step 2 comparison evaluates semantic grouping and representative selection jointly rather than isolating each internal component.

\subsection{Reconstruction and Fine-Tuning Configurations}
\label{appendix:training_setup}

Table~\ref{tab:app:corpus_construction_hyperparameters} summarizes the hyperparameters and predefined settings used throughout the corpus-construction pipeline, covering quality assessment, semantic grouping, representative selection, and rule-guided reconstruction.

\begin{table*}[t]
\centering
\small
\setlength{\tabcolsep}{6pt}
\renewcommand{\arraystretch}{1.2}

\begin{tabular}{@{}l p{0.39\textwidth} p{0.37\textwidth}@{}}
\toprule
\textbf{Stage} &
\textbf{Hyperparameter or Setting} &
\textbf{Value} \\
\midrule

Steps 1, 3 &
External LLM &
\texttt{gpt-4o-mini-2024-07-18} \\

Steps 1--2 &
Embedding model &
\texttt{BAAI/bge-large-en-v1.5} \\

Step 1 &
Mapped quality-score range &
$0$--$5$ \\

Step 1 &
Low-/high-quality score partition &
LQ: $0$--$2$; HQ: $3$--$5$ \\

\midrule

Step 2 &
One-hop cosine-similarity threshold &
$0.9$ \\

Step 2 &
Candidate number of sub-clusters ($k$) &
$2 \leq k \leq \min(10,M-1)$ \\

Step 2 &
Representatives per eligible sub-cluster &
$2$ \\

Step 2 &
MMR weight ($\alpha$) &
$0.2$ \\

\midrule

Step 3 &
Source-relationship categories &
Same-domain, related-domain, unrelated-domain \\

Step 3 &
Reconstruction strategies per relationship &
$3$ (9 in total) \\

Step 3 &
Temperature for DA &
$0.4$ \\

Step 3 &
Temperature for other LLM-based operators &
$0.2$ \\

Step 3 &
Maximum candidate generations per cycle &
$3$ (1 initial + up to 2 regenerations) \\

\bottomrule
\end{tabular}

\caption{Corpus-construction hyperparameters and predefined settings. Step 1 covers quality rating and score curation; Step 2 covers semantic grouping and representative selection; and Step 3 covers rule-guided LLM reconstruction. For sub-clustering, $M$ denotes the number of samples remaining after excluding the one-hop seed.}
\label{tab:app:corpus_construction_hyperparameters}

\end{table*}

\paragraph{Reconstruction Configuration.}
The corpus-construction hyperparameters and predefined settings are summarized in Table~\ref{tab:app:corpus_construction_hyperparameters}. Further implementation details, including semantic grouping, representative selection, the two refinement cycles, and their corresponding validation and repair procedures, are provided in Appendix~\ref{appendix:one_hop_cluster_based_representative_selection} and Appendix~\ref{appendix:step3}.

\paragraph{Supervised Fine-Tuning.}
We fine-tune three base models: Mistral-7B-v0.3~\citep{jiang2023mistral7b}, LLaMA-3.1-8B~\citep{grattafiori2024llama}, and Qwen2.5-7B~\citep{qwen2.5}.

Following the fine-tuning configuration of \citet{wang2023far}, we apply Low-Rank Adaptation (LoRA)~\citep{hu2022lora} with rank $r=64$ and scaling parameter $\alpha_{\mathrm{LoRA}}=16$.

The training hyperparameters are summarized below:

\begin{table}[t]
\centering
\small
\begin{tabular}{lc}
\hline
\textbf{Hyperparameter} & \textbf{Value} \\
\hline
LoRA rank & 64 \\
LoRA scaling parameter & 16 \\
Batch size & 128 \\
Learning rate & $1\times10^{-4}$ \\
Training epochs & 5 \\
LoRA dropout & 0.1 \\
Warm-up ratio & 0.03 \\
Maximum input length & 2048 tokens \\
\hline
\end{tabular}
\caption{LoRA fine-tuning hyperparameters used across the three base models.}
\label{tab:app:sft_hyperparameters}
\end{table}

Fine-tuning experiments are conducted using eight NVIDIA H20 or A800 GPUs. The same fine-tuning configuration is applied across the evaluated training datasets to maintain comparability.

\paragraph{Random Seeds and Result Reporting.}
For Mistral-7B-v0.3 and LLaMA-3.1-8B, we conduct two downstream SFT runs with seeds 42 and 47 for the Full Set, the two direct-fusion ablations, and our reconstructed dataset.

For these configurations, we report the mean and sample standard deviation across the two runs. The remaining baseline results in the corresponding main tables are based on single runs.

The reconstructed training dataset is held fixed across repeated SFT runs. Consequently, these experiments characterize downstream fine-tuning variability rather than variability arising from independently rerunning the complete reconstruction pipeline.

\subsection{Computational Cost and Validation-Set Requirements}
\label{appendix:computational_cost}

\paragraph{Cost Comparison.}
The proposed framework incurs computational costs during offline data construction, including external-LLM calls for quality assessment, domain analysis, candidate generation, validation, and failure-guided refinement.

To characterize its practical resource requirements, we compare it with representative baselines exhibiting different computational profiles. Table~\ref{tab:appendix_gpu_api_validation_analysis} reports the average API cost per output sample, measured GPU runtime, and task-specific validation-set requirements.

\begin{table*}[!htb]
    \setlength{\aboverulesep}{0pt}
    \setlength{\belowrulesep}{0pt}
    \caption{Representative comparison of average API cost per generated output sample, GPU runtime, and task-specific validation-set requirements. The methods have different computational profiles, so GPU usage and external-LLM API cost should be considered jointly rather than as interchangeable resource measures. GPU runtime refers to downstream fine-tuning and excludes external-LLM inference and other offline data-construction operations, which are accounted for separately where applicable.
}
    \centering
    \scalebox{0.9}{
        \begin{tabular}{lcccc}
        \hline
        \noalign{\vskip 1mm} 
        & LESS~\citeyearpar{xia2024less} & MathFusion~\citeyearpar{pei2025mathfusion} & Evol-Instruct~\citeyearpar{xu2024wizardlm} & Ours \\
        \hline
        Average API Cost (in USD) & 0 & 0.004 & 0.003 & 0.005 \\
        GPU Runtime (in GPU-hours) & 152.5 & 17 & 17 & 17 \\
        Validation Set & Required & Not Required & Not Required & Not Required \\
        \hline
        \end{tabular}
    }
    \label{tab:appendix_gpu_api_validation_analysis}
\end{table*}

LESS performs gradient-based selection without external-LLM API calls but requires task-specific validation data and additional GPU computation. In contrast, MathFusion, Evol-Instruct, and our framework rely on external-LLM processing during offline data construction.

Our framework incurs a higher average API cost per output sample than the two representative LLM-based augmentation methods, reflecting its multi-stage generation, validation, and repair procedure.

Importantly, these costs arise during offline dataset construction. Once the reconstructed training data have been generated, downstream fine-tuning and inference do not require rerunning the external reconstruction pipeline.

\paragraph{Validation-Set Requirements.}
Methods such as LESS and RDS+ require task-specific validation examples to estimate training-sample utility or representation-based relevance. In contrast, our reconstruction procedure operates on the low-scoring source samples and their semantic relationships without requiring a task-specific validation set during data construction.

Accordingly, the absence of a validation-set requirement in Table~\ref{tab:appendix_gpu_api_validation_analysis} refers to the construction procedure rather than the availability of benchmark test data used for downstream evaluation.

\paragraph{External-LLM Reproducibility.}
The quality-rating and reconstruction stages depend on an external LLM. Although the model configuration, prompt templates, and rule-guided control procedures are held fixed, repeated API calls may produce different intermediate outputs.

Moreover, the external scoring and generation model may affect both the composition of the low-scoring source set and the quality of reconstructed samples. Our experiments evaluate the framework under a fixed external-LLM configuration rather than systematically varying the scoring or generation model.

The repeated SFT experiments therefore characterize training variability given a fixed reconstructed dataset. They do not establish the variability of independently rerunning the entire rating, clustering, reconstruction, and fine-tuning pipeline.

\section{Additional Results and Ablation Studies}
\label{appendix:additional_results}

This section presents supplementary experimental results that complement the main evaluation in Section~4. We first report the complete Qwen2.5-7B comparison, followed by additional data-size ablations that separately examine intra-cluster and inter-cluster reconstruction and their combinations with the high-scoring source data. Finally, we analyze the distributions of curated quality scores before and after reconstruction.

\subsection{Additional Model Evaluations}
\label{appendix:emprical_observations}

\paragraph{Qwen2.5-7B Evaluation.}
Table~\ref{appendix:tab_main_table_qwen} reports the complete Qwen2.5-7B evaluation, including the reference configurations, data-selection methods, instruction-synthesis baselines, and reconstruction ablations described in Appendix~\ref{appendix:baseline}.

Unlike the five-benchmark evaluations for Mistral-7B-v0.3 and LLaMA-3.1-8B, this comparison excludes TruthfulQA and reports the average across MMLU, GSM8K, BBH, and TyDiQA.

Our reconstructed dataset achieves the highest reported four-benchmark average of 69.3, compared with 65.2 for the LQ Set, 68.0 for the HQ Set, and 68.8 for the Full Set. These correspond to improvements of 4.1, 1.3, and 0.5 points, respectively.

The largest improvement over the LQ Set occurs on TyDiQA, where our method achieves 69.3 compared with 55.4, an increase of 13.9 points. Improvements are also observed on GSM8K (+2.0) and BBH (+0.5), whereas MMLU decreases slightly by 0.1 points.

These results indicate that reconstructed supervision can improve average downstream performance while using substantially fewer training samples than the complete source pool. However, the gains are task-dependent and should not be interpreted as uniform improvements across all benchmarks.

\begin{table*}[!htb]
    \setlength{\aboverulesep}{0pt}
    \setlength{\belowrulesep}{0pt}
    
    \caption{Performance comparison across four downstream benchmarks using Qwen2.5-7B as the fine-tuning base model. Unless otherwise specified, the training dataset contains 54,888 samples. The best and second-best average scores are highlighted in \textbf{\color{red}{bold red}} and \textbf{bold black}, respectively. Values in parentheses for our method indicate the absolute performance differences relative to the LQ Set on each benchmark and the overall average.}

    \label{appendix:tab_main_table_qwen}
    \centering
    \scalebox{0.9}{
        \begin{tabular}{lccccc}
            \toprule
            \textbf{Dataset} & \makecell{\textbf{MMLU} \\ (EM $\uparrow$)} & \makecell{\textbf{GSM8K} \\ (EM $\uparrow$)} & \makecell{\textbf{BBH} \\ (EM $\uparrow$)} & \makecell{\textbf{TydiQA} \\ (1-Shot F1 $\uparrow$)} & \textbf{Average $\uparrow$}\\
            \midrule
            \multicolumn{6}{>{\columncolor{lightgreen}}c}{\textbf{Base Model: Qwen2.5-7B}} \\
            Vanilla Base Model & 71.8 & 83.5 & 58.1 & 25.3 & 59.7 \\
            LQ Set (123786) & 69.3 & 77.5 & 58.6 & 55.4 & 65.2 \\
            HQ Set (131247) & 72.2 & 79.0 & 60.4 & 60.2 & 68.0 \\
            Full Set (300932) & 72.0 & 78.0 & 59.8 & 65.2 & \textbf{68.8} \\
            \noalign{%
                \setlength{\arrayrulewidth}{1pt}
                \setlength{\dashlinedash}{5pt}
                \setlength{\dashlinegap}{2pt}
            }%
            \cdashline{1-6} 
            Completion Length & 67.2 & 75.5 & 57.9 & 62.7 & 65.8 \\
            $\mathrm{KNN}_{10}$ & 70.4 & 77.5 & 57.7 & 63.2 & 67.2 \\
            Perplexity & 70.1 & 76.0 & 52.9 & 63.7 & 65.7 \\
            Random Selection & 69.3 & 75.5 & 57.3 & 65.4 & 66.9 \\
            \noalign{%
                \setlength{\arrayrulewidth}{1pt}
                \setlength{\dashlinedash}{5pt}
                \setlength{\dashlinegap}{2pt}
            }%
            \cdashline{1-6} 
            AlpaGasus (Random) & 65.8 & 74.0 & 58.1 & 57.6 & 63.9 \\
            IFD & 63.9 & 68.5 & 53.2 & 52.1 & 59.4 \\
            Superfiltering & 68.3 & 76.0 & 55.0 & 59.4 & 64.7 \\
            DEITA & 68.6 & 76.0 & 59.4 & 59.5 & 65.9 \\
            DEITA (Our Curated Score) & 68.0 & 73.0 & 57.6 & 62.6 & 65.3 \\
            RDS+ & 69.1 & 78.5 & 55.4 & 55.1 & 64.5 \\
            RDS+ (Best) & 69.1 & 78.5 & 57.9 & 55.1 & 65.2 \\
            $\mathrm{DS}^{2}$ & 67.2 & 79.5 & 58.1 & 61.6 & 66.6 \\
            LESS & 69.5 & 75.5 & 58.1 & 63.8 & 66.7 \\
            \noalign{%
                \setlength{\arrayrulewidth}{1pt}
                \setlength{\dashlinedash}{5pt}
                \setlength{\dashlinegap}{2pt}
            }%
            \cdashline{1-6} 
            MathFusion (conditional) & 67.9 & 70.5 & 59.3 & 55.4 & 63.3 \\
            MathFusion (parallel) & 68.0 & 65.5 & 61.3 & 41.3 & 59.0 \\
            MathFusion (sequential) & 68.3 & 64.0 & 58.6 & 52.3 & 60.8 \\
            MathFusion & 68.0 & 74.0 & 62.3 & 54.7 & 64.8 \\
            Evol-Instruct & 69.0 & 79.5 & 62.4 & 60.2 & 67.8 \\
            Self-Instruct & 65.3 & 75.0 & 57.0 & 61.3 & 64.7 \\
            \cdashline{1-6} 
            1-to-1 Rewriting/ Enhancement & 64.8 & 82.0 & 59.8 & 40.9 & 61.9 \\
            Direct Corpora Fusion Without Step2\&3 & 63.4 & 80.5 & 58.7 & 53.1 & 63.9 \\
            Direct Corpora Fusion Without Step3 & 65.0 & 79.5 & 58.2 & 47.0 & 62.4 \\
            \cdashline{1-6} 
            Ours & 69.2 (-0.1) & 79.5 (+2.0) & 59.1 (+0.5) & 69.3 (+13.9) & \textbf{\textcolor{red}{69.3 (+4.1)}} \\
            \bottomrule
        \end{tabular}
    }
\end{table*}

\subsection{Supplementary Data-Size Ablation}
\label{appendix:ablation}

\paragraph{Experimental Configurations.}
The main-text data-size ablation reports two-seed results for 20\%, 60\%, and 100\% of our complete reconstructed dataset. Here, we additionally provide the original single-run results across all five data proportions (20\%, 40\%, 60\%, 80\%, and 100\%) and extend the analysis by separately evaluating intra-cluster and inter-cluster reconstruction and their combinations with the HQ Set.

The complete reconstructed dataset contains 54,888 samples, comprising 15,488 intra-cluster and 39,400 inter-cluster reconstructed samples.

Beyond the complete-dataset ablation, we consider two additional groups of experiments.

\begin{itemize}[leftmargin=*]

    \item \textbf{Reconstruction-only configurations.}
    We independently sample different proportions of the intra-cluster and inter-cluster reconstructed datasets.

    These configurations are denoted as \textit{Intra-$x$\%} and \textit{Inter-$x$\%}, where $x$\% refers to the proportion of the corresponding reconstruction-generated dataset.

    Since the two datasets contain different numbers of samples, identical percentages do not imply identical training-data sizes.

    These experiments are evaluated on MMLU, GSM8K, BBH, and TyDiQA, with averages computed over the four benchmarks.
    
    \item \textbf{HQ-augmented configurations.}
    We additionally investigate whether reconstructed data complements the original high-scoring source samples.
    
    The three configurations are \textit{HQ+Intra-$x$\%}, \textit{HQ+Inter-$x$\%}, and \textit{HQ+Ours-$x$\%}.
    
    Each configuration retains the complete HQ Set of 131,247 samples and adds $x$\% of the intra-cluster reconstructed dataset (15,488 samples), the inter-cluster reconstructed dataset (39,400 samples), or our complete reconstructed dataset (54,888 samples), respectively.
    
    Consequently, the total training-data size increases with the augmentation proportion and differs across the three configurations.
    
\end{itemize}

The HQ-augmentation experiments, including those for Qwen2.5-7B, are evaluated across five benchmarks, including TruthfulQA. Their averages should therefore be interpreted separately from the four-benchmark Qwen comparison in Appendix~\ref{appendix:emprical_observations} and the four-benchmark reconstruction-only ablations.

\paragraph{Single-Run Data-Size Results.}
Table~\ref{tab:ablation_ours_mistral_llama} reports the original seed-42 results for all five proportions of our complete reconstructed dataset. These results provide the benchmark-level performance for the additional 40\% and 80\% configurations, complementing the two-seed summaries for 20\%, 60\%, and 100\% in main-text Table~3.

\paragraph{Intra-Cluster Reconstruction: Performance with Partial Data.}
Table~\ref{appendix:ablation_intra_qwen_mistral_llama} reports the intra-cluster reconstruction results for Qwen2.5-7B, Mistral-7B-v0.3, and LLaMA-3.1-8B.

A notable finding is that even relatively small subsets of intra-cluster reconstructed data can provide competitive downstream performance.

For Qwen2.5-7B, Intra-20\% and Intra-60\% both achieve a four-benchmark average of 70.9, exceeding the HQ Set (68.0) and Full Set (68.8) reference configurations.

For Mistral-7B-v0.3, increasing the intra-cluster proportion from 20\% to 60\% improves the average from 52.2 to 53.6. The latter exceeds both the HQ Set (53.1) and Full Set (52.4) under the corresponding four-benchmark evaluation.

For LLaMA-3.1-8B, Intra-60\% achieves 60.0, exceeding the HQ Set average of 59.5 but remaining below the Full Set average of 61.6.

These findings demonstrate that partial intra-cluster reconstructed datasets can provide useful supervision across all three base models. However, performance is not strictly monotonic with data volume. For example, the Qwen averages at 20\%, 40\%, and 60\% are 70.9, 69.3, and 70.9, respectively.

\paragraph{Inter-Cluster Reconstruction: Effect of Data Volume.}
Table~\ref{appendix:ablation_inter_mistral_llama} presents the inter-cluster reconstruction experiments for Mistral-7B-v0.3 and LLaMA-3.1-8B.

Under Mistral-7B-v0.3, the four-benchmark average increases from 52.1 at Inter-20\% to 54.6 at Inter-100\%. The latter exceeds both the HQ Set (53.1) and Full Set (52.4).

For LLaMA-3.1-8B, the corresponding average increases from 57.4 at Inter-20\% to 59.1 at Inter-100\%, although intermediate proportions exhibit small fluctuations. Unlike Mistral, the full inter-cluster dataset does not exceed the HQ Set (59.5) or Full Set (61.6) reference configurations.

These results suggest that inter-cluster reconstruction can provide substantial gains, particularly for Mistral, but its effectiveness varies across fine-tuning models.

Moreover, because the intra-cluster and inter-cluster datasets differ in size and composition, these experiments do not isolate source heterogeneity independently of training-data volume.

\paragraph{Complementarity with High-Scoring Data: LLaMA-3.1-8B.}
Table~\ref{appendix:ablation_hq+inter_llama} evaluates the combination of the complete HQ Set with different proportions of inter-cluster reconstructed data under LLaMA-3.1-8B.

Adding 40\% inter-cluster data produces an average of 56.0, matching the HQ-only baseline. Increasing the proportion to 60\% and 100\% raises the average to 57.8 and 58.1, respectively.

The HQ+Inter-100\% configuration therefore exceeds the HQ Set (56.0) by 2.1 points and the Full Set (57.7) by 0.4 points.

This result provides evidence that inter-cluster reconstructed samples can contribute complementary supervision beyond the original high-scoring data in this experimental setting.

\paragraph{HQ-Augmented Reconstruction under Qwen2.5-7B.}
Tables~\ref{appendix:ablation_hq+intra_qwen} and~\ref{appendix:ablation_hq+inter_mistral_qwen} present the Qwen2.5-7B results obtained by augmenting the HQ Set with intra-cluster and inter-cluster reconstructed samples, respectively.

For intra-cluster augmentation, the average improves from 60.6 at HQ+Intra-40\% to 63.5 at HQ+Intra-100\%. The latter exceeds both the HQ Set (62.0) and Full Set (63.3).

For inter-cluster augmentation, the highest reported average is 63.8 at HQ+Inter-80\%, exceeding the HQ Set and Full Set by 1.8 and 0.5 points, respectively.

These results show that both reconstruction mechanisms can improve the performance of HQ-only fine-tuning in selected configurations.

Although the best reported inter-cluster configuration achieves a higher average than the best reported intra-cluster configuration, these settings use different proportions and numbers of added samples. Therefore, this comparison does not establish that inter-cluster reconstruction is intrinsically superior to intra-cluster reconstruction.

\paragraph{Combining Both Reconstruction Mechanisms with the HQ Set.}
Table~\ref{appendix:ablation_hq+entp_mistral_llama_qwen} reports the HQ+Ours configurations, where the complete HQ Set is augmented with different proportions of our reconstructed dataset, comprising both intra-cluster and inter-cluster samples.

For Mistral-7B-v0.3, HQ+Ours-100\% achieves a five-benchmark average of 51.2, exceeding both the HQ Set (50.3) and Full Set (50.6).

For LLaMA-3.1-8B, HQ+Ours-100\% reaches 57.7, outperforming the HQ Set (56.0) and matching the Full Set (57.7).

For Qwen2.5-7B, HQ+Ours-100\% achieves 63.1, exceeding the HQ Set (62.0) but remaining slightly below the Full Set (63.3).

These results suggest that reconstructed data can complement high-scoring source data across different base models. However, the magnitude of improvement depends on the model and augmentation configuration, and combining both reconstruction mechanisms does not guarantee superiority over all reference settings.

\paragraph{Summary and Scope of the Supplementary Ablations.}
Taken together, Tables~\ref{appendix:ablation_intra_qwen_mistral_llama}--\ref{appendix:ablation_hq+entp_mistral_llama_qwen} provide three additional observations.

First, useful downstream supervision can be obtained from partial reconstructed datasets, sometimes exceeding the performance of substantially larger raw-data configurations.

Second, reconstruction-data volume generally affects downstream performance, but the relationship is not uniformly monotonic across all model and dataset configurations.

Third, adding reconstructed samples to the HQ Set can improve performance, indicating potential complementarity between original high-scoring supervision and reconstructed examples.

These results should not be interpreted as universal scaling laws or as evidence that a particular reconstruction mechanism is always superior. The comparisons involve different training-data sizes, source compositions, and evaluation settings, and the supplementary experiments do not independently isolate every internal reconstruction component.

\subsection{Curated Score Distributions}
\label{appendix:score_distributions}

In addition to downstream performance, we examine changes in the distribution of curated overall quality scores following reconstruction.

The LQ Set contains source samples with curated overall scores of 0--2. The reconstructed samples are subsequently assessed using the adopted quality-rating and score-curation procedure to obtain corresponding quality estimates.

\begin{figure*}[t]
    \centering
    \includegraphics[
        width=\textwidth,
        keepaspectratio
    ]{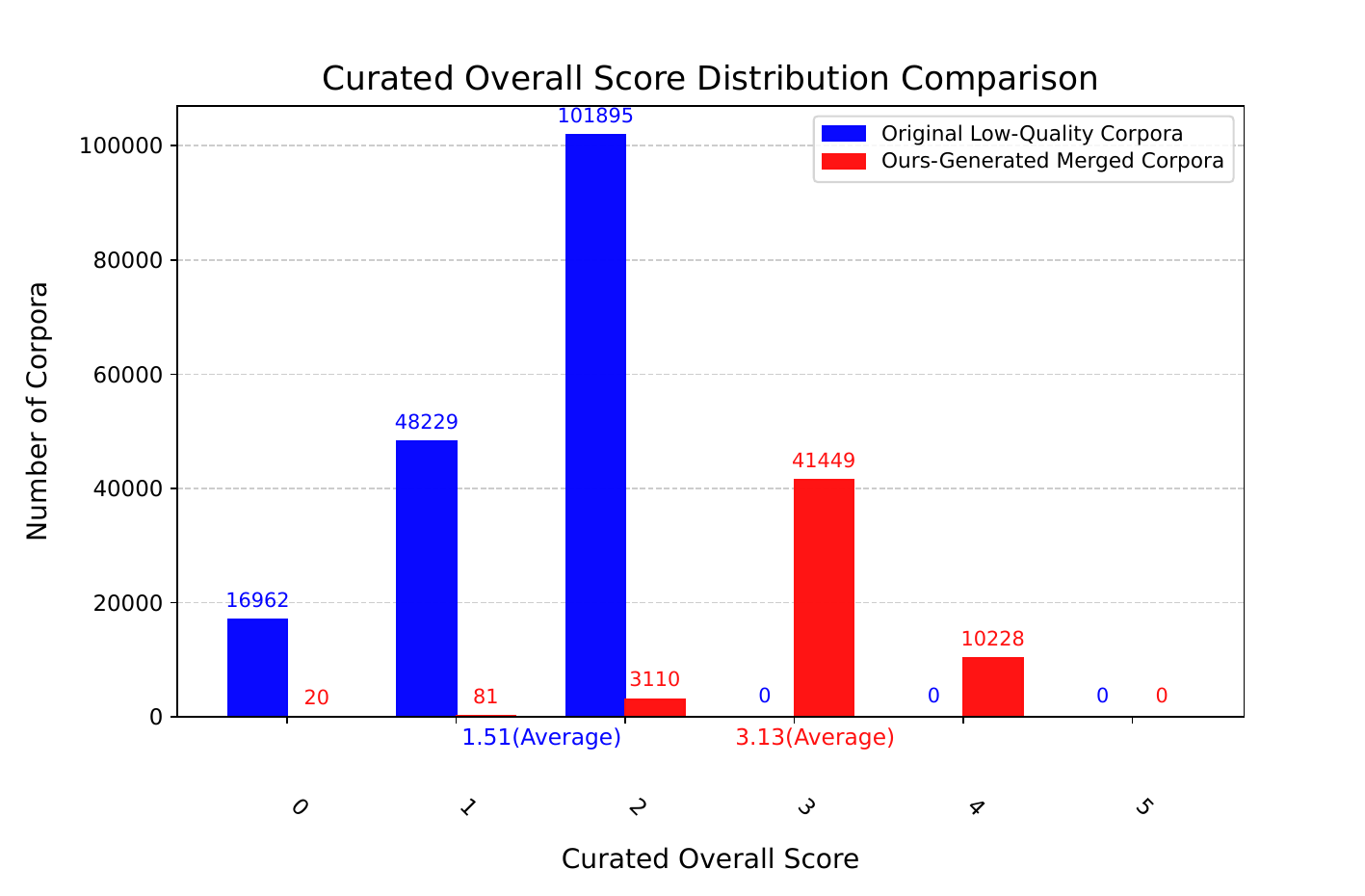}
    
    \caption{Comparison of curated overall quality-score distributions between the original low-scoring source pool before structure-dependence filtering and the reconstructed dataset. Scores range from 0 to 5, with larger values indicating higher model-estimated quality.}

    \label{fig:app:score_distributions}
\end{figure*}

Figure~\ref{fig:app:score_distributions} compares the curated quality-score distributions of the original low-scoring source samples and the reconstructed dataset.

The reconstructed dataset exhibits a shift toward higher score categories, with a larger concentration of samples around scores 2--4 compared with the original low-scoring source distribution.

This shift is consistent with the intended objective of reconstructing more complete and informative instruction-response pairs from low-scoring source data.

However, the curated scores are model-based estimates rather than independently verified quality annotations. Consequently, the distributional changes do not, by themselves, establish factual correctness or faithful integration of source information.

These complementary aspects of reconstruction quality are examined through the source-grounded manual audit reported in Appendix~F.

\section{Manual Quality Audit and Qualitative Examples}
\label{appendix:manual_quality_audit}

This section evaluates reconstruction quality beyond the internal curated quality scores discussed in Appendix~\ref{appendix:score_distributions} and the downstream benchmark results in Section~4. We report a small-scale, source-grounded manual audit, followed by a verified successful reconstruction and two previously showcased examples that were reassessed as failures. The qualitative cases illustrate the evaluation criteria; the aggregate findings are based on the randomly sampled audit rather than on selected examples.

\subsection{Audit Scope and Annotation Protocol}
\label{appendix:manual_audit_protocol}

\paragraph{Sample and evaluation unit.}
We randomly sampled 30 reconstructed instruction--response samples and examined their generated content alongside the original source samples from which they were constructed. The sample comprises 13 pairwise cross-cluster reconstructions (termed \emph{inter-class fusion} in the original audit) and 17 multi-source intra-cluster reconstructions. Each reconstructed sample, rather than each underlying source item, serves as one evaluation unit.

\paragraph{Manual Annotation and Consensus.}
Each sampled reconstruction was evaluated by multiple human annotators who were not personally acquainted with one another. The annotators assessed each sample according to the five binary criteria defined in Appendix~\ref{appendix:manual_audit_criteria}. For each sample and criterion, the reported binary label was determined by strict-majority consensus, requiring agreement from more than half of the annotators. The resulting five-dimensional quality vector consists of these majority-agreed criterion labels.

\paragraph{Evaluation objective.}
The audit examines whether reconstruction meaningfully preserves and integrates source information into a coherent, usable instruction--response sample. It is separate from the LLM-based quality-rating and score-curation procedure used for the distributional analysis. It is not an independent verification of the external factual correctness of the original source materials.

We distinguish \emph{inherited source information}, including potentially inaccurate or fictional content already present in an input, from \emph{reconstruction-introduced information}, such as unsupported factual claims, entities, or relationships introduced during generation. An inherited claim is not automatically treated as a reconstruction failure merely because its external truth has not been established. By contrast, a newly introduced claim that is unsupported by the input sources, or a contradiction of source content, constitutes a source-faithfulness failure.

\subsection{Manual Evaluation Criteria}
\label{appendix:manual_audit_criteria}

Each reconstructed sample is evaluated using the following five binary criteria. A score of 1 indicates that the condition is satisfied, and 0 indicates failure.

\begin{enumerate}
    \item \textbf{Source integration (SI).} The generated sample meaningfully incorporates the core information from its source samples, instead of omitting major distinct inputs, simply concatenating them, or substantially drifting from the original content and task intent.

    \item \textbf{Context completeness (CC).} The generated question or instruction includes sufficient context to be understood and answered without relying on essential information that has been omitted.

    \item \textbf{Question clarity (QC).} The generated question is well formed, unambiguous, and answerable.

    \item \textbf{Answer faithfulness and consistency (AF).} The response addresses the generated question, remains consistent with the provided sources and generated context, and does not add unsupported or contradictory factual information.

    \item \textbf{Overall transformation usability (OU).} The reconstructed instruction--response sample is coherent, source-grounded, and usable for downstream supervised fine-tuning without major structural, task-level, or consistency-related revision.
\end{enumerate}

The five labels are recorded in the order \textbf{[SI, CC, QC, AF, OU]}. A sample is counted as passing \emph{all five criteria} only if each label equals 1. In the case of highly redundant inputs, source integration does not require repeating every individual sentence or example; preserving the shared core content and task intent is sufficient. However, omission of major, distinct source concepts is counted as an integration failure. These manually assigned criteria are used to assess the resulting samples; they should not be confused with the automatic ICD/FAC validation checks that guide reconstruction in Appendix~\ref{appendix:step3}.

\subsection{Aggregate Audit Results}
\label{appendix:manual_audit_results}

\paragraph{Criterion-level outcomes.}
Table~\ref{tab:app:manual_audit_results} summarizes the five criterion-wise pass rates and the proportion passing all five conditions. Context completeness and question clarity each pass for 28 of the 30 samples (93.3\%), whereas only 20 (66.7\%) pass source integration and 18 (60.0\%) pass answer faithfulness and consistency. Overall transformation usability and the conjunction of all five conditions each pass for 14 samples (46.7\%).

\begin{table}[t]
\centering
\small
\setlength{\tabcolsep}{4pt}
\renewcommand{\arraystretch}{1.12}
\begin{tabular}{@{}p{0.59\columnwidth}cc@{}}
\toprule
\textbf{Manual quality criterion} & \textbf{Passed} & \textbf{Rate} \\
\midrule
Source integration & 20/30 & 66.7\% \\
Context completeness & 28/30 & 93.3\% \\
Question clarity & 28/30 & 93.3\% \\
Answer faithfulness and consistency & 18/30 & 60.0\% \\
Overall transformation usability & 14/30 & 46.7\% \\
\midrule
All five criteria & 14/30 & 46.7\% \\
\bottomrule
\end{tabular}
\caption{Manual source-grounded quality audit of 30 randomly sampled reconstructed instruction--response samples. Each criterion is assigned a binary pass/fail label. The last row requires all five criteria to pass simultaneously.}
\label{tab:app:manual_audit_results}
\end{table}

These results distinguish surface-level formulation quality from successful source-grounded transformation. A fluent, answerable question can still omit important source information, introduce unsupported details, or yield an inconsistent answer. In particular, the 46.7\% all-criteria pass rate shows that the adopted automatic reconstruction checks do not guarantee that every final sample satisfies the stricter manual standards.

\paragraph{Source integration by reconstruction setting.}
Table~\ref{tab:app:manual_audit_by_setting} separates source-integration outcomes by the two sampled reconstruction settings. Of the 13 pairwise cross-cluster cases, 11 pass source integration (84.6\%). Of the 17 multi-source intra-cluster cases, 9 pass (52.9\%). These counts sum to the 20 source-integration passes in Table~\ref{tab:app:manual_audit_results}.

\begin{table}[t]
\centering
\small
\setlength{\tabcolsep}{3pt}
\renewcommand{\arraystretch}{1.1}
\begin{tabular}{@{}p{0.51\columnwidth}ccc@{}}
\toprule
\textbf{Reconstruction setting} & \textbf{Audited} & \textbf{Passed} & \textbf{Rate} \\
\midrule
Pairwise cross-cluster & 13 & 11/13 & 84.6\% \\
Multi-source intra-cluster & 17 & 9/17 & 52.9\% \\
\bottomrule
\end{tabular}
\caption{Source-integration pass rates by reconstruction setting within the 30-sample manual audit. The comparison is descriptive and is not controlled for source count, topic, or input complexity.}
\label{tab:app:manual_audit_by_setting}
\end{table}

The lower observed integration rate for multi-source intra-cluster reconstruction is consistent with the additional difficulty of condensing several potentially distinct inputs into one instruction--response pair. However, the settings differ in both the number and composition of source inputs. Given the small sample sizes, this difference should not be interpreted as a controlled causal effect of reconstruction type or as a statistically established population-level difference.

\paragraph{Observed failure modes.}
The manual review identifies several types of failures: omitting a substantial part of the supplied source information; introducing unsupported claims or connections; drifting from the original task; generating internally inconsistent or incomplete answers; and mechanically combining unrelated concepts without a coherent semantic relationship. These failure modes explain why high rates of question clarity and context completeness do not translate into an equally high rate of usable, source-grounded reconstructions.

\subsection{Source-Grounded Qualitative Examples}
\label{appendix:manual_audit_examples}

We next illustrate the five criteria with one successful reconstruction drawn from the random audit and two earlier showcased reconstructions that were subsequently reassessed using the same criteria. The latter two examples are presented as qualitative failure cases and are not asserted to be part of the random 30-sample audit. Table~\ref{tab:app:manual_audit_example_labels} summarizes their manual annotations.

\begin{table}[t]
\centering
\footnotesize
\renewcommand{\arraystretch}{1.15}
\resizebox{\columnwidth}{!}{%
\begin{tabular}{@{}lccccc@{}}
\toprule
\textbf{Example} & \textbf{SI} & \textbf{CC} & \textbf{QC} & \textbf{AF} & \textbf{OU} \\
\midrule
Sick-leave email & 1 & 1 & 1 & 1 & 1 \\
La Paz / botany & 0 & 0 & 1 & 0 & 0 \\
Apps as companions & 0 & 0 & 0 & 0 & 0 \\
\bottomrule
\end{tabular}%
}
\caption{Case-level manual quality judgments for the illustrated reconstructions. The sick-leave example belongs to the random audit, while the other two are reassessed failure cases. SI: Source Integration; CC: Context Completeness; QC: Question Clarity; AF: Answer Faithfulness and Consistency; OU: Overall Transformation Usability.}
\label{tab:app:manual_audit_example_labels}
\end{table}

\subsubsection{Verified Success: Sick-Leave Email}
\label{appendix:manual_success_case}

The following reconstruction was among the 30 randomly sampled outputs and passed all five manual criteria. The two source samples request essentially the same task: writing an email to colleagues about a week-long sickness-related absence. We reproduce their actual wording, including linguistic errors in the original inputs, to make the source-to-output relationship inspectable.

\paragraph{Source sample A.}
\begin{quote}\small
\noindent\textbf{User:} Can you write me an polite E-Mail to inform my colleagues that I'll be sick for the entire week, please.

\smallskip
\noindent\textbf{Assistant:} Dear colleagues

I've been seek for the entire week and I wasn't able to work on our project. Please understand my issue and forgive me.

[Your name]
\end{quote}

\paragraph{Source sample B.}
\begin{quote}\small
\noindent\textbf{User:} Can you write me an polite E-Mail to inform my colleagues that I'll be sick for the entire week, please.

\smallskip
\noindent\textbf{Assistant:} Dear colleagues,

I hope this email finds you all well. Unfortunately, I wanted to let you know that due to unexpected circumstances, I will be unable to come to work this week as I am feeling unwell.

I apologize for any inconvenience this may cause, and I will do my best to catch up on any missed work when I return. In the meantime, please feel free to reach out to me if there are any urgent matters that require my attention and I will do my best to respond as soon as I am able.

Thank you for your understanding and support during this time. I look forward to rejoining the team as soon as I am able.

Best regards,

[Your Name]
\end{quote}

\paragraph{Reconstructed sample.}
\begin{quote}\small
\noindent\textbf{User:} Dear colleagues,

I hope this message finds you well. I wanted to inform you that I will be on sick leave for the entire week due to health issues. I apologize for any disruption this may cause to our ongoing projects. I appreciate your understanding and support during this time.

In the meantime, I will do my best to stay updated on our work. If there are any urgent matters, please feel free to reach out to me. I look forward to rejoining you all soon.

Best regards,

[Your Name]

What should I do if I need to provide further updates while I'm away?

\smallskip
\noindent\textbf{Assistant:} If you need to provide further updates while you're away, consider setting up an out-of-office email response that informs colleagues of your absence and provides alternative contacts for urgent matters.
\end{quote}

\paragraph{Source-grounded assessment.}
The reconstruction preserves the shared central task and salient content: notifying colleagues of a week-long absence, addressing possible disruption to ongoing work, thanking colleagues for their understanding, and handling urgent matters during the absence. Because the two source samples are highly redundant, retaining their overlapping core information does not require repeating every sentence. The additional follow-up question concerns communication during the same absence, and its answer offers a relevant general action without inventing specific facts about the person or workplace. Accordingly, the recorded quality vector is $[1,1,1,1,1]$.

\subsubsection{Failure: Unsupported Bridging in the La Paz Example}
\label{appendix:manual_failure_lapaz}

\begin{figure*}[p]
    \centering
    \includegraphics[
        width=\textwidth,
        height=0.78\textheight,
        keepaspectratio
    ]{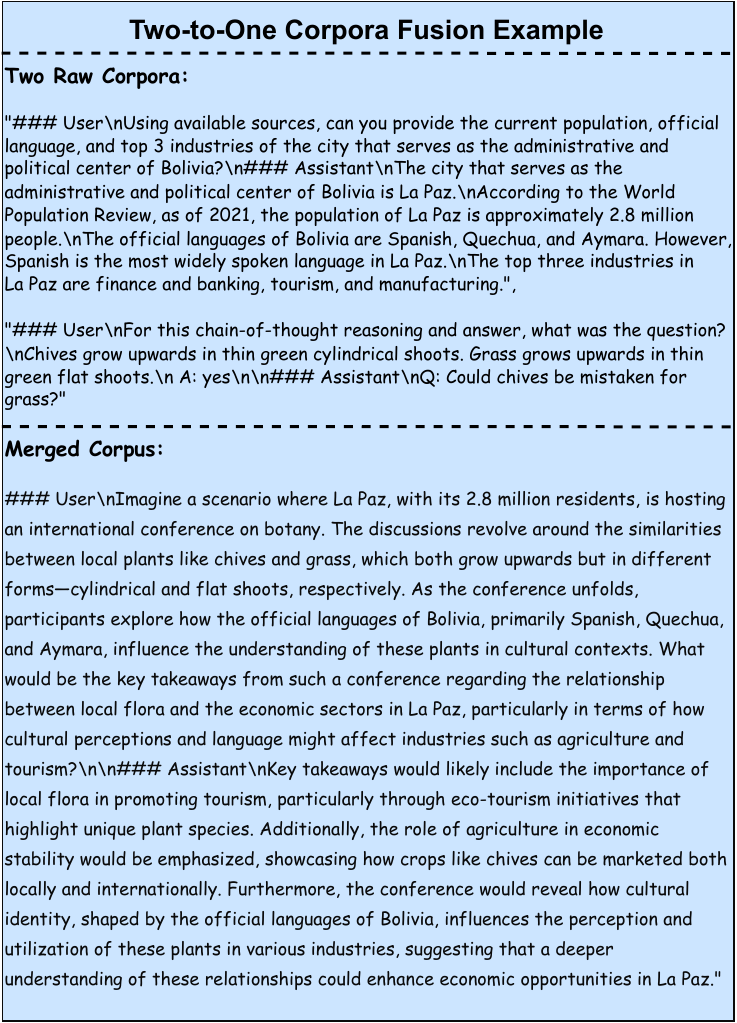}

    \caption{Reassessed reconstruction failure: unsupported bridging in the La Paz example. The generated botanical-conference scenario introduces information not supported by the source samples and omits major source concepts. The manual quality vector is $[0,0,1,0,0]$.}
    \label{fig:app:lapaz_failure}
\end{figure*}

As illustrated in Figure~\ref{fig:app:lapaz_failure}, the first reassessed example combines a source about La Paz, including its population, languages, and principal industries, with a separate source question about whether chives could be mistaken for grass. The generated sample introduces an international botany conference and additional claims about local flora, agriculture, eco-tourism, crop marketing, and the influence of language on plant utilization. These connecting elements are not supported by the original source inputs. The output also omits major distinct information about La Paz, including finance and banking and manufacturing.

The example is grammatically understandable, but its cross-topic narrative is built on unsupported additions rather than meaningful preservation of the sources. The reassessed quality vector is $[0,0,1,0,0]$: it passes question clarity but fails source integration, context completeness, answer faithfulness and consistency, and overall usability. Its failure illustrates why apparent narrative coherence does not by itself demonstrate source-grounded reconstruction.

\subsubsection{Failure: Task Drift in the App-Review Example}
\label{appendix:manual_failure_apps}

\begin{figure*}[p]
    \centering
    \includegraphics[
        width=\textwidth,
        height=0.78\textheight,
        keepaspectratio
    ]{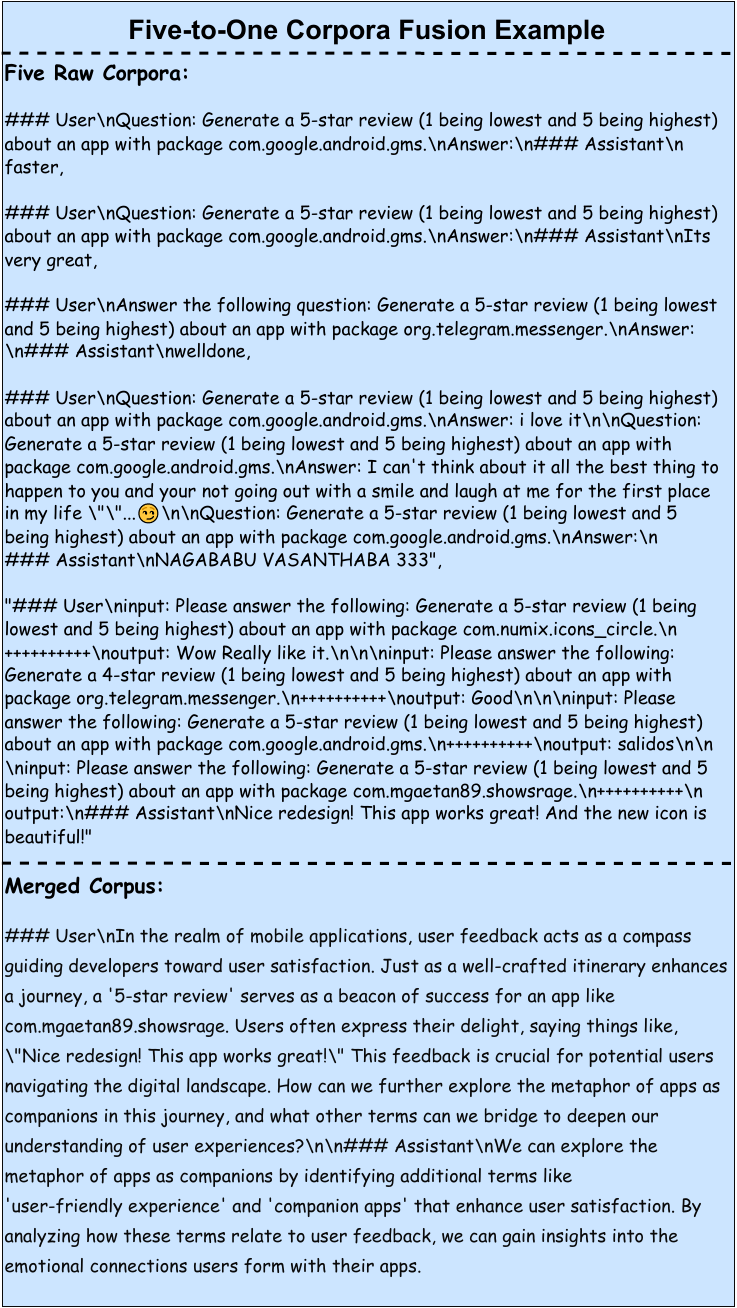}

    \caption{Reassessed reconstruction failure: task drift in the app-review example. The generated discussion of apps as companions does not preserve the original review-writing task and introduces unsupported abstractions. The manual quality vector is $[0,0,0,0,0]$.}
    \label{fig:app:apps_failure}
\end{figure*}

As illustrated in Figure~\ref{fig:app:apps_failure}, the second reassessed example originates from samples that request short application reviews conditioned on a specified application package and target star rating. Instead of preserving this concrete review-writing task, the reconstruction shifts to an abstract discussion of apps as companions, digital journeys, and emotional connections between users and applications. These ideas are not grounded in the original review-generation requests, and the resulting response no longer performs the intended source task.

The reassessed quality vector is $[0,0,0,0,0]$. This case illustrates severe task drift and unsupported abstraction: creative or fluent language is insufficient when the generated instruction--response pair loses the core purpose of its source samples.

\subsection{Illustrative Contrast with Direct One-Step Fusion}
\label{appendix:direct_fusion_qualitative_example}

A separate example from the direct one-step fusion baseline described in Appendix~\ref{appendix:baseline} illustrates why linguistic concatenation alone is not a sufficient reconstruction strategy. In that example, one source requests a list of leading semiconductor companies, while the other asks which of a black mamba and a black sea bass is a fish. The baseline-generated instruction largely places the two unrelated source questions next to each other, but its output answers only the fish question (\emph{Black Sea Bass}) and does not address the semiconductor request. This is an illustrative baseline output, not a sample from our 30-output audit and not an additional manually scored result. It demonstrates the risk of preserving superficial source text without generating a coherent, fully answered combined task.

\subsection{Interpretation and Limitations}
\label{appendix:manual_audit_limitations}

The audit provides evidence about dimensions that are not directly established by the internal curated quality-score distribution or downstream benchmark scores. The contrast between high question-level pass rates and the substantially lower source-faithfulness and all-criteria pass rates indicates that well-formed reconstruction does not guarantee faithful source integration. The selected examples make this distinction concrete, including a successful transformation of redundant inputs and two failures caused by unsupported bridging or task drift.

The evaluation has important limitations. First, it covers only 30 samples out of 54,888 reconstructed examples; criterion-wise rates and differences between reconstruction settings are descriptive rather than precise estimates for the full dataset. Second, the audit does not independently verify the truth of all inherited source claims. Third, although each reported binary label is supported by strict-majority consensus among multiple human annotators, we do not report a chance-corrected inter-annotator agreement statistic.

The complete sample-level annotation records, including the reconstruction setting, number of source samples, five binary labels, annotation notes, and the distinction between inherited and reconstruction-introduced issues, are planned for release following acceptance.

Taken together, the findings support a narrower interpretation than would be implied by fluency or model-based score improvements alone: rule-guided reconstruction can produce useful source-grounded instruction data, but it does not eliminate failures of information preservation, grounding, or task consistency. These limitations motivate stronger source-aware validation and broader human evaluation in future work.

\begin{table*}[!htb]
    \setlength{\aboverulesep}{0pt}
    \setlength{\belowrulesep}{0pt}
    \caption{
    \rev{
        The fine-tuning base models are Mistral-7B-v0.3 and Llama-3.1-8B. We report single-run results for 20\%, 40\%, 60\%, 80\%, and 100\% of the reconstructed dataset, with seed 42 used for these configurations. The Average is computed over the four reported benchmarks. Best and second-best Average scores for each base model are highlighted in \textbf{\textcolor{red}{bold red}} and \textbf{bold black}, respectively. Parenthesized values indicate absolute Average-score differences relative to the corresponding single-run LQ Set baseline.
    }
    }
    \label{tab:ablation_ours_mistral_llama}
    \centering
    \scalebox{1}{
    \begin{tabular}{lccccc}
        \toprule
        \textbf{Dataset} & \makecell{\textbf{MMLU} \\ (EM $\uparrow$)} & \makecell{\textbf{GSM8K} \\ (EM $\uparrow$)} & \makecell{\textbf{BBH} \\ (EM $\uparrow$)} & \makecell{\textbf{TydiQA} \\ (1-Shot F1 $\uparrow$)} & \textbf{Average $\uparrow$}\\
        \midrule
        \multicolumn{6}{>{\columncolor{lightgreen}}c}{\textbf{Base Model: Mistral-7B-v0.3}} \\
        LQ Set (123786) &  47.5 & 43.5 & 52.7 & 41.1 & \rev{46.2} \\
        HQ Set (131247) & 58.4 & 46.0 & 55.5 & 52.5 & 53.1 \\
        Full Set (300932) & 60.0 & 43.5 & 52.5 & 53.4 & 52.4 \\
        \noalign{%
            \setlength{\arrayrulewidth}{1pt}
            \setlength{\dashlinedash}{5pt}
            \setlength{\dashlinegap}{2pt}
        }%
        \cdashline{1-6} 
        Ours-20\% & 59.3 & 41.5 & 54.6 & 55.4 & 52.7 (+6.5) \\
        Ours-40\% & 58.9 & 42.0 & 50.5 & 56.8 & 52.1 (+5.9) \\
        Ours-60\% & 59.1 & 45.5 & 52.3 & 56.2 & 53.3 (+7.1) \\
        Ours-80\% & 58.5 & 44.0 & 53.8 & 57.4 & \textbf{53.4 (+7.2)} \\[1pt]
        Ours-100\% & 58.6 & 44.0 & 53.8 & 58.3 & \textbf{\textcolor{red}{53.7 (+7.5)}} \\
        \multicolumn{6}{>{\columncolor{lightgreen}}c}{\textbf{Base Model: Llama-3.1-8B}} \\
        LQ Set (123786) & 52.7 & 57.0 & 61.0 & \rev{43.9} & \rev{53.7} \\
        HQ Set (131247) & 62.3 & 57.5 & 59.3 & 58.9 & 59.5 \\
        Full Set (300932) & 63.5 & 61.0 & 59.1 & 62.8 & \textbf{\textcolor{red}{61.6}} \\
        \noalign{%
            \setlength{\arrayrulewidth}{1pt}
            \setlength{\dashlinedash}{5pt}
            \setlength{\dashlinegap}{2pt}
        }%
        \cdashline{1-6} 
        Ours-20\% & 63.9 & 57.5 & 61.5 & 52.5 & 58.9 (+5.2) \\
        Ours-40\% & 62.3 & 56.5 & 58.0 & 56.5 & 58.3 (+4.6)\\
        Ours-60\% & 62.3 & 57.5 & 60.1 & 57.2 & 59.3 (+5.6)\\
        Ours-80\% & 62.0 & 56.0 & 61.5 & 57.8 & 59.3 (+5.6)\\
        Ours-100\% & 61.7 & 54.5 & 60.7 & 61.3 & \textbf{59.6 (+5.9)} \\[1pt]
    \bottomrule
    \end{tabular}
    }
\end{table*}

\begin{table*}[ht]
    \setlength{\aboverulesep}{0pt}
    \setlength{\belowrulesep}{0pt}
    
    \caption{Data-size ablation of intra-cluster reconstructed samples across four downstream benchmarks using Qwen2.5-7B, Mistral-7B-v0.3, and LLaMA-3.1-8B as fine-tuning base models. Intra-$x$\% denotes the use of $x$\% of the 15,488 intra-cluster reconstructed samples. The best and second-best average scores within each base-model setting are highlighted in \textbf{\textcolor{red}{bold red}} and \textbf{bold black}, respectively. Values in parentheses indicate the absolute   differences in average performance relative to the corresponding LQ Set.}

    \label{appendix:ablation_intra_qwen_mistral_llama}
    \centering
    \begin{tabular}{lccccc}
        \toprule
        \textbf{Dataset} & \makecell{\textbf{MMLU} \\ (EM $\uparrow$)} & \makecell{\textbf{GSM8K} \\ (EM $\uparrow$)} & \makecell{\textbf{BBH} \\ (EM $\uparrow$)} & \makecell{\textbf{TydiQA} \\ (1-Shot F1 $\uparrow$)} & \textbf{Average $\uparrow$}\\
        \midrule
        \multicolumn{6}{>{\columncolor{lightgreen}}c}{\textbf{Base Model: Qwen2.5-7B}} \\
        Vanilla Base Model & 71.8 & 83.5 & 58.1 & 25.3 & 59.7 \\
        LQ Set (123786) & 69.3 & 77.5 & 58.6 & 55.4 & 65.2 \\
        HQ Set (131247) & 72.2 & 79.0 & 60.4 & 60.2 & 68.0 \\
        Full Set (300932) & 72.0 & 78.0 & 59.8 & 65.2 & 68.8 \\
        \noalign{%
            \setlength{\arrayrulewidth}{1pt}
            \setlength{\dashlinedash}{5pt}
            \setlength{\dashlinegap}{2pt}
        }%
        \cdashline{1-6} 
        Intra-20\% & 71.2 & 86.5 & 58.1 & 67.8 & \textbf{\textcolor{red}{70.9 (+5.7)}} \\
        Intra-40\% & 71.0 & 84.5 & 59.4 & 62.4 & \textbf{69.3 (+4.1)} \\
        Intra-60\% & 70.6 & 81.5 & 61.4 & 69.9 & \textbf{\textcolor{red}{70.9 (+5.7)}} \\
        \multicolumn{6}{>{\columncolor{lightgreen}}c}{\textbf{Base Model: Mistral-7B-v0.3}} \\
        Vanilla Base Model & 59.7 & 38.0 & 47.6 & 54.8 & 50.0 \\
        LQ Set (123786) &  47.5 & 43.5 & 52.7 & 41.1 & 46.2 \\
        HQ Set (131247) & 58.4 & 46.0 & 55.5 & 52.5 & \textbf{53.1} \\
        Full Set (300932) & 60.0 & 43.5 & 52.5 & 53.4 & 52.4 \\
        \noalign{%
            \setlength{\arrayrulewidth}{1pt}
            \setlength{\dashlinedash}{5pt}
            \setlength{\dashlinegap}{2pt}
        }%
        \cdashline{1-6} 
        Intra-20\% & 59.6 & 40.0 & 52.9 & 56.1 & 52.2 (+6.0) \\
        Intra-40\% & 59.9 & 39.5 & 54.5 & 55.1 & 52.3 (+6.1)\\
        Intra-60\% & 60.1 & 43.5 & 53.5 & 57.3 & \textbf{\textcolor{red}{53.6 (+7.4)}} \\
        \multicolumn{6}{>{\columncolor{lightgreen}}c}{\textbf{Base Model: Llama-3.1-8B}} \\
        Vanilla Base Model & 64.1 & 58.0 & 55.3 & 22.1 & 49.9 \\
        LQ Set (123786) & 52.7 & 57.0 & 61.0 & 43.9 & 53.7 \\
        HQ Set (131247) & 62.3 & 57.5 & 59.3 & 58.9 & 59.5 \\
        Full Set (300932) & 63.5 & 61.0 & 59.1 & 62.8 & \textbf{\textcolor{red}{61.6}} \\
        \noalign{%
            \setlength{\arrayrulewidth}{1pt}
            \setlength{\dashlinedash}{5pt}
            \setlength{\dashlinegap}{2pt}
        }%
        \cdashline{1-6} 
        Intra-20\% & 63.9 & 54.5 & 57.5 & 52.0 & 57.0 (+3.3) \\
        Intra-40\% & 64.0 & 59.5 & 60.6 & 53.9 & 59.5 (+5.8)\\
        Intra-60\% & 63.6 & 56.5 & 59.6 & 60.1 & \textbf{60.0 (+6.3)} \\
    \bottomrule
    \end{tabular}
\end{table*}

\begin{table*}[ht]
    \setlength{\aboverulesep}{0pt}
    \setlength{\belowrulesep}{0pt}
    
    \caption{Data-size ablation of inter-cluster reconstructed samples across four downstream benchmarks using Mistral-7B-v0.3 and LLaMA-3.1-8B as fine-tuning base models. Inter-$x$\% denotes the use of $x$\% of the 39,400 inter-cluster reconstructed samples. The best and second-best average scores within each base-model setting are highlighted in \textbf{\textcolor{red}{bold red}} and \textbf{bold black}, respectively. Values in parentheses indicate the absolute differences in average performance relative to the corresponding LQ Set.}

    \label{appendix:ablation_inter_mistral_llama}
    \centering
    \begin{tabular}{lccccc}
        \toprule
        \textbf{Dataset} & \makecell{\textbf{MMLU} \\ (EM $\uparrow$)} & \makecell{\textbf{GSM8K} \\ (EM $\uparrow$)} & \makecell{\textbf{BBH} \\ (EM $\uparrow$)} & \makecell{\textbf{TydiQA} \\ (1-Shot F1 $\uparrow$)} & \textbf{Average $\uparrow$}\\
        \midrule
        \multicolumn{6}{>{\columncolor{lightgreen}}c}{\textbf{Base Model: Mistral-7B-v0.3}} \\
        Vanilla Base Model & 59.7 & 38.0 & 47.6 & 54.8 & 50.0 \\
        LQ Set (123786) &  47.5 & 43.5 & 52.7 & 41.1 & 46.2 \\
        HQ Set (131247) & 58.4 & 46.0 & 55.5 & 52.5 & 53.1 \\
        Full Set (300932) & 60.0 & 43.5 & 52.5 & 53.4 & 52.4 \\
        \noalign{%
            \setlength{\arrayrulewidth}{1pt}
            \setlength{\dashlinedash}{5pt}
            \setlength{\dashlinegap}{2pt}
        }%
        \cdashline{1-6} 
        Inter-20\% & 59.4 & 37.0 & 54.3 & 57.5 & 52.1 (+5.9) \\
        Inter-40\% & 59.1 & 42.5 & 51.8 & 57.1 & 52.6 (+6.4)\\
        Inter-60\% & 58.9 & 45.0 & 52.1 & 56.3 & 53.1 (+6.9)\\
        Inter-80\% & 58.9 & 47.0 & 52.2 & 56.7 & \textbf{53.7 (+7.5)}\\
        Inter-100\% & 58.0 & 49.0 & 53.1 & 58.1 & \textbf{\textcolor{red}{54.6 (+8.4)}} \\
        \multicolumn{6}{>{\columncolor{lightgreen}}c}{\textbf{Base Model: Llama-3.1-8B}} \\
        Vanilla Base Model & 64.1 & 58.0 & 55.3 & 22.1 & 49.9 \\
        LQ Set (123786) & 52.7 & 57.0 & 61.0 & 43.9 & 53.7 \\
        HQ Set (131247) & 62.3 & 57.5 & 59.3 & 58.9 & \textbf{59.5} \\
        Full Set (300932) & 63.5 & 61.0 & 59.1 & 62.8 & \textbf{\textcolor{red}{61.6}} \\
        \noalign{%
            \setlength{\arrayrulewidth}{1pt}
            \setlength{\dashlinedash}{5pt}
            \setlength{\dashlinegap}{2pt}
        }%
        \cdashline{1-6} 
        Inter-20\% & 63.6 & 55.0 & 57.7 & 53.2 & 57.4 (+3.7) \\
        Inter-40\% & 62.1 & 58.5 & 58.6 & 53.4 & 58.2 (+4.5) \\
        Inter-60\% & 62.3 & 55.5 & 57.8 & 55.9 & 57.9 (+4.2) \\
        Inter-80\% & 62.3 & 55.5 & 58.4 & 55.8 & 58.0 (+4.3) \\
        Inter-100\% & 61.9 & 60.5 & 59.9 & 54.0 & 59.1 (+5.4) \\
    \bottomrule
    \end{tabular}
\end{table*}


\begin{table*}[ht]
    \setlength{\aboverulesep}{0pt}
    \setlength{\belowrulesep}{0pt}
    
    \caption{Performance comparison across five downstream benchmarks using LLaMA-3.1-8B as the fine-tuning base model. HQ+Inter-$x$\% denotes the combination of the complete HQ Set (131,247 samples) with $x$\% of the 39,400 inter-cluster reconstructed samples. The best and second-best average scores are highlighted in \textbf{\textcolor{red}{bold red}} and \textbf{bold black}, respectively. Values in parentheses indicate the absolute differences in average performance relative to the HQ Set.}

    \label{appendix:ablation_hq+inter_llama}
    \centering
    \begin{tabular}{lcccccc}
        \toprule
        \textbf{Dataset} & \makecell{\textbf{MMLU} \\ (EM $\uparrow$)} & \makecell{\textbf{TruthfulQA} \\ \rev{(I $\times$ T $\uparrow$)}} & \makecell{\textbf{GSM8K} \\ (EM $\uparrow$)} & \makecell{\textbf{BBH} \\ (EM $\uparrow$)} & \makecell{\textbf{TydiQA} \\ (1-Shot F1 $\uparrow$)} & \textbf{Average $\uparrow$}\\
        \midrule
        \multicolumn{7}{>{\columncolor{lightgreen}}c}{\textbf{Base Model: Llama-3.1-8B}} \\
        Vanilla Base Model & 64.1 & 32.9 & 58.0 & 55.3 & 22.1 & 46.5 \\
        LQ Set (123786) & 52.7 & 44.3 & 57.0 & 61.0 & 43.9 & 51.8 \\
        HQ Set (131247) & 62.3 & 41.8 & 57.5 & 59.3 & 58.9 & 56.0 \\
        Full Set (300932) & 63.5 & 42.0 & 61.0 & 59.1 & 62.8 & 57.7 \\
        \noalign{%
            \setlength{\arrayrulewidth}{1pt}
            \setlength{\dashlinedash}{5pt}
            \setlength{\dashlinegap}{2pt}
        }%
        \cdashline{1-7} 
        HQ+Inter-40\% & 62.0 & 45.3 & 58.0 & 59.9 & 55.0 & 56.0 (+0.0) \\
        HQ+Inter-60\% & 63.3 & 44.6 & 61.0 & 62.4 & 57.5 & \textbf{57.8 (+1.8)}\\
        HQ+Inter-100\% & 62.5 & 44.6 & 65.0 & 60.6 & 57.8 & \textbf{\textcolor{red}{58.1 (+2.1)}} \\
    \bottomrule
    \end{tabular}
\end{table*}


\begin{table*}[ht]
    \setlength{\aboverulesep}{0pt}
    \setlength{\belowrulesep}{0pt}
    
    \caption{Performance comparison across five downstream benchmarks using Qwen2.5-7B as the fine-tuning base model. HQ+Intra-$x$\% denotes the combination of the complete HQ Set (131,247 samples) with $x$\% of the 15,488 intra-cluster reconstructed samples. The best and second-best average scores are highlighted in  \textbf{\textcolor{red}{bold red}} and \textbf{bold black}, respectively.}

    \label{appendix:ablation_hq+intra_qwen}
    \centering
    \begin{tabular}{lcccccc}
        \toprule
        \textbf{Dataset} & \makecell{\textbf{MMLU} \\ (EM $\uparrow$)} & \makecell{\textbf{TruthfulQA} \\ \rev{(I $\times$ T $\uparrow$)}} & \makecell{\textbf{GSM8K} \\ (EM $\uparrow$)} & \makecell{\textbf{BBH} \\ (EM $\uparrow$)} & \makecell{\textbf{TydiQA} \\ (1-Shot F1 $\uparrow$)} & \textbf{Average $\uparrow$}\\
        \midrule
        \multicolumn{7}{>{\columncolor{lightgreen}}c}{\textbf{Base Model: Qwen2.5-7B}} \\
        Vanilla Base Model & 71.8 & 11.1 & 83.5 & 58.1 & 25.3 & 50.0 \\
        LQ Set (123786) & 69.3 & 43.5 & 77.5 & 58.6 & 55.4 & 60.9 \\
        HQ Set (131247) & 72.2 & 38.2 & 79.0 & 60.4 & 60.2 & 62.0 \\
        Full Set (300932) & 72.0 & 41.4 & 78.0 & 59.8 & 65.2 & \textbf{63.3} \\
        \noalign{%
            \setlength{\arrayrulewidth}{1pt}
            \setlength{\dashlinedash}{5pt}
            \setlength{\dashlinegap}{2pt}
        }%
        \cdashline{1-7} 
        HQ+Intra-40\% & 72.0 & 36.3 & 72.5 & 58.7 & 63.6 & 60.6 \\
        HQ+Intra-60\% & 72.0 & 36.5 & 80.0 & 58.1 & 64.1 & 62.1 \\
        HQ+Intra-100\% & 71.0 & 45.8 & 76.5 & 59.4 & 64.7 & \textbf{\textcolor{red}{63.5}} \\
    \bottomrule
    \end{tabular}
\end{table*}

\begin{table*}[ht]
    \setlength{\aboverulesep}{0pt}
    \setlength{\belowrulesep}{0pt}
    
    \caption{Performance comparison across five downstream benchmarks using Qwen2.5-7B as the fine-tuning base model. HQ+Inter-$x$\% denotes the combination of the complete HQ Set (131,247 samples) with $x$\% of the 39,400 inter-cluster reconstructed samples. The best and second-best average scores are highlighted in \textbf{\textcolor{red}{bold red}} and \textbf{bold black}, respectively.}

    \label{appendix:ablation_hq+inter_mistral_qwen}
    \centering
    \begin{tabular}{lcccccc}
        \toprule
        \textbf{Dataset} & \makecell{\textbf{MMLU} \\ (EM $\uparrow$)} & \makecell{\textbf{TruthfulQA} \\ \rev{(I $\times$ T $\uparrow$)}} & \makecell{\textbf{GSM8K} \\ (EM $\uparrow$)} & \makecell{\textbf{BBH} \\ (EM $\uparrow$)} & \makecell{\textbf{TydiQA} \\ (1-Shot F1 $\uparrow$)} & \textbf{Average $\uparrow$}\\
        \midrule
        \multicolumn{7}{>{\columncolor{lightgreen}}c}{\textbf{Base Model: Qwen2.5-7B}} \\
        Vanilla Base Model & 71.8 & 11.1 & 83.5 & 58.1 & 25.3 & 50.0 \\
        LQ Set (123786) & 69.3 & 43.5 & 77.5 & 58.6 & 55.4 & 60.9 \\
        HQ Set (131247) & 72.2 & 38.2 & 79.0 & 60.4 & 60.2 & 62.0 \\
        Full Set (300932) & 72.0 & 41.4 & 78.0 & 59.8 & 65.2 & \textbf{63.3} \\
        \noalign{%
            \setlength{\arrayrulewidth}{1pt}
            \setlength{\dashlinedash}{5pt}
            \setlength{\dashlinegap}{2pt}
        }%
        \cdashline{1-7} 
        HQ+Inter-20\% & 71.6 & 37.3 & 73.0 & 59.5 & 64.0 & 61.1 \\
        HQ+Inter-40\% & 71.7 & 42.0 & 75.0 & 57.0 & 61.7 & 61.5 \\
        HQ+Inter-60\% & 71.6 & 44.8 & 74.0 & 57.1 & 63.1 & 62.1 \\
        HQ+Inter-80\% & 71.4 & 41.9 & 81.5 & 59.7 & 64.5 & \textbf{\textcolor{red}{63.8}} \\
    \bottomrule
    \end{tabular}
\end{table*}


\begin{table*}[ht]
    \setlength{\aboverulesep}{0pt}
    \setlength{\belowrulesep}{0pt}
    
    \caption{Performance comparison across five downstream benchmarks using Mistral-7B-v0.3, LLaMA-3.1-8B, and Qwen2.5-7B as fine-tuning base models. HQ+Ours-$x$\% denotes the combination of the complete HQ Set (131,247 samples) with $x$\% of our reconstructed dataset (54,888 samples), comprising both intra-cluster and inter-cluster reconstructions. The best and second-best average scores within each base-model setting are highlighted in \textbf{\textcolor{red}{bold red}} and \textbf{bold black}, respectively.}

    \label{appendix:ablation_hq+entp_mistral_llama_qwen}
    \centering
    \begin{tabular}{lcccccc}
        \toprule
        \textbf{Dataset} & \makecell{\textbf{MMLU} \\ (EM $\uparrow$)} & \makecell{\textbf{TruthfulQA} \\ \rev{(I $\times$ T $\uparrow$)}} & \makecell{\textbf{GSM8K} \\ (EM $\uparrow$)} & \makecell{\textbf{BBH} \\ (EM $\uparrow$)} & \makecell{\textbf{TydiQA} \\ (1-Shot F1 $\uparrow$)} & \textbf{Average $\uparrow$}\\
        \midrule
        \multicolumn{7}{>{\columncolor{lightgreen}}c}{\textbf{Base Model: Mistral-7B-v0.3}} \\
        Vanilla Base Model & 59.7 & 30.4 & 38.0 & 47.6 & 54.8 & 46.1 \\
        LQ Set (123786) & 47.5 & 43.7 & 43.5 & 52.7 & 41.1 & 45.7 \\
        HQ Set (131247) & 58.4 & 39.2 & 46.0 & 55.5 & 52.5 & 50.3 \\
        Full Set (300932) & 60.0 & 43.5 & 43.5 & 52.5 & 53.4 & 50.6 \\
        \noalign{%
            \setlength{\arrayrulewidth}{1pt}
            \setlength{\dashlinedash}{5pt}
            \setlength{\dashlinegap}{2pt}
        }%
        \cdashline{1-7} 
        HQ+Ours-40\% & 57.8 & 42.4 & 45.0 & 54.4 & 55.2 & \textbf{51.0} \\
        HQ+Ours-60\% & 58.2 & 45.8 & 45.0 & 52.4 & 54.5 & \textbf{\textcolor{red}{51.2}} \\
        HQ+Ours-100\% & 57.2 & 47.2 & 46.0 & 52.2 & 53.4 & \textbf{\textcolor{red}{51.2}} \\
        \multicolumn{7}{>{\columncolor{lightgreen}}c}{\textbf{Base Model: Llama-3.1-8B}} \\
        Vanilla Base Model & 64.1 & 32.9 & 58.0 & 55.3 & 22.1 & 46.5 \\
        LQ Set (123786) & 52.7 & 44.3 & 57.0 & 61.0 & 43.9 & 51.8 \\
        HQ Set (131247) & 62.3 & 41.8 & 57.5 & 59.3 & 58.9 & 56.0 \\
        Full Set (300932) & 63.5 & 42.0 & 61.0 & 59.1 & 62.8 & \textbf{\textcolor{red}{57.7}} \\
        \noalign{%
            \setlength{\arrayrulewidth}{1pt}
            \setlength{\dashlinedash}{5pt}
            \setlength{\dashlinegap}{2pt}
        }%
        \cdashline{1-7} 
        HQ+Ours-40\% & 62.5 & 44.6 & 59.0 & 58.7 & 57.4 & 56.4 \\
        HQ+Ours-60\% & 62.7 & 43.0 & 59.5 & 61.3 & 61.5 & \textbf{57.6} \\
        HQ+Ours-100\% & 62.7 & 47.2 & 61.5 & 61.1 & 56.1 & \textbf{\textcolor{red}{57.7}} \\
        \multicolumn{7}{>{\columncolor{lightgreen}}c}{\textbf{Base Model: Qwen2.5-7B}} \\
        Vanilla Base Model & 71.8 & 11.1 & 83.5 & 58.1 & 25.3 & 50.0 \\
        LQ Set (123786) & 69.3 & 43.5 & 77.5 & 58.6 & 55.4 & 60.9 \\
        HQ Set (131247) & 72.2 & 38.2 & 79.0 & 60.4 & 60.2 & 62.0 \\
        Full Set (300932) & 72.0 & 41.4 & 78.0 & 59.8 & 65.2 & \textbf{\textcolor{red}{63.3}} \\
        \noalign{%
            \setlength{\arrayrulewidth}{1pt}
            \setlength{\dashlinedash}{5pt}
            \setlength{\dashlinegap}{2pt}
        }%
        \cdashline{1-7} 
        HQ+Ours-40\% & 72.1 & 44.3 & 75.5 & 59.4 & 60.7 & 62.4 \\
        HQ+Ours-60\% & 72.2 & 43.0 & 74.0 & 60.9 & 59.6 & 61.9 \\
        HQ+Ours-100\% & 71.7 & 45.1 & 76.0 & 59.0 & 63.4 & \textbf{63.1} \\
    \bottomrule
    \end{tabular}
\end{table*}

\clearpage

\clearpage

\clearpage

\clearpage




\clearpage

\end{document}